\documentclass[Afour,sageh,times]{sagej}
\usepackage{moreverb,url}

\usepackage[colorlinks,bookmarksopen,bookmarksnumbered,citecolor=red,urlcolor=red]{hyperref}

\newcommand\BibTeX{{\rmfamily B\kern-.05em \textsc{i\kern-.025em b}\kern-.08em
T\kern-.1667em\lower.7ex\hbox{E}\kern-.125emX}}

\usepackage{graphics} 
\usepackage{epsfig} 
\usepackage{times} 
\usepackage{amsmath} 
\usepackage{amssymb}  
\usepackage{bm}
\usepackage{multirow}
\usepackage{float}
\usepackage{subfig}
\usepackage{url}
\usepackage{flushend}
\usepackage{booktabs} 
\usepackage{bbding}
\usepackage{xcolor} 
\newcommand{\mathcolorbox}[2]{\colorbox{#1}{$\displaystyle #2$}}
\usepackage{soul} 
\definecolor{lightblue}{rgb}{0.812, 0.906, 0.922}
\definecolor{lightgreen}{rgb}{0.651, 0.780, 0.415}
\definecolor{lightorange}{rgb}{0.984, 0.757, 0.678}
\definecolor{FFDF70}{rgb}{1.000, 0.875, 0.439}
\definecolor{F46E49}{rgb}{0.956863, 0.431373, 0.286275}
\definecolor{81D0BB}{rgb}{0.505882, 0.815686, 0.733333}
\definecolor{CFE7EB}{rgb}{0.811765, 0.905882, 0.921569}
\definecolor{FBC1AD}{rgb}{0.984314, 0.756863, 0.678431}
\definecolor{7DABCF}{rgb}{0.490196, 0.670588, 0.811765}
\definecolor{377E22}{rgb}{0.216, 0.494, 0.133}
\definecolor{E76D7E}{rgb}{0.909804, 0.427451, 0.494118}
\definecolor{F2B1BB}{rgb}{0.949, 0.694, 0.733}
\definecolor{B6DDBB}{rgb}{0.714, 0.867, 0.733}
\definecolor{45AAB4}{rgb}{0.270, 0.666, 0.706}

\usepackage{tikz}     

\newcommand{\trajectorySegment}[2]{
    {\begin{tikzpicture}[scale=0.2, baseline=-0.6ex]
        \foreach \i in {0,...,3} {
            \ifnum\i<#1
                \filldraw[fill=white] (\i,0) rectangle ++(1,1);
            \else\ifnum\i<#2
                \filldraw[fill=gray] (\i,0) rectangle ++(1,1);
            \else
                \filldraw[fill=white] (\i,0) rectangle ++(1,1);
            \fi\fi
        }
    \end{tikzpicture}}
}

\def\volumeyear{2016}

\begin{document}

\runninghead{Song et al.}

\title{\textcolor{F46E49}{R}obot \textcolor{F46E49}{T}rajectron \textcolor{F46E49}{V2}: A Probabilistic Shared Control Framework for Navigation}

\author{Pinhao Song$^{1}$, Yurui Du$^{2}$, Ophelie Saussus$^3$, Sofie De Schrijver$^{3,4}$, Irene Caprara$^3$, Peter Janssen$^3$, Renaud Detry$^{1,2}$}

\affiliation{$^1$KU Leuven, Dept. Mechanical Engineering, Research unit Robotics, Automation and Mechatronics. \\
$^2$KU Leuven, Dept. Electrical Engineering, Research unit Processing Speech and Images.\\
$^3$KU Leuven, Dept. Neurosciences, Laboratory for Neuro- and Psychophysiology.\\
$^4$Department of Electrical and Computer Engineering, University of Washington.\\
Corresponding author: Pinhao Song, \textit{pinhao.song@kuleuven.be}
}


\begin{abstract}


We propose a probabilistic shared-control solution for navigation, called Robot Trajectron V2 (RT-V2), that enables accurate intent prediction and safe, effective assistance in human–robot interaction. RT-V2 jointly models a user’s long-term behavioral patterns and their noisy, low-dimensional control signals by combining a prior intent model with a posterior update that accounts for real-time user input and environmental context. The prior captures the multimodal and history-dependent nature of user intent using recurrent neural networks and conditional variational autoencoders, while the posterior integrates this with uncertain user commands to infer desired actions. We conduct extensive experiments to validate RT-V2 across synthetic benchmarks, human–computer interaction studies with keyboard input, and brain–machine interface experiments with non-human primates. Results show that RT-V2 outperforms the state of the art in intent estimation, provides safe and efficient navigation support, and adequately balances user autonomy with assistive intervention. By unifying probabilistic modeling, reinforcement learning, and safe optimization, RT-V2 offers a principled and generalizable approach to shared control for diverse assistive technologies. Code will be available in \url{https://mousecpn.github.io/RTV2_page/}.

\end{abstract}


\maketitle

\section{Introduction}
Shared control is a collaborative approach between a human operator and a robot, designed to reduce operator workload and facilitate the more efficient and safer completion of complex tasks with the robot. This approach is widely used in various fields, such as subsea maintenance, surgery, driving, and assistive devices. In the context of assistive devices, it enables individuals with disabilities to regain autonomy through technologies like robotic wheelchairs and manipulators. The key challenge here is that while robots have many degrees of freedom, the input devices available to disabled users are typically low-DoF and noisy due to the nature of their disabilities. Examples include chin joysticks \citep{rulik2022control} and neural implants \citep{hochberg2012reach}. Using these devices to control a wheelchair or robot arm can be slow, tiring, and prone to errors. Shared control addresses these issues by identifying the user's intent and facilitating smoother, more effortless goal achievement. To enhance this process, shared control often utilizes additional sources of information to interpret user input in context. For instance, cameras can provide images of the surrounding environment to aid in this interpretation.


Accurately assisting and executing a user's desired action requires understanding their intent. 
Predicting user intent is particularly challenging due to three defining characteristics:
(i) \textbf{Multi-modal}: a user may approach a goal through different sub-optimal paths; 
(ii) \textbf{Non-Markovian}: past experiences continue to influence current actions (e.g., a previous car accident may make a driver more cautious); 
(iii) \textbf{Non-stationary}: user performance fluctuates, improving when they are focused and declining when they are fatigued.
This complexity creates a dynamic interplay between the user's and the assistive controller's authority. Users often require more control to effectively convey their intent, while assistive controllers may need increased authority to reduce the user's effort. This tension gives rise to two critical dilemmas frequently faced by assistive controllers:

\begin{figure}[!t]
  \centering
  \includegraphics[width=\linewidth]{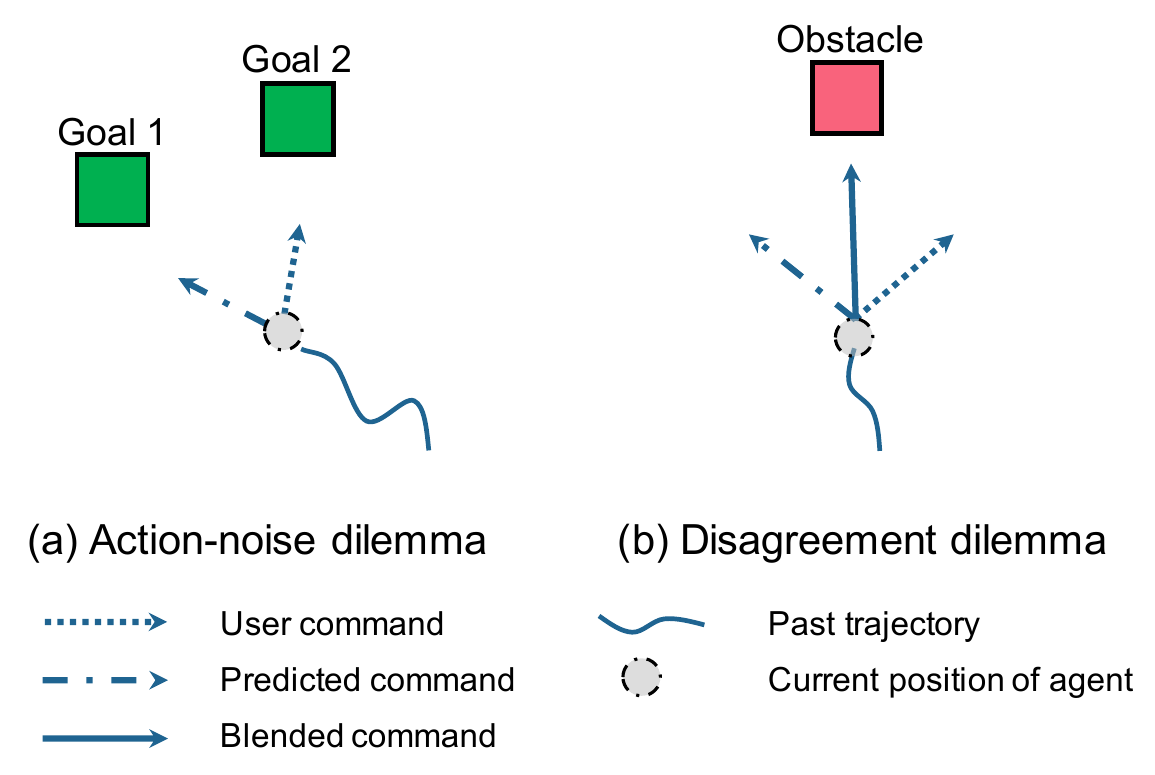}
  \caption{
  In this figure, the dot represents a mobile robot under shared control by an autonomous agent and a human user with an input device. The autonomous agent is designed to assist the user by identifying their intent through contextual cues.
    The figure illustrates two common dilemmas in shared control systems:
    (a) The user intends to achieve one of two potential goals, but the autonomous agent is perplexed by a sudden change in the user's command.
    (b) The user's goal is to avoid an obstacle, which conflicts with the autonomous agent's predicted path due to the task's multi-modal nature. See text for details.
  }
  \label{fig: dilemma}
  \vspace{-0.2cm}
\end{figure}

\textbf{(i) Action--noise dilemma}: 
An assistive controller must execute the user’s intended actions while filtering out noise from the user interface. However, this task is complicated by the challenge of distinguishing noise from the effect of the three defining characteristics of user intent listed above. 
Over-reliance on the assistive controller may reduce noise effectively but risks suppressing genuine changes in the user’s intent. Conversely, relying more on the user preserves their autonomy but fails to alleviate their control burden or input noise. For example, as illustrated in Fig.~\ref{fig: dilemma}~(a), at the current timestep, the assistive controller predicts an action toward goal 1, while the user issues a command toward goal 2. Treating the user’s command as noise and filtering it out may result in the user reaching an unintended goal if the command represents their true intent. On the other hand, fully relying on the user’s command offers no assistance if the command is indeed noise.
Some approaches attempt to address this dilemma heuristically \citep{demeester2008user,RT}. By introducing a disagreement threshold, the system differentiates between intended actions and noise, returning control to the user when the threshold is exceeded. However, this method does not fundamentally solve the problem, as there is no guarantee that a user command exceeding the threshold represents true intent. Additionally, the process of setting such a threshold lacks theoretical justification.
To fully resolve the action–noise dilemma, the assistive controller must adaptively blend actions by considering the uncertainties of both the user and the environment.

\textbf{(ii) Disagreement Dilemma}: A common approach to shared control involves linearly blending the user’s commands with those of the controller, facilitated by an arbitrator \citep{policyblending, RT, maeda2022blending}. However, as Trautman points out \citep{psc}, tasks such as collision avoidance often allow for multiple equally optimal trajectories due to the multi-modal nature of human intent. The user may select any of these trajectories, which can differ from the controller's prediction.
When the user disagrees with the assistive controller at a given timestep, blending a safe user command with a safe controller-proposed action may unintentionally result in an unsafe shared action. For example, as shown in Fig.~\ref{fig: dilemma}~(b), the assistive controller predicts an action to the left, while the user commands movement to the right. A linear blend of these two actions could lead to a collision with the obstacle.
Existing approaches, such as those based on probabilistic models \citep{psc}, constraint-based shared control \citep{iregui2021reconfigurable}, and model predictive control (MPC) \citep{lu2019model}, attempt to address this issue by implicitly blending policies. However, these methods often reduce to linear blending, limiting their ability to fully resolve the disagreement dilemma.
To effectively tackle this challenge, the assistive controller should adopt a multi-modal blending strategy. For instance, it could generate multiple trajectory proposals and blend the user’s command with the proposal most aligned with their intent, ensuring both safety and responsiveness.

In this paper, we propose an assistive controller named \emph{Robot Trajectron V2} (RT-V2) for navigation tasks. RT-V2 is designed within a Bayesian framework, as:
\begin{equation}
    p(i|u,c) \propto  p(u|i)p(i|c) \label{eq: bayes}
\end{equation}
where $i$ denotes the user's intent, $u$ denotes the current user's command, and $c$ denotes environmental context. 
$p(i|c)$ is the prior model, $p(u|i)$ is the likelihood, which models the uncertainty of the user interface, and $p(i|u,c)$ is the posterior we want to model. 
The prior model $p(i|c)$ is trained to imitate the user’s behavior patterns, $p(u|i)$ is the likelihood which models the uncertainty of the user interface, and the posterior estimation of the user’s intent $p(i|u,c)$ combines prior uncertainty with the uncertainty in commands from the user interface. This dual consideration of prior and posterior uncertainties implicitly addresses the action-noise dilemma.
The prior model is constructed using recurrent neural networks and a conditional variational auto-encoder framework and trained in an imitation learning (IL) style, effectively capturing the non-Markovian and multi-modal characteristics of user behavior. The multi-modality of the prior estimate enables seamless blending with the user’s current command, facilitating accurate estimation of intended actions and resolving the disagreement dilemma. Since the prior model is trained by IL, it suffers from causal confusion because it learns correlations between states, actions, and rewards from demonstrations without distinguishing causal relationships from spurious ones. Since the model typically minimizes a supervised loss between the expert's actions and their own predictions, it might copy associations that exist in the data but are not actually responsible for successful task execution. To overcome the causal confusion inherent in IL, caused by the lack of feedback from the environment, we introduce Imagined Rollout Reinforcement Learning. In this approach, RT-V2 simulates interactions with the environment to obtain reward feedback, thereby improving its autonomous navigation capabilities. Furthermore, a sampling-based optimization method with safety constraints is employed to ensure the controller's actions are collision-free and safe. We conduct comprehensive experiments to validate the effectiveness of RT-V2. Trajectory prediction experiments demonstrate its high accuracy in intent estimation, while navigation experiments confirm its robust navigation performance. We conducted shared autonomy experiments with human users who provided input via a keyboard interface, and with monkeys equipped with a BMI interface (neural implants). Our experiments reveal that RT-V2 achieves optimal shared control in terms of agreeability, safety, and efficiency.

In summary:
\begin{itemize}
    \item Our paper proposes Robot Trajectron V2 (RT-V2), a Bayesian-based assistive controller designed for navigation tasks. RT-V2 models user behavior using a prior trained on past data and a posterior that adapts to real-time user commands, addressing both action-noise and disagreement dilemmas.
    \item The prior model is built using a recurrent neural network and a conditional variational autoencoder (CVAE), enabling it to capture multi-modal and non-Markovian aspects of human intent. This enhances the controller’s ability to accurately interpret and blend user commands in a dynamic shared control setting.
    \item To overcome causal confusion in imitation learning, we introduce Imagined Rollout Reinforcement Learning, where RT-V2 simulates future interactions to receive reward signals and refine its autonomous navigation capabilities.
    \item A sampling-based trajectory optimization method with safety constraints is employed to ensure the controller's actions are collision-free.
\end{itemize}

The novel contributions of the paper are:
\begin{itemize}
    \item A shared-control model grounded in a probabilistic formulation of the intention prior and posterior, and their acquisition from data via a combination of imitation learning, reinforcement learning, and sampling-based optimization.
    \item Extensive experiments show that RT-V2 achieves high accuracy in intent estimation and safe, efficient navigation. Tests with human users (keyboard interface) and monkey users (BMI interface) demonstrate its effectiveness in optimizing shared autonomy with respect to agreeability, safety, and efficiency.
\end{itemize}

\section{Related Work}

\subsection{Shared Control}
Shared control involves the cooperative determination of a policy by both an autonomous agent and the user, with the goal of improving performance and safety in robot manipulation. This approach is widely utilized in various applications, including assistive driving \citep{lu2019model}, wheelchair control \citep{vanhooydonck2003shared}, and BMI-controlled manipulation \citep{xu2020shared}.
Shared control can be divided into three key components: intent estimation, planning, and policy blending. The controller first identifies the user’s intent based on historical \citep{RT,ziebart2008maximum} and contextual information \citep{zhang2022human}. It then generates planning proposals \citep{demeester2008user, maeda2022blending} and blends the user’s command with the controller’s proposed action \citep{policyblending}.

To efficiently assist users, it is crucial for shared control systems to understand their intent. Early studies \citep{vanhooydonck2003shared,goodrich2003seven,demeester2008user} suggest that requiring users to explicitly specify their intent is inefficient and sometimes unfeasible for several reasons: First, in tasks such as assistive wheelchair control, users reactively determine their desired path based on the dynamically changing environment. Second, for some user groups, explicitly stating their goal can be cognitively and physically challenging, or even impossible (e.g., in BMI-controlled settings). Consequently, contemporary research places emphasis on harnessing implicit cues such as user commands and environmental sensing to deduce user intent.
A widely used framework for intent estimation is Inverse Reinforcement Learning (IRL), also known as Inverse Optimal Control. IRL leverages a parameterized family of reward functions to fit human demonstration datasets, inferring the user’s intended goal by modeling them as an intent-driven agent maximizing cumulative rewards. The most notable method within this framework is MaxEnt IRL \citep{ziebart2008maximum}, which has inspired numerous shared control approaches and shown promising performance even in complex, cluttered environments \citep{policyblending, javdani2018shared, muelling2017autonomy, gottardi2022shared, fu2025a}.
However, as Ivanovic et al. \citep{ivanovic2020multimodal} point out, multi-modal planning in IRL may be computationally intractable due to its reliance on the unnormalized log-probability density function. Beyond IRL, human-in-the-loop methods have also been proposed to learn reward functions \citep{reddy2018shared, wu2023human}. These methods, while effective, require users to provide additional feedback during training, making them labor-intensive and impractical in certain scenarios.

Planning involves guiding the robot to the intended goal while ensuring collision avoidance and adherence to constraints. Various methods can be utilized for planning in shared control, including Artificial Potential Fields (APFs) \citep{gottardi2022shared}, Model Predictive Control (MPC) \citep{lu2019model}, and the Dynamic Window Approach (DWA) \citep{lei2022intention}. Additionally, reinforcement learning agents are frequently employed as planners in shared control scenarios \citep{reddy2018shared, singh2023probabilistic, backman2023reinforcement}.
Another prevalent approach is constraint-based shared control, which formulates the problem as a constraint-based optimization task \citep{sct, iregui2021reconfigurable}. The primary challenge in shared control planning is effectively coordinating with the user’s multi-modal behaviors. If the planner generates only a single trajectory toward the most likely goal, it risks creating a disagreement dilemma. To mitigate this, planners must consider multiple trajectories for all potential goals. However, this approach is computationally expensive, as many of the generated trajectories may remain unused.

Policy Blending often begins with linear blending \citep{policyblending, gottardi2022shared, luo2024human}, where accurately estimating the arbitrator to balance the user’s command and the autonomous agent's action is critical. Compared to constant arbitrators \citep{gottardi2022shared} or those based on distance metrics \citep{xu2020shared}, confidence-based arbitrators generally perform better as they incorporate uncertainty in intent estimation \citep{RT, policyblending}.
Some researchers frame policy blending within a probabilistic context. For instance, Trautman \citep{psc} introduced the Probabilistic Shared Control (PSC) framework, which simplifies to linear blending and has been applied to tasks like wheelchair navigation by Zezh et al. \citep{pscwheel, psccompare}. However, many of these approaches model user intent as a Gaussian distribution, with the user's current input as the mean. This simplification overlooks the multi-modal and non-stationary characteristics of user behavior, limiting their effectiveness in dynamic, real-world scenarios.
Demeester et al. \citep{demeester2008user} proposed a Bayesian framework for maneuvering assistance in wheelchair navigation. This approach models and estimates complex user intents while explicitly accounting for the uncertainty associated with the user’s intent. Beyond intent estimation, the framework incorporates user-specific characteristics and intent uncertainty into the decision-making process for assistive actions. By doing so, it effectively addresses the multi-modality and non-stationarity inherent in user behavior.
Some approaches leverage reinforcement learning to frame shared control \citep{reddy2018shared, schaff2020residual, oh2020natural}, reducing the reliance on predefined assumptions about user behavior and the environment. This enables the end-to-end training of twin policies. However, these methods typically require extensive user-in-the-loop data collection, which can be labor-intensive. To address this, some researchers simulate user interactions to generate data \citep{reddy2018shared, backman2023reinforcement}, while others allow the system to take sub-optimal actions within an acceptable deviation from the user's input \citep{schaff2020residual, oh2020natural}. Despite these efforts, such strategies still rely on assumptions about user behavior.

In this work, we propose RT-V2, which frames shared control as imitating the user's intended decision-making process and inferring intended actions posteriorly. While RT-V2 builds on our previous work, RT \citep{RT}, it introduces significant enhancements. Unlike RT, RT-V2 incorporates contextual information and implicitly integrates intent estimation, planning, and policy blending into a unified framework. Trained on human demonstrations, RT-V2 captures the multi-modality and non-stationarity of human behaviors. It combines the uncertainty derived from past dynamics and context with the user's uncertainty to infer the posterior estimation of the user's next intended action, ultimately guiding the user toward the intended goal.

\subsection{Imitation Learning}
Imitation Learning (IL), also referred to as Learning from Demonstration (LfD) or Behavior Cloning (BC), focuses on training a policy that mimics expert behavior. IL is widely applied in various robotics tasks \citep{ILsurvey, ILsurveyDriving, rtx} due to its efficiency in data collection and training. However, a key limitation of IL is the assumption of independent and identically distributed (i.i.d.) data, which does not hold in sequential decision-making tasks, where the states encountered depend on the acting policy. This misalignment can lead to the drift problem: if the policy makes a mistake, the error compounds, potentially causing larger deviations from the expert demonstrations over time. To mitigate this, DAgger \citep{dagger} introduces a method in which the current policy collects data with human labeling throughout the training process. However, DAgger requires continuous human expert involvement, which is labor-intensive.

Another limitation of Imitation Learning (IL) is its susceptibility to causal confusion due to the absence of reward signals. Unlike reinforcement learning, where policies learn from both successes and failures, IL-trained policies do not encounter failure cases and thus struggle to establish a relationship between observations and actions, especially in environments with large observation and action spaces. To address this issue, Generative Adversarial Imitation Learning (GAIL) \citep{gail} introduces an adversarial discriminator to evaluate the current policy. Recent advancements in model-based reinforcement learning (RL) \citep{ha2018world, dreamer, tdmpc} involve learning an environment model and extracting the policy from this model. IL can benefit from this model-based RL framework, as policies trained via IL can perform imagined rollouts by interacting with the model and receiving reward signals \citep{kolev2024efficient}.

A key challenge in Imitation Learning (IL) is modeling multi-modal and non-Markovian behaviors, as human experts do not always behave consistently under identical observations, and their actions are influenced by past states. World Models \citep{ha2018world} tackle this by using Recurrent Neural Networks (RNNs) to encode state memories in a compact latent space. MILE \citep{mile} extends this concept to learn non-Markovian behaviors in autonomous driving. To model multi-modal behavior, many trajectory prediction methods \citep{trajectron, trajectron++, song2020pip}, which can also be framed as a form of imitation learning, employ latent variable models such as Conditional Variational Autoencoders (CVAE) \citep{cvae}. These models enable sampling of diverse future trajectories based on past trajectories.

In this work, the proposed RT-V2 uses RNNs to encode state memories and a CVAE framework \citep{cvae} to model multi-modal behaviors. Within a model-based imitation learning (MBIL) framework, RT-V2 can perform imagined rollouts and receive rewards by interacting with the simulated environment.

\begin{figure*}[!t]
  \centering
  \includegraphics[width=0.8\linewidth]{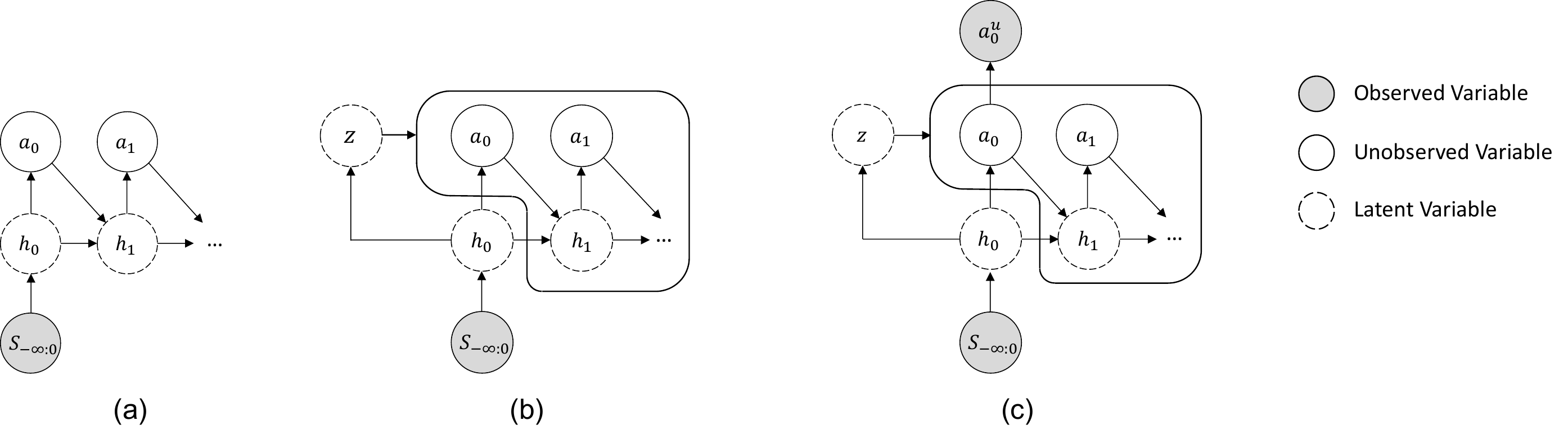}
  \caption{
  The graphical models of the decision-making process. We represent the trajectory in terms of relative timesteps, with $t=0$ denoting the current timestep. (a) We assume the user’s action is conditional independent of past actions given the past states, and that each state is conditionally dependent on the immediately preceding state and action. Besides, a latent state $\bm{h}$ is introduced to represent a memory of the past states. (b) We incorporate a latent variable $\bm{z}$ to account for the multi-modality of the user’s intent. Consequently, both $\bm{a}_t$ and the transition of $\bm{h}_t$ become conditionally dependent on $\bm{z}$ in addition. (c) We observe the current user command $\bm{a}^\text{u}_0$ from the user interface. Based on this observation, we aim to infer (i.e., posteriorly estimate) the user’s current desired action $\bm{a}_0$. See text for details.
  }
  \label{fig: graphical model}
\end{figure*}

\section{Methodology Overview}

This paper introduces an assistive controller, RT-V2, designed to help users navigate around obstacles and reach their goals. A navigation problem can be simplified as follows: in a workspace, there are $M$ points of interest (potential goals) $\{\bm{g}_m\}_{m=1}^{M}$ and $N$ obstacles $\{\bm{o}_n\}_{n=1}^{N}$. A user controls an omnidirectional (virtual) robot via a noisy interface (neural implant, joystick, ...). The assistive controller processes these noisy commands, works to separate their signal from noise and infer intent, and issues robot commands that realize the perceived user intent. 


Our method approaches shared control from a Bayesian perspective (see Eq.\ \ref{eq: bayes}). The methodology, detailed in Sections \ref{il4sc} through \ref{scwc}, is organized as follows:

\smallskip\noindent Section \ref{il4sc}: We first train an intent estimator using imitation learning. This estimator processes the robot’s historical trajectory and contextual cues to predict an expected future trajectory. Notably, at this stage, there is not yet any notion of shared control, assistance, or user input. Instead, we simply train a model in a supervised way to predict how the onset of a robot motion is likely to continue. The model's input is a ``past'' trajectory, and its output is the robot's ``future'' trajectory. Our choosing of the word \emph{intent} is motivated by our assumption that completing a trajectory implicitly captures the intention of the user or agent that generated the trajectories the model is trained on. In our Bayesian formulation, this estimator serves as the \emph{intent prior} $p(i|c)$. It contributes to the computation of the \emph{intent posterior} $p(i|u,c)$ discussed below. 

\smallskip\noindent Section \ref{irrl}: 
Because imitation learning lacks direct environmental feedback, the intent prior may produce trajectories that collide with obstacles. To address this source of causal confusion, we introduce Imagined Rollout Reinforcement Learning: We generate simulated rollouts with the intent prior, and tune trajectory prediction with reward signals provided by an approximate environment.

\smallskip\noindent Section \ref{pdm}: By contrast to Section \ref{il4sc}, this section explicitly models \emph{user input} in addition to robot trajectories. We model user intent based on both the past trajectory \emph{and} the current user input.
    Specifically, we adopt an agentic perspective, where the user provides a noisy command $u$ at each timestep. We model the likelihood $p(u|i)$ and combine it with the prior $p(i|c)$ (the intent estimator) to obtain the posterior $p(i|u,c)$.

\smallskip\noindent Section \ref{scwc}: Finally, we propose a controller built upon this intent model to ensure safe navigation. A sampling-based optimization method is used to compute collision-free actions by maximizing the posterior $p(i|u,c)$ under a no-collision constraint.

\section{Imitation Learning for the Prior Model} \label{il4sc}
In this section, we derive a prior model of the user's behavior through imitation learning. 
Specifically, we train a policy, dubbed \textit{twin policy}, to encode the user's motion behavior. The twin policy encodes how the user typically moves, or, in other words, where the user \emph{intends} to move next. Training is achieved by minimizing the discrepancy between the twin policy and a \emph{goal-conditioned} intention policy, which represents the desired user behavior for reaching a specific goal.
To mathematically formulate the problem, we model the twin policy and goal-conditioned intention user policy as stochastic distributions $\pi_{T}(\bm{a}|\bm{s})$ and $\pi_{I}(\bm{a}|\bm{s}, \bar{\bm{g}})$, respectively. Here, $\bm{a} \in \mathbb{R}^2$ is a velocity action, and $\bar{\bm{g}} \in \{\bm{g}_m\}_{m=1}^{M}$ denotes the user's intended goal. We define $\bm{s}$ as an abstract state that includes all the information necessary to determine $\bm{a}$. The explicit definition of $\bm{s}$ will be provided later, but it does not affect the subsequent derivation.
The optimization objective is to align the twin policy with the user’s intention by minimizing the \emph{total variation} $\mathcal{D}_{\textnormal{TV}}$ of the two policies:
\begin{equation}
    \pi_{T}^* = \mathop{\textnormal{min}}_{\pi_{T}}~\mathcal{D}_{\textnormal{TV}}(\pi_T(\cdot|\bm{s}),\pi_I(\cdot|\bm{s}, \bar{\bm{g}})). \label{eq: ideal shared control}
\end{equation}
Unfortunately, the user's true goal $\bar{\bm{g}}$ is privileged information that we cannot measure. Instead, we propose to minimize a similar quantity in which possible goals are marginalized:
\begin{equation}
\mathop{\textnormal{min}}_{\pi_{T}}~\mathcal{D}_{\textnormal{KL}}(\pi_{T}(\cdot|\bm{s}),\pi_{I}(\cdot|\bm{s})), \label{eq: suboptimal shared control}
\end{equation}
where $\pi_{I}(\bm{a}|\bm{s}) = \sum_{\bm{g}}~p(\bm{g}) \pi_{I}(\bm{a}|\bm{s},\bm{g})$, and $\mathcal{D}_{\textnormal{KL}}(\cdot,\cdot)$ is the KL divergence. The relevance of this expression is supported by the following theorem, whose proof can be found in the appendix:

\noindent \emph{\textbf{Theorem 1}: Given $\pi_{I}(\bm{a}|\bm{s}) = \sum_{\bm{g}}~p(\bm{g}) \pi_{I}(\bm{a}|\bm{s}, \bm{g})$,  we can bound the
sub-optimality of any policy $\pi_{T}(\bm{a}|\bm{s})$ as: 
}
\begin{equation}
\begin{aligned}
    & \mathcal{D}_{\textnormal{TV}}(\pi_T(\cdot|\bm{s}), \pi_I(\cdot|\bm{s}, \bm{g})) \\
    &  \leq \sqrt{2\mathcal{D}_{\textnormal{KL}}(\pi_T(\cdot|\bm{s}),\pi_I(\cdot|\bm{s}))} + \mathcal{D}_{\textnormal{TV}}(\pi_I(\cdot|\bm{s}), \pi_I(\cdot|\bm{s}, \bm{g}))
\end{aligned}
\end{equation}
According to \emph{Theorem 1}, the total variation between the twin policy and the goal-conditioned intention user policy is bounded by the KL divergence between the twin policy and the unconditioned intention user policy, plus the total variation between the unconditioned and goal-conditioned intention user policy. Because $\mathcal{D}_{\textnormal{TV}}(\pi_I(\cdot|\bm{s}), \pi_I(\cdot|\bm{s}, \bm{g}))$ is independent of the twin policy $\pi_T(\bm{a}|\bm{s})$, we can approach $\pi_{I}(\bm{a}|\bm{s},\bar{\bm{g}})$ by training $\pi_T(\bm{a}|\bm{s})$ to imitate $\pi_I(\bm{a}|\bm{s})$.

Unfortunately, analytically solving Expression \ref{eq: suboptimal shared control} is impossible because $\pi_{I}$ cannot be explicitly obtained. 
Thus, we seek an empirical optimization goal by transforming Expression \ref{eq: suboptimal shared control} into maximizing the log-likelihood of the intended trajectories:
\begin{equation}
\mathop{\textnormal{max}}~\mathbb{E}_{\bm{\zeta}} [\textnormal{log}~p(\bm{s}_\trajectorySegment{2}{4}, \bm{a}_\trajectorySegment{1}{3}|\bm{s}_\trajectorySegment{0}{2}, \bm{a}_\trajectorySegment{0}{1} )],  \label{eq: substitute mle}
\end{equation}
where we represent the trajectory in terms of relative timesteps, and tabular subscripts compactly denote the following four segments of the trajectory: $\bm{*}_{-T+1:-1}$, $\bm{*}_0$, $\bm{*}_{1:H-1}$, and $\bm{*}_{H}$, respectively. For example, $\bm{a}_\trajectorySegment{0}{1}$ is equivalent to $\bm{a}_{-T+1:-1}$, and $\bm{s}_\trajectorySegment{2}{4}$ denotes $\bm{s}_{1:H}$. Our model aims to predict ``future'' trajectories from ``past'' trajectories. We use $t=0$ to represent the ``current'' timestep. The variable $\bm{s}_t$ is the state at $t$, incorporating the current dynamics and contextual information, and $\bm{a}_t$ is the velocity action at $t$. The future (intended) trajectory $\bm{\zeta} = (\bm{s}_\trajectorySegment{0}{4},\bm{a}_\trajectorySegment{0}{3})$ is sampled from the interaction between the user and the environment with failure samples filtered out, $(\bm{s}_\trajectorySegment{0}{2},\bm{a}_\trajectorySegment{0}{1})$ is the past trajectory and $(\bm{s}_\trajectorySegment{2}{4},\bm{a}_\trajectorySegment{1}{3})$ is the future trajectory. To further decompose Eq.~\ref{eq: substitute mle}, we make the following assumption:

\noindent \emph{\textbf{Assumption 1}: We assume that the user's action is conditionally independent of the past actions given the past states, and a state is conditionally independent of past states and actions given the state and action that directly precede it. For instance, $\bm{a}_0$ is only dependent on $\bm{s}_{\leq 0}$, and $\bm{s}_{0}$ is conditionally independent of $\bm{s}_{\leq-2}$ and $\bm{a}_{\leq-2}$ given $\bm{s}_{-1}$ and $\bm{a}_{-1}$. Fig.~\ref{fig: graphical model}~(a) expresses this assumption graphically.
}

\noindent \emph{\textbf{Remark}: This is a restrictive assumption, but it is less restrictive compared to prior works \citep{ezeh2017probabilistic,aigner2000modeling} which assumes the robot's action depends on the latest sensor and user signal.}

\noindent We introduce a latent state $\bm{h}_t$, which encodes a memory of past states $\bm{s}_{\leq t}$. Because the past states are observed and the future states are not, we compute the $\bm{h}$ instead of $\bm{s}$ for a future trajectory. Thus, by leveraging \emph{Assumption 1}, we compute $p(\bm{h}_\trajectorySegment{1}{4}, \bm{a}_\trajectorySegment{1}{3}|\bm{s}_\trajectorySegment{0}{2})$ instead of $p(\bm{s}_\trajectorySegment{2}{4}, \bm{a}_\trajectorySegment{1}{3}|\bm{s}_\trajectorySegment{0}{2},\bm{a}_\trajectorySegment{0}{1})$, which simplifies the problem.
Moreover, under \emph{Assumption~1} $\bm{a}_t$ is conditionally independent of $\bm{a}_{\leq t-1}$ and $\bm{s}_{\leq t}$ given $\bm{h}_t$, and we can decompose $p(\bm{h}_\trajectorySegment{1}{4}, \bm{a}_\trajectorySegment{1}{3}|\bm{s}_\trajectorySegment{0}{2})$ as:
\begin{equation}
\begin{aligned}
    & p(\bm{h}_\trajectorySegment{1}{4}, \bm{a}_\trajectorySegment{1}{3}|\bm{s}_\trajectorySegment{0}{2})\\ 
    & =   p(\bm{h}_0|\bm{s}_\trajectorySegment{0}{2}) \prod_{t=0}^{H-1}  \mathcolorbox{F2B1BB}{p(\bm{h}_{t+1}|\bm{h}_{t}, \bm{a}_{t})} \mathcolorbox{B6DDBB}{p(\bm{a}_{t}|\bm{h}_{t})},
    \label{eq: decomposed mle2}
\end{aligned}
\end{equation}
where $\mathcolorbox{F2B1BB}{p(\bm{h}_{t+1}|\bm{h}_{t}, \bm{a}_{t})}$ is the state transition distribution, and  $\mathcolorbox{B6DDBB}{p(\bm{a}_{t}|\bm{h}_{t})}$ is the action distribution. The graphical model of Eq.~\ref{eq: decomposed mle2} can be represented in Fig.~\ref{fig: graphical model}~(a).

\begin{figure*}[!t]
  \centering
  \includegraphics[width=\linewidth]{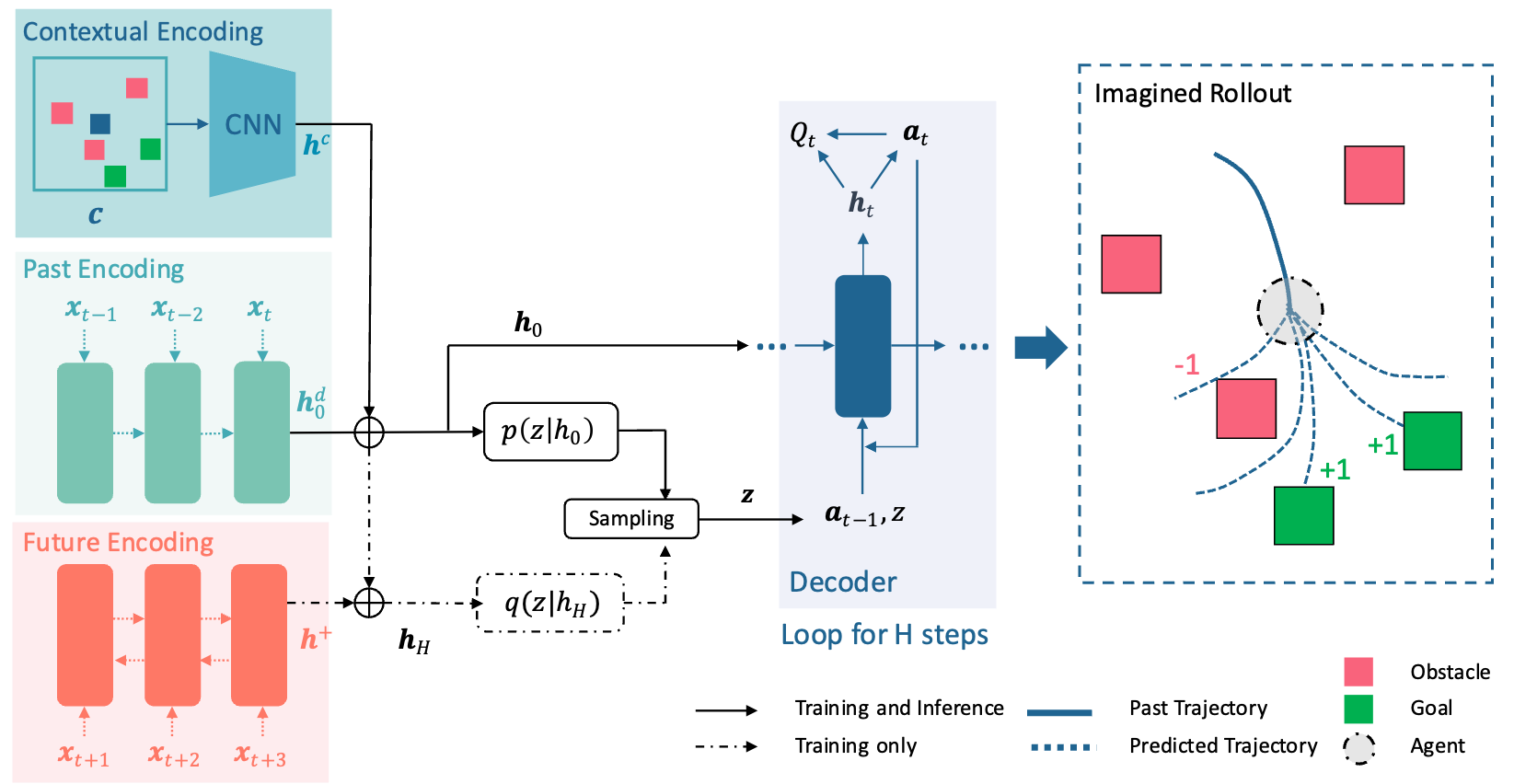}
  \caption{The architecture of the Robot Trajectron V2. The model follows a CVAE framework. RT-V2 encodes dynamics and contextual information to generate future trajectories (imagined rollouts) and receives rewards by interacting with local environments. See text for details.}
  \label{fig: rtv2}
\end{figure*}

To model the high multi-modality of the user's intent, as RT \citep{RT}, we mimic the CVAE framework \citep{cvae, trajectron} and introduce a latent variable $\bm{z}$, which obeys a Multinoulli distribution, to facilitate the encoding of a low-dimensional, multi-modal representation, as:
\begin{equation}
\begin{aligned}
    & p(\bm{h}_{0}|\bm{s}_\trajectorySegment{0}{2}) p(\bm{h}_\trajectorySegment{2}{4},\bm{a}_\trajectorySegment{1}{3}|\bm{h}_{0})\\
    & =   \sum_{\bm{z}} p(\bm{h}_{0}|\bm{s}_\trajectorySegment{0}{2}) p(\bm{h}_\trajectorySegment{2}{4}, \bm{a}_\trajectorySegment{1}{3}|\bm{h}_{0},\bm{z}) p(\bm{z}|\bm{h}_{0})\\
     & =    p(\bm{h}_{0}|\bm{s}_\trajectorySegment{0}{2}) \sum_{\bm{z}} p(\bm{z}|\bm{h}_{0})\\
     &~~~~~~~~~~~~~~\prod_{t=0}^{H-1}  \mathcolorbox{F2B1BB}{p(\bm{h}_{t+1}|\bm{h}_{t}, \bm{a}_{t}, \bm{z})} \mathcolorbox{B6DDBB}{p(\bm{a}_{t}|\bm{h}_{t}, \bm{z})}.
    \label{eq: decomposed mle with latent}
\end{aligned}
\end{equation}
The graphical model of Eq.~\ref{eq: decomposed mle with latent} can be represented in Fig.~\ref{fig: graphical model}~(b), which introduces $\bm{z}$ compared to Fig.~\ref{fig: graphical model}~(a). Thus, $\mathcolorbox{F2B1BB}{p(\bm{h}_{t+1}|\bm{h}_{t}, \bm{a}_{t})}$ and  $\mathcolorbox{B6DDBB}{p(\bm{a}_{t}|\bm{h}_{t})}$ in Eq.~\ref{eq: decomposed mle with latent} can be rewritten as $\mathcolorbox{F2B1BB}{p(\bm{h}_{t+1}|\bm{h}_{t}, \bm{a}_{t}, \bm{z})}$ and  $\mathcolorbox{B6DDBB}{p(\bm{a}_{t}|\bm{h}_{t}, \bm{z})}$, respectively.
This latent variable $\bm{z}$ denotes the ``maneuver class'' reflecting the probable maneuver directions for the user to execute \citep{trajectron,song2020pip}.
To train the model, we maximize the likelihood in Eq.~\ref{eq: decomposed mle2} by minimizing, per CVAE practice \citep{beta-vae,cvae}, the $\beta$-weighted evidence-based lower bound (ELBO) loss:
\begin{equation}
\begin{aligned}
     & \mathcal{L}_{\text{ELBO}} \\
     & =  ~ -\mathbb{E}_{\bm{z} \sim p(\bm{z}|\bm{h}_{H})}[\textnormal{log}~p(\bm{h}_\trajectorySegment{2}{4}, \bm{a}_\trajectorySegment{1}{3}|\bm{s}_\trajectorySegment{0}{2}, \bm{z})] \\
    &~~ + \beta D_{KL}(q(\bm{z}|\bm{h}_{H})||p(\bm{z}|\bm{h}_{0})) \\
    & =   - \mathbb{E}_{\bm{z} \sim p(\bm{z}|\bm{h}_{H})}[p(\bm{h}_{0}|\bm{s}_\trajectorySegment{2}{4}) \\ 
    & ~~~~~~~~~~~~~~~\sum_{t=0}^{H-1} \textnormal{log}~ p(\bm{h}_{t+1}|\bm{h}_{t}, \bm{a}_{t},\bm{z}) p(\bm{a}_{t}|\bm{h}_{t},\bm{z})] \\
    &~~ + \beta D_{KL}(q(\bm{z}|\bm{h}_{H})||p(\bm{z}|\bm{h}_{0})).
    \label{eq: ELBO}
\end{aligned}
\end{equation}
where the maneuver class posterior $q(\bm{z}|\bm{h}_{H})$ is introduced to guide the training of $p(\bm{z}|\bm{h}_{0})$ by minimizing their KL divergence.

The proposed method, named RT-V2, is implemented as follows (shown in Fig.~\ref{fig: rtv2}):
\begin{equation}
\begin{aligned}
    & \textnormal{Prior Latent Encoder} & \bm{h}_{0} \sim \delta(\bm{h} - h_{\theta}(\bm{s}_\trajectorySegment{0}{2})), \\
    & \textnormal{Posterior Latent Encoder} &\bm{h}_{H} \sim \delta(\bm{h} - h_{\kappa}(\bm{s}_\trajectorySegment{0}{4})), \\
    & \textnormal{Maneuver Class Prior} & \bm{z} \sim p_{\omega}(\bm{z}|\bm{h}_{0}), \\
    & \textnormal{Maneuver Class Posterior} & \bm{z} \sim q_{\eta}(\bm{z}|\bm{h}_{H}), \\
    & \textnormal{Actor Layer} & \bm{a}_t \sim p_{\phi}(\bm{a}_{t}|\bm{h}_{t}, \bm{z}), \\
    & \textnormal{Latent Dynamics} & \bm{h}_{t+1} \sim \delta(\bm{h} - t_{\psi}(\bm{h}_t, \bm{a}_t,\bm{z})) ,  
    \label{eq: components}
\end{aligned}
\end{equation}
where ${\omega}$, ${\eta}$ and ${\phi}$ denote the learnable parameters of the neural representation underlying probability distributions $q_{\omega}$, $p_{\eta}$, and $q_{\phi}$. ${\theta}$, ${\kappa}$ and ${\psi}$ are the learnable parameters of the prior encoder $h_{\theta}(\cdot)$, posterior encoder $h_{\kappa}(\cdot)$, and the latent dynamics $t_{\psi}(\cdot)$. We use the Dirac function $\delta(\cdot)$ because it is a static environment and the transition between states is deterministic. We will discuss the implementation details in the following subsections.

\subsection{Latent Encoder}
We use neural networks parameterized with $\theta = (\theta_l, \theta_c)$ to encode the past state trajectory $\bm{s}_\trajectorySegment{0}{2} = \{\bm{x}_{t}, \bm{c}_{t}\}_{t=-T+1}^{0}$, where $\bm{x}_t = [X_{t}, \dot{X}_{t}, \ddot{X}_{t}] \in \mathbb{R}^{6}$ is the dynamics information of the robot (i.e., position, velocity and acceleration) and $\bm{c}_t$ is the contextual information at timestep $t$. Since we work in a static environment, we can simplify the $\bm{s}_\trajectorySegment{0}{2}$ as $\{\bm{x}_\trajectorySegment{0}{2}, \bm{c}\}$. We use an LSTM as a past encoder to encode the past dynamics $\bm{x}_\trajectorySegment{0}{2}$, as $\bm{h}^{\textnormal{d}}_0 = \textnormal{LSTM}(\bm{x}_\trajectorySegment{0}{2};\theta_l)$. 
Then, we represent contextual information as a $H \times H$ local occupancy map with three channels $\bm{c} \in \mathbb{R}^{3 \times H \times H}$. The first channel represents the robot's position, for which we assign one to the center pixel. The second and third channels represent the position of the potential goals and obstacles, as:
\begin{equation}
\begin{aligned}
    &\bm{c}(1,x,y) = \bigvee_{m=1}^{M} \mathbb{I}((x,y)=\bm{g}_{m}), \\
    &\bm{c}(2,x,y) = \bigvee_{n=1}^{N} \mathbb{I}((x,y)=\bm{o}_{n}).
\end{aligned}
\end{equation}
We use convolutional layers to encode the contextual information as $\bm{h}^{\textnormal{c}} = \textnormal{Conv}(\bm{c};\theta_c)$. Thus, the latent past encoding is $\bm{h}_{0} = [\bm{h}^{\textnormal{d}}_0; \bm{h}^{\textnormal{c}}]$.
Besides, the posterior latent encoding is obtained by concatenating the latent future encoding, which is obtained as:
\begin{equation}
    \bm{h}^{+} = \textnormal{BiLSTM}(\bm{x}_\trajectorySegment{2}{4};\kappa),
\end{equation}
with $\bm{h}_{0}$, as $\tilde{\bm{h}}_{H}=[\bm{h}^{+};\bm{h}^{\textnormal{d}}_0; \bm{h}^{\textnormal{c}}]$. We use $\tilde{\bm{h}}_{H}$ as the posterior latent encoding in Eq. \ref{eq: components} instead of $\bm{h}_{H}=[\bm{h}^{\textnormal{d}}_{H};\bm{h}^{c}]$, where $\bm{h}^{\textnormal{d}}_{H}$ is obtained by feeding $\bm{x}_\trajectorySegment{0}{4}$ into the past encoder because it improves the training stability with same information encoded.

   
\subsection{Maneuver Class Sampling}
We model maneuver class prior $p_{\omega}$ and posterior $q_{\eta}$ as Bernoulli distributions whose parameters are generated with multi-layer perceptrons (MLPs), as:
\begin{align}
    & B_\omega = \textnormal{MLP}(\bm{h}_{0};\omega),\\
    & B_\eta = \textnormal{MLP}(\tilde{\bm{h}}_{H};\eta).
\end{align}
We sample $\bm{z}$ from $q_{\eta}$ during the training, while from $p_{\omega}$ during the inference.

\subsection{Decoding Future Trajectory}
When decoding the future trajectory, we use LSTM model to encode the latent dynamics, as:
\begin{equation}
    \bm{h}_{t+1} = t_{\psi}(\bm{h}_t, \bm{a}_t,\bm{z}) = \textnormal{LSTM}_t([\bm{h}_{t},\bm{a}_{t},\bm{z}];{\psi}).
\end{equation}
We model the actor layer $p_{\phi}$ with velocity-space Gaussian Mixture Models (GMMs) updated at each timestep. We denote the parameters of the GMMs at time $t$ with $G_t = \{(\bm{\mu}_t^{z}, \bm{\Sigma}_t^{z}, \alpha_t^{z})\}_{z=1}^Z$, where $Z$ is the number of maneuver classes (or the number of Gaussian components) and $\alpha_t^{z}=p(\bm{z}|\bm{h}_{0})$ (or $q(\bm{z}|\bm{h}_{H})$ during training).
The rest of the GMMs' parameters are modeled by MLP, as follows:
\begin{equation}
    \{(\bm{\mu}_t^{z}, \bm{\Sigma}_t^{z})\}_{z=1}^Z = \textnormal{MLP}(\bm{h}_{t};\phi).  \label{eq:velocitygmm}
\end{equation}
We predict the action at time $t$ via sampling, as $\bm{a}_{t} \sim \textnormal{GMMs}(G_t)$. 

\section{Imagined Rollout Reinforcement Learning} \label{irrl}
A crucial problem of imitation learning is causal confusion. The model is trained on successful trials that contain no failure cases. Thus, the model does not know how to behave when encountering an unsafe condition. Besides, the model is trained to imitate the behavior instead of learning through feedback from interacting with the true environment. Therefore, it is difficult for the model to capture a causal relation between the action taken and the contextual information, leading to a low goal achievement rate and a high collision rate. To address this limitation, we propose Imagined Rollout Reinforcement Learning, in which the model can be trained with rewards from imagined action rollout interacting with an approximate environment.
In our navigation task, we use an approximate dynamics model \citep{trajectron} to integrate the action trajectory $\bm{a}_\trajectorySegment{1}{3}$ to position trajectory $\xi = X_\trajectorySegment{2}{4}$, as:
\begin{equation}
    X_{t+1} =  X_{t} + \bm{a}_{t} \Delta t. \label{eq: dynamics}
\end{equation}

\noindent \emph{\textbf{Remark}: In many robotics tasks, approximate dynamics models are often obtainable and work well \citep{bhardwaj2022storm,lu2019model}. Besides, because this framework is based on model-based RL \citep{ha2018world,dreamer,williams2017information}, the dynamics model can be learned to obtain the imagined trajectory if an approximate dynamics model is not accessible.}

As a probabilistic model, RT-V2 can sample many imagined trajectories by sampling actions from the actor layers for each step and rolling them out through Eq.~\ref{eq: dynamics}. Then, since we have the local map, we can assign a reward $r$ to every imagined trajectory, as:
\begin{equation}
    r = \begin{cases}
    & 1,  \ \ \ \text{if $\xi$ reaches one of the goals} \\
    & -1, \ \text{if $\xi$ collides with one of the obstacles}\\
    & 0, \ \ \ \textnormal{otherwise},
     \end{cases} 
\end{equation}


\noindent We follow the Actor-Critic framework \citep{ac} to train RT-V2 with Q function $Q_{\varphi}(\bm{h}_t, \bm{a}_t)$ parametrized as MLPs. The Q-function parameters are trained to minimize the soft Bellman residual:
\begin{equation}
    \mathcal{L}_{Q} = \mathbb{E}_{\substack{\bm{a}_t \sim p(\bm{a}_t|\bm{h}_t) \\ \bm{h}_t \sim p(\bm{h}_{t+1}|\bm{h}_t, \bm{a}_t) }}[\frac{1}{2}(Q_{\varphi}(\bm{h}_t, \bm{a}_t)- \hat{Q}(\bm{h}_t, \bm{a}_t))], \label{eq: co}
\end{equation}
with
\begin{equation}
    \hat{Q}(\bm{h}_t, \bm{a}_t) = \textnormal{Clip}(r_{\textnormal{GAE}}(\bm{h}_t, \bm{a}_t) + \gamma Q_{\bar{\varphi}}(\bm{h}_{t+1}, \bm{a}_{t+1}), -1, 1),\label{eq: ao}
\end{equation}
where $\gamma$ is the discount factor, $\bar{\varphi}$ is an exponentially moving average of $\varphi$, and $r_{\textnormal{GAE}}$ is generalized advantage estimation \citep{gae}, which can reduce the variance of advantage estimation.
The actor can be trained to minimize the actor loss as:
\begin{equation}
    \mathcal{L}_{A} = \mathbb{E}_{\bm{a}_t \sim p(\bm{a}_t|\bm{h}_t)}[-r_{\textnormal{GAE}}(\bm{h}_t, \bm{a}_t) \cdot \textnormal{log}~p(\bm{a}_t|\bm{h}_t)].
\end{equation}
As Soft Actor-Critic (SAC) \citep{sac}, we introduce a maximum entropy term in the training goal to (i) encourage exploration in imagined rollouts and (ii) retain the multi-modal expressiveness of the RT-V2. 
Instead of maximizing, per SAC practice, the entropy estimated by the Monte-Carlo method, we maximize the entropy lower bound \citep{huber2008entropy}, which can be written as:
\begin{align}
    & H(\sum_{\bm{z}}q_{\eta}(\bm{z}|\bm{h}_{H})p_{\phi}(\cdot|\bm{h}_t, \bm{z})) \\ 
    & ~~~~~~\geq  - \sum_{\bm{z}} q_{\eta}(\bm{z}|\bm{h}_{H})~ \textnormal{log}\sum_{\bm{z}'} q_{\eta}(\bm{z}|\bm{h}_{H}) e_t(\bm{z},\bm{z}'), \nonumber \label{eq:entropy}\\
    & e_t(\bm{z},\bm{z}') = \mathcal{N}(\bm{\mu}^{z}_{t}|\bm{\mu}^{z'}_{t}, \bm{\Sigma}^{z}_t+\bm{\Sigma}^{z'}_t).
\end{align}
Notably, SAC normally computes the entropy of the RL policy $\pi(\bm{a}|\bm{s})$. However, since RT-V2 is a multi-modal model incorporating a latent variable $\bm{z}$, we marginalize over $\bm{z}$ in $p_{\phi}(\bm{a}_t|\bm{h}_t, \bm{z})$ to represent RT-V2's policy in Eq.~\ref{eq:entropy}.
Thus, we write the entropy loss as follows:
\begin{equation}
    \mathcal{L}_{\textnormal{ent}} = \sum_{t=0}^{H-1} \sum_{\bm{z}} q_{\eta}(\bm{z}|\bm{h}_{H})~ \textnormal{log}\sum_{\bm{z}'} q_{\eta}(\bm{z}|\bm{h}_{H}) e_t(\bm{z},\bm{z}').
\end{equation}

The total loss of RT-V2 can be written as:
\begin{equation}
    \mathcal{L} = \mathcal{L}_{\textnormal{ELBO}} + \mathcal{L}_{Q} + \mathcal{L}_{A} + \mathcal{L}_{\textnormal{ent}}. \label{eq: total loss}
\end{equation}

\section{Posterior Decision Making} \label{pdm}

Until now, we have focused exclusively on modeling the user's behavior. We have done so first via imitation learning (Sec.\ \ref{il4sc}), where the model is trained to imitate successful user trajectories, i.e., trajectories that reach a goal without colliding with any obstacle. Like all imitation learning models, the model of Sec.\ \ref{il4sc} is poised to struggle when placed in an unsafe situation (e.g., near an obstacle). To endow the model with robustness to those cases, we augmented it with imagined-rollout RL in Sec.\ \ref{irrl}, specifically training the model to avoid generating trajectories that collide with an obstacle. The results are a model that can robustly mimic the user's behavior, i.e., predict future trajectories based on historical motion.

In this section, we expand this behavior model to take user input into consideration, in addition to historical motion. We design a means of estimating the desired action conditioned on the current user command as the past trajectory. By training with the loss function described in Eq.~\ref{eq: total loss}, RT-V2 models the decision-making distribution $p(\bm{h}_\trajectorySegment{2}{4},\bm{a}_\trajectorySegment{1}{3}|\bm{s}_\trajectorySegment{0}{2})$, capturing the behavioral pattern of the intention policy, the ideal policy envisioned by the user. At the current timestep, we observe the user command $\bm{a}^{\text{u}}_{0}$, which is generated through a noisy input device. To achieve a better informed estimation of the intended action sequence $\{\bm{a}_t\}_{t=0}^{H-1}$, we incorporate the observed user command $\bm{a}^{\text{u}}_{0}$ and infer the posterior decision-making distribution as follows:
\begin{equation}
     p(\bm{h}_\trajectorySegment{2}{4}, \bm{a}_\trajectorySegment{1}{3}|\bm{s}_\trajectorySegment{0}{2}, \bm{a}^{\text{u}}_{0}). \label{eq: posterior decision making}·
\end{equation}
To calculate this distribution, we make the following assumption:


\noindent \emph{\textbf{Assumption 2}: We assume that the current user command $\bm{a}^{\text{u}}_{0}$ solely depends on the user's desired action $\bm{a}_0$.
}

\noindent \emph{\textbf{Remark}: Assumption 2 describes a process in which the user first generates a desired action $\bm{a}_0$ in their mind, and then inputs $\bm{a}_0^{\textnormal{u}}$ through the user interface.
}

\noindent Using \emph{Assumption 2}, we can represent the graphical model of Eq.~\ref{eq: posterior decision making} as shown in Fig.~\ref{fig: graphical model}~(c), in which an arrow from $\bm{a}_0$ to $\bm{a}_0^u$ is added compared to Fig.~\ref{fig: graphical model}~(b).
Since $\bm{a}_0$ cannot be observed, we introduce $\bm{a}_0^u$ to help posteriorly estimate $\bm{a}_0$. Following the same technique in Eq.~\ref{eq: decomposed mle with latent}, we can decompose Eq.~\ref{eq: posterior decision making} as:
\begin{equation}
\begin{aligned}
     & p(\bm{h}_\trajectorySegment{2}{4}, \bm{a}_\trajectorySegment{1}{3}|\bm{s}_\trajectorySegment{0}{2}, \bm{a}^{\text{u}}_{0}) \\
      &  = p(\bm{h}_{0}|\bm{s}_\trajectorySegment{0}{2})\\
      &~~ \sum_{\bm{z}}~p(\bm{h}_{1}|\bm{h}_{0}, \bm{a}_{0}, \bm{z}) \mathcolorbox{lightblue}{p(\bm{a}_{0}|\bm{h}_{0},\bm{z}, \bm{a}_0^{\textnormal{u}})}
     \mathcolorbox{lightorange}{p(\bm{z}|\bm{h}_{0}, \bm{a}_0^{\textnormal{u}})} \\
     & ~~\cdot \prod_{t=1}^{H-1} p(\bm{h}_{t+1}|\bm{h}_{t}, \bm{a}_{t}, \bm{z}) p(\bm{a}_{t}|\bm{h}_{t},\bm{z}).
    \label{eq: shared control goal}
\end{aligned}
\end{equation}
Compared to Eq.~\ref{eq: decomposed mle with latent}, the terms $p(\bm{z}|\bm{h}_{0})$ and $p(\bm{a}_0|\bm{h}_{0},\bm{z})$ become $\mathcolorbox{lightorange}{p(\bm{z}|\bm{h}_{0}, \bm{a}_0^{\textnormal{u}})}$ and $\mathcolorbox{lightblue}{p(\bm{a}_{0}|\bm{h}_{0},\bm{z}, \bm{a}_0^{\textnormal{u}})}$ in Eq.~\ref{eq: shared control goal}, respectively. The remaining terms in Eq.~\ref{eq: shared control goal} are identical to those in Eq.~\ref{eq: decomposed mle with latent}. To compute $p(\bm{h}_\trajectorySegment{2}{4}, \bm{a}_\trajectorySegment{1}{3}|\bm{s}_\trajectorySegment{0}{2}, \bm{a}_0^{\textnormal{u}})$, we must first calculate $\mathcolorbox{lightorange}{p(\bm{z}|\bm{h}_{0}, \bm{a}_0^{\textnormal{u}})}$ and $\mathcolorbox{lightblue}{p(\bm{a}_{0}|\bm{h}_{0},\bm{z}, \bm{a}_0^{\textnormal{u}})}$. We can define $\mathcolorbox{lightorange}{p(\bm{z}|\bm{h}_{0}, \bm{a}_0^{\textnormal{u}})}$ as:
\begin{equation}
     \mathcolorbox{lightorange}{p(\bm{z}|\bm{h}_{0}, \bm{a}_0^{\text{u}})} = \frac{p(\bm{a}_0^{\text{u}}| \bm{h}_{0}, \bm{z})}{p(\bm{a}_0^{\text{u}}|\bm{h}_{0})} p_{\omega}(\bm{z}|\bm{h}_{0}).
\end{equation}
Since each mode $\bm{z}$ corresponds to a component in action-GMMs, we can use these GMMs to evaluate $p(\bm{a}_0^{\text{u}}| \bm{h}_{0}, \bm{z})$ and $p(\bm{a}_0^{\text{u}}|\bm{h}_{0})$ as follows:
\begin{equation}
     p(\bm{a}_0^{\text{u}}|\bm{h}_{0}, \bm{z}) = \mathcal{N}(\bm{a}_0^{\text{u}}|\bm{\mu}^{z}_0, \bm{\Sigma}^{z}_0)
\end{equation}
\begin{equation}
     p(\bm{a}_0^{\text{u}}|\bm{h}_{0}) = \sum_{\bm{z}} p_{\omega}(\bm{z}|\bm{h}_{0}) p(\bm{a}_0^{\text{u}}|\bm{h}_{0}, \bm{z}).
\end{equation}

For $\mathcolorbox{lightblue}{p(\bm{a}_{0}|\bm{h}_{0},\bm{z}, \bm{a}_0^{\textnormal{u}})}$, we can express the posterior as the product of the prior and likelihood, as follows:
\begin{equation}
        \mathcolorbox{lightblue}{p(\bm{a}_0|\bm{h}_{0}, \bm{z}, \bm{a}_0^{\text{u}})} \propto p(\bm{a}_0|\bm{h}_{0}, \bm{z}) p(\bm{a}_0^{\text{u}}|\bm{a}_{0}),
\end{equation}
where $p(\bm{a}_0|\bm{h}_{0}, \bm{z})$ represents the prior action distribution as defined in Eq.~\ref{eq: decomposed mle with latent}, and $p(\bm{a}_0^{\text{u}}|\bm{a}_{0})$ accounts for the measurement uncertainty. 

\noindent \emph{\textbf{Assumption 3}: We assume that the intent measurement distribution $p(\bm{a}_0^{\text{u}}|\bm{a}_{0})$ is a Gaussian distribution, as:
\begin{equation}
     p(\bm{a}_0^{\text{u}}|\bm{a}_{0}) = \mathcal{N}(\bm{a}_0^{\text{u}}|\bm{a}_0, \bm{\Sigma}^{\text{sys}}),
\end{equation}
where $\bm{\Sigma}^{\text{sys}}$ represents the system variance of the user interface.
}

\noindent In practice, $\bm{\Sigma}^{\text{sys}}$ can either be estimated from the system or manually adjusted as a hyperparameter. Since both $p(\bm{a}_0|\bm{h}_{0}, \bm{z})$ and $p(\bm{a}_0^{\text{u}}|\bm{a}_{0})$ are Gaussian, $\mathcolorbox{lightblue}{p(\bm{a}_0|\bm{h}_{0}, \bm{z}, \bm{a}_0^{\text{u}})}$ can be rewritten as:
\begin{align}
    & \mathcolorbox{lightblue}{p(\bm{a}_0|\bm{h}_{0}, \bm{z}, \bm{a}_0^{\text{u}})} = \mathcal{N}(\bm{a}_0|\tilde{\bm{\mu}}_0^z, \tilde{\bm{\Sigma}}_0^z),\\
    & \bm{K} = \bm{\Sigma}_0^z(\bm{\Sigma}_0^z+\bm{\Sigma}^{\text{sys}})^{-1},\\
    & \tilde{\bm{\mu}}_0^z = \bm{\mu}_0^z + \bm{K}(\bm{a}_0^{\textnormal{u}}-\bm{\mu}_0^z), \\
    & \tilde{\bm{\Sigma}}_0^z = (\bm{I}-\bm{K})\bm{\Sigma}_0^z.
\end{align}
With $\mathcolorbox{lightorange}{p(\bm{z}|\bm{h}_{0}, \bm{a}_0^{\textnormal{u}})}$ and $\mathcolorbox{lightblue}{p(\bm{a}_{0}|\bm{h}_{0},\bm{z}, \bm{a}_0^{\textnormal{u}})}$, we can derive the posterior distribution $p(\bm{h}_\trajectorySegment{2}{4}, \bm{a}_\trajectorySegment{1}{3}|\bm{s}_\trajectorySegment{0}{2}, \bm{a}_0^{\textnormal{u}})$. This distribution yields a better-informed estimate of the intended action sequence $\{\bm{a}_t\}_{t=0}^{H-1}$, incorporating the current user's command $\bm{a}^{\textnormal{u}}_0$.

\section{Sampling-based Control with Constraint} \label{scwc}
A RL agent, especially in imitation learning, is not always guaranteed to act safely, often violating the robot’s constraints. For example, in navigation tasks, constraints include collision avoidance and adherence to maximum velocity limits. In robot manipulation, constraints may involve respecting joint limits and avoiding self-collisions. This challenge becomes even more pronounced when the agent is tasked with predicting the user’s intent, as users may not be fully aware of the robot's physical constraints. Thus, we aim to sample the action sequence that maximizes Expression~\ref{eq: posterior decision making} without violating the constraints. Specifically, we propose the safety likelihood function defined as:
\begin{equation}
\begin{aligned}
     & \tilde{p}_{\textnormal{safe}}(\bm{s}_\trajectorySegment{2}{4}, \bm{a}_\trajectorySegment{1}{3}|\bm{s}_\trajectorySegment{0}{2}, \bm{a}^{\textnormal{u}}_0) \\
    & =   p(\bm{s}_\trajectorySegment{2}{4}, \bm{a}_\trajectorySegment{1}{3}|\bm{s}_\trajectorySegment{0}{2}, \bm{a}^{\textnormal{u}}_0) \tilde{p}_{\textnormal{con}}(\bm{s}_\trajectorySegment{2}{4}). \label{eq: safe mle}
\end{aligned}
\end{equation}
Here, $\tilde{p}_{\textnormal{safe}}$ denotes an unnormalized distribution. Since our objective is to take its argmax, normalization is unnecessary.
The term $\tilde{p}_{\textnormal{con}}(\bm{s}_\trajectorySegment{2}{4})$ represents the unnormalized constraint function, which can be written as:
\begin{equation}
    \tilde{p}_{\textnormal{con}}(\bm{s}_{T+1:T+H}) = \begin{cases}
    & 0,  \ \text{if the constraint is violated,} \\
    & 1, \ \text{otherwise}.
     \end{cases}
     \label{eq: constraints}
\end{equation} 
Maximizing the safety likelihood in Eq.~\ref{eq: safe mle} simultaneously increases the likelihood of the user’s desired action sequence while reducing the risk of constraint violations.

\noindent \emph{\textbf{Remark}:
Constraints can also be represented as soft constraints, such as manipulability \citep{haviland2020purely}, which can be expressed as $\tilde{p}_{\textnormal{con}}(\bm{s}\trajectorySegment{2}{4}) \propto \exp(-\textnormal{cost}(\bm{s}\trajectorySegment{2}{4}))$. Multiple constraints can be combined multiplicatively, e.g., $\tilde{p}_{\textnormal{con}} = \tilde{p}_{\textnormal{con1}} \cdot \tilde{p}_{\textnormal{con2}} \cdot \tilde{p}_{\textnormal{con3}} \dots$.}

In this work, we use obstacle collision avoidance as a constraint. However, maximizing Eq.~\ref{eq: safe mle} is a highly multi-modal and non-convex optimization problem, making it challenging to calculate the result analytically. Inspired by Model Predictive Path Integral (MPPI) \citep{williams2017information}, we aim to utilize a sampling-based method to maximize Expression~\ref{eq: safe mle} by sampling trajectories to iteratively improve solutions.
We begin by defining the free energy of the system as follows:
\begin{align}
   & \mathcal{F}(\bm{A}) =- \lambda \mathop{\textnormal{log}} \mathbb{E}_{P}[\textnormal{exp}(-\frac{1}{\lambda}S(\bm{A}))],\\
   & S(\bm{A}) = \log \min(\tilde{p}_{\textnormal{safe}}(\bm{A}), \epsilon\big),
\end{align}
where $\bm{A} = \bm{a}_\trajectorySegment{1}{3}$, $\epsilon > 0$ is a small constant introduced to avoid numerical issues in the logarithm, and $P$ is a probability distribution over the actions of the uncontrolled system, written as:
\begin{equation}
     P(\bm{A}) = \prod_{t=0}^{H-1} \mathcal{N}(\bm{a}_t|\bm{0}, \bm{\Sigma}),
\end{equation}
Free energy represents the portion of a system's energy available to perform work. Our objective is to minimize the free energy, which corresponds to identifying the action trajectory that maximizes the safety likelihood as described in Expression~\ref{eq: safe mle}. To achieve this, we utilize a proposal action distribution $Q$ and express the free energy in terms of $Q$ by reformulating the expectation with respect to $Q$, as:
\begin{equation}
\begin{aligned}
    & \mathcal{F}(\bm{A})  =   - \lambda \textnormal{log}~( \mathbb{E}_{Q}[\frac{P(\bm{A})}{Q(\bm{A})}\textnormal{exp}(-\frac{1}{\lambda}S(\bm{A}))])\\
    & \leq -  \mathbb{E}_{Q}[\lambda \textnormal{log}~(\frac{P(\bm{A})}{Q(\bm{A})} ) - S(\bm{A})] ~~\textnormal{(Jensen's inequality)} \\
    & =   \mathbb{E}_{Q}[S(\bm{A})] + \lambda \mathcal{D}_{\textnormal{KL}}(Q||P), \label{eq: free energy lb}
\end{aligned}
\end{equation}
where $Q$ can be written as:
\begin{equation}
    Q(\bm{A}) = \prod_{t=0}^{H-1} \mathcal{N}(\bm{a}_t|\bm{u}_t, \bm{\Omega}_t).
\end{equation}
where $\bm{U}=\bm{u}_\trajectorySegment{1}{3}$ is the mean action sequence of $Q$.
Inequality~\ref{eq: free energy lb} demonstrates that free energy serves as a lower bound for the expected cost under the proposal distribution, in addition to the control cost represented by the KL divergence. Consequently, identifying a control distribution that achieves this lower bound minimizes both the expected cost and the control cost. Importantly, when $\frac{P(\bm{A})}{Q(\bm{A})} \propto \frac{1}{\textnormal{exp}(-\frac{1}{\lambda}S(\bm{A}))}$, the inequality~\ref{eq: free energy lb} becomes an equality. In this case, we can define the optimal control distribution $Q^*$ as:
\begin{equation}
    Q^*(\bm{A}) = \frac{\textnormal{exp}(-\frac{1}{\lambda}S(\bm{A}))}{\mathbb{E}_{P}[\textnormal{exp}(-\frac{1}{\lambda}S(\bm{A}))]}P(\bm{A}). 
\end{equation}
If we substitute $Q^*(\bm{A})$ for $Q(\bm{A})$ in Inequality~\ref{eq: free energy lb}, the lower bound can be achieved. 
Thus, the problem can be reformulated from minimizing the expected cost and control cost to minimizing the divergence between the action proposal distribution and the optimal control distribution.
Thus, the optimal actions sequence $\bm{U}^*=\bm{u}_\trajectorySegment{1}{3}^*$ can be obtained by:
\begin{equation}
\begin{aligned}
     & \bm{U}^*  = \mathop{\textnormal{argmin}}_{\bm{U}} \mathcal{D}_{\textnormal{KL}}(Q^*||Q)\\
     & = \mathop{\textnormal{argmin}}_{\bm{U}} \int Q^*(\bm{A})\mathop{\textnormal{log}}\frac{Q^*(\bm{A})}{Q(\bm{A})} \mathrm{d}\bm{A} \\
     & = \mathop{\textnormal{argmin}}_{\bm{U}} \int Q^*(\bm{A})\mathop{\textnormal{log}}(\frac{Q^*(\bm{A})}{P(\bm{A})} \frac{P(\bm{A})}{Q(\bm{A})}) \mathrm{d}\bm{A}\\
     & = \mathop{\textnormal{argmin}}_{\bm{U}} \int \underbrace{Q^*(\bm{A})\mathop{\textnormal{log}}(\frac{Q^*(\bm{A})}{P(\bm{A})} )}_{\text{Independent of} ~\bm{U}} - Q^*(\bm{A})\mathop{\textnormal{log}}(\frac{P(\bm{A})}{Q(\bm{A})}) \mathrm{d}\bm{A}\\
     & = \mathop{\textnormal{argmax}}_{\bm{U}} \int Q^*(\bm{A})\textnormal{log}~\frac{Q(\bm{A})}{P(\bm{A})} \mathrm{d}\bm{A}. \label{eq: Q* KL}
\end{aligned}
\end{equation}
It can be proved from Eq.~\ref{eq: Q* KL} that the optimal action at time $t$ is the expected input under the optimal distribution:
\begin{equation}
    \bm{u}_t^* = \int~Q^*(\bm{A})\bm{a}_t \mathrm{d}\bm{A}. \label{eq: optimal action}
\end{equation}
Since $Q^*$ is complex, high-dimensional, and intractable, direct sampling becomes challenging, making it impossible to analytically calculate the expectation in Eq.~\ref{eq: optimal action}. To address this, we use importance sampling to estimate the expectation in Eq.~\ref{eq: optimal action} by drawing samples from the proposal distribution $Q$, which is computationally convenient to sample from.
We use the actor layer $p_{\phi}(\bm{a}_{t}|\bm{h}_{t}, \bm{z}) = \mathcal{N}(\bm{a}_t|\bm{\mu}_t^z, \bm{\Sigma}_t^z)$ for mode $\bm{z}$ in Eq.~\ref{eq: components} to initialize $Q$ as $Q^{z}$, 
\begin{equation}
      Q^{z}(\bm{A}) = \prod_{t=0}^{H-1} p_{\phi}(\bm{a}_t|\bm{h}_t,\bm{z}) 
      =  \prod_{t=0}^{H-1} \mathcal{N}(\bm{a}_t|\bm{u}_t^z, \bm{\Omega}_t^z),
\end{equation}
where $\bm{u}_t^z = \bm{\mu}_t^z$ and $\bm{\Omega}_t^z = \bm{\Sigma}_t^z$. Then $\bm{u}_t^{z,*}$ can be estimated from the proposal action distribution $Q^z$ via importance sampling:
\begin{equation}
        \bm{u}_t^{z,*} =  \int~\frac{Q^*(\bm{A})}{Q^{z}(\bm{A})}Q^{z}(\bm{A})\bm{a}_t \mathrm{d}\bm{A}
        =  \mathbb{E}_{Q^z}[w^z(\bm{A})\bm{a}_t],
\end{equation}
with the importance sampling weight $w^z(\bm{A})$ as:
\begin{equation}
\begin{aligned}
    & w^z(\bm{A}) =  \frac{Q^*(\bm{A})}{Q^{z}(\bm{A})}
    =  \frac{Q^*(\bm{A})}{P(\bm{A})}\frac{P(\bm{A})}{Q^{z}(\bm{A})}\\
    & =   \textnormal{exp}(-\frac{1}{\lambda}S(\bm{A})+\sum_{t=0}^{H-1} (\bm{u}^{z}_t)^T(\bm{\Omega}^z_t)^{-1}(\bm{a}_t - \frac{1}{2}\bm{u}^{z}_t)).
\end{aligned}
\end{equation}
Given $N$ samples $\{\bm{A}^n\}_{n=1}^N$ drawn from $Q^{z}$, we can have the iterative update law of $Q^z$:
\begin{equation}
    \bm{u}_t^{z,i+1} = \bm{u}_t^{z,i} + \sum_{n=1}^N w^z(\bm{A}^n)(\bm{a}_t^{n} - \bm{u}_t^{z,i}).
\end{equation}
We represent the result of this optimization $\bm{u}_\trajectorySegment{1}{3}^{z,\infty}$ for the proposal action distribution $Q^{z}$ as $\bm{U}^{z, *}$. We solve $\bm{U}^{z, *}$ for all $\bm{z}$, and the final optimal action sequence $\bm{U}^*$ can be calculated as:
\begin{equation}
    \bm{U}^* = \mathop\textnormal{argmax}_{\bm{U}^{z, *}}~ \tilde{p}_{\textnormal{safe}}(\bm{U}^{z, *}).
\end{equation}
By comparing the results from multiple proposal distributions and selecting the minimum, we can address the multi-modal and non-convex optimization problem posed by Expression~\ref{eq: safe mle}. At each timestep, we compute only the current step action $\bm{a}^*_0$. In the subsequent step, Expression~\ref{eq: safe mle} is re-evaluated to determine the next action. This process is iteratively repeated.

\noindent \emph{\textbf{Remark}: This work specifically addresses scenarios where the action space is identical to the control input space. However, the proposed approach has the potential to be extended to scenarios where the action space differs from the control input space, such as 6-DoF robot manipulation with users controlling the end-effector \citep{RT}. In such cases, sampling would need to occur in the control (joint) space, with evaluations carried out in the action space. Addressing multi-modal optimization in these conditions is a challenging task. Nonetheless, advancements in Model Predictive Path Integral (MPPI) methods \citep{honda2023stein, rastgar2024priest} offer promising solutions, paving the way to extend this framework to various tasks in the future.}

\begin{table*}[!t]
  \centering
  \caption{The performance of trajectory prediction in Traj1M and navigation in the simulation environment. ``-'' denotes that this metric does not apply to this method.}
  \resizebox{\linewidth}{!}{
  \begin{tabular}{c|cccccc|cc|ccc}
    \toprule
    Method & Map  & AO. & CO. & Clip Q & Reinforce & MaxEnt &  ADE (mm) & FDE (mm) & SR (\%) & CR (\%) & OOT (\%) \\
    \midrule
    RT &  & & & & & &  226.42 & 372.51 & - & - & - \\
    \midrule
    RT-V2  & \Checkmark & \Checkmark & \Checkmark & \Checkmark & GAE & LB & 158.61 & 209.15 & 93 & 6 & 1 \\
    RT-V2 w. Constraint  & \Checkmark & \Checkmark & \Checkmark & \Checkmark & GAE & LB & - & - & 100 & 0 & 0 \\
    \midrule
    \multicolumn{12}{c}{Ablation Studies}\\
    \midrule
    Pure IL& \Checkmark  &  &  &  & && 181.19 & 260.72 & 28 & 30 & 32 \\
     + AO. & \Checkmark  & \Checkmark & &  & Return & & 163.83 & 219.21 & 80 & 13 & 6 \\
     + CO.& \Checkmark  & \Checkmark & \Checkmark &  & Advantage & & 161.83 & 219.21 & 85 & 8 & 9 \\
     + GAE& \Checkmark  & \Checkmark & \Checkmark &  & GAE & & 171.33 & 240.23 & 87 & 8 & 5 \\
     + Clip Q& \Checkmark  & \Checkmark & \Checkmark & \Checkmark & GAE & & 161.66 & 219.18 & 89 & 9 & 2 \\
     + MaxEnt (MC)& \Checkmark  & \Checkmark & \Checkmark & \Checkmark & GAE & MC & 161.44 & 217.36 & 77 & 20 & 3 \\
    \bottomrule
  \end{tabular}
  }
  \label{tab:Sim}
\end{table*}

\section{Experiments}
We conducted comprehensive experiments to evaluate the proposed RT-V2 method. Our experiments can be categorized into 3 groups: \emph{Offline-Exp}, \emph{Human-Exp}, and \emph{BMI-Exp}.
\begin{itemize}
    \item \emph{Offline-Exp}: Using simulated data, we first assessed the RT-V2's prior model for its trajectory prediction capabilities, i.e., its ability to derive intent from historical trajectory data. In addition, we evaluated the model's ability to function as a navigation controller, in a simulated environment.
    \item \emph{Human-Exp}: This set of experiments involved a shared autonomy setup with human participants using a keyboard interface in a simulated environment. The primary goal was to test RT-V2's effectiveness in assisting users with obstacle avoidance and achieving desired goals.
    \item \emph{BMI-Exp}:The generalizability of RT-V2 across different interfaces was demonstrated through two key experiments where user input was collected from brain-machine interfaces (BMIs) surgically implanted in our laboratory's subject monkeys. First, we conducted an offline trajectory prediction experiment, similar to \emph{Offline-Exp}. We then tested RT-V2 in a shared autonomy setup, mirroring our \emph{Human-Exp}.
\end{itemize}

\begin{figure}[!t]
  \centering
  \includegraphics[width=\linewidth]{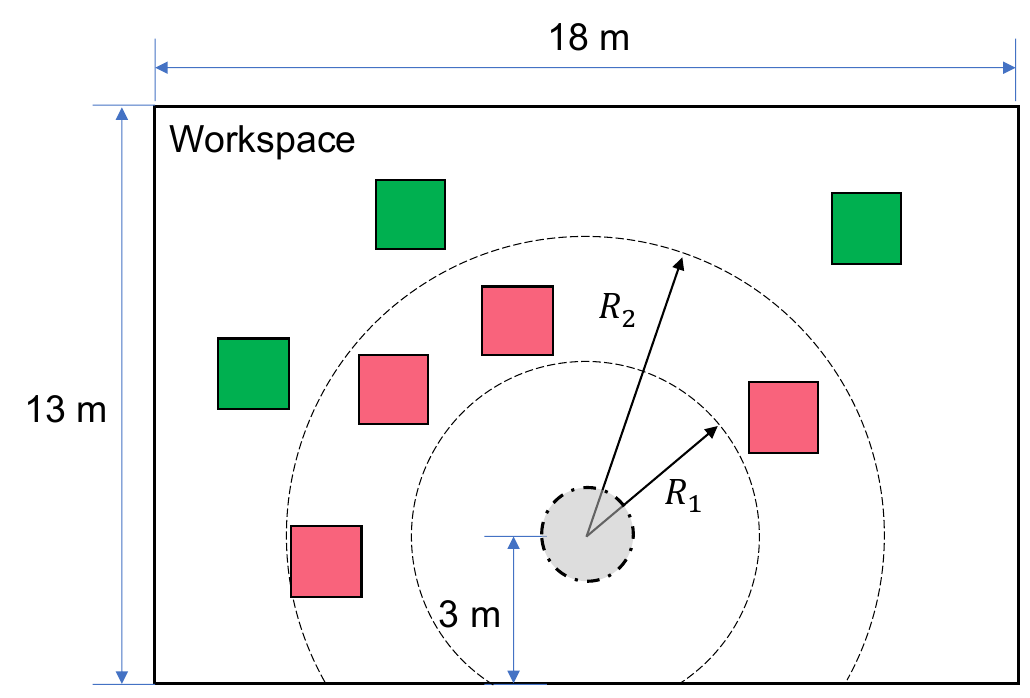}
  \caption{The setup of the simulation workspace. The gray circle with a dash border line denotes the initial position of the virtual robot. The pink boxes denote obstacles. The green boxes denote goals.}
  \label{fig: sim setup}
\end{figure}

\begin{figure*}[!t]
  \centering
  \includegraphics[width=\linewidth]{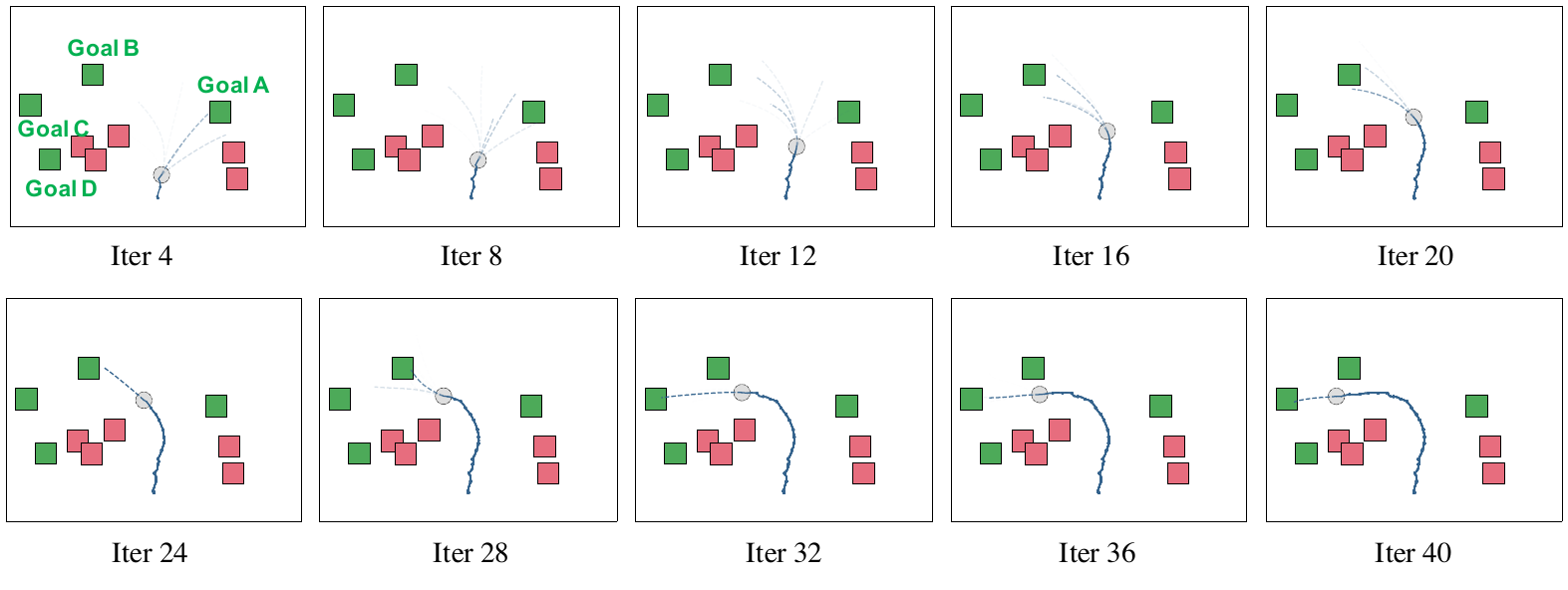}
  \caption{Visualization of RT-V2's generated trajectories. We visualize the trajectory of all the action-GMMs' modes every 4 iterations for better visibility. The gray circle with the dash border line is the robot. The blue line denotes the past trajectory, while the blue dash lines denote the predicted trajectories. The opacity of the predicted trajectories denotes the value of $q_{\omega}({\bm{z}|\bm{h}_{0}})$. Pink boxes denote obstacles while green boxes denote potential goals.}
  \label{fig: trajectory prediction}
\end{figure*}


\begin{figure*} [t!]
\centering
    \resizebox{\linewidth}{!}{
    \subfloat[\label{nav_a}]{
    \includegraphics[scale=0.17]{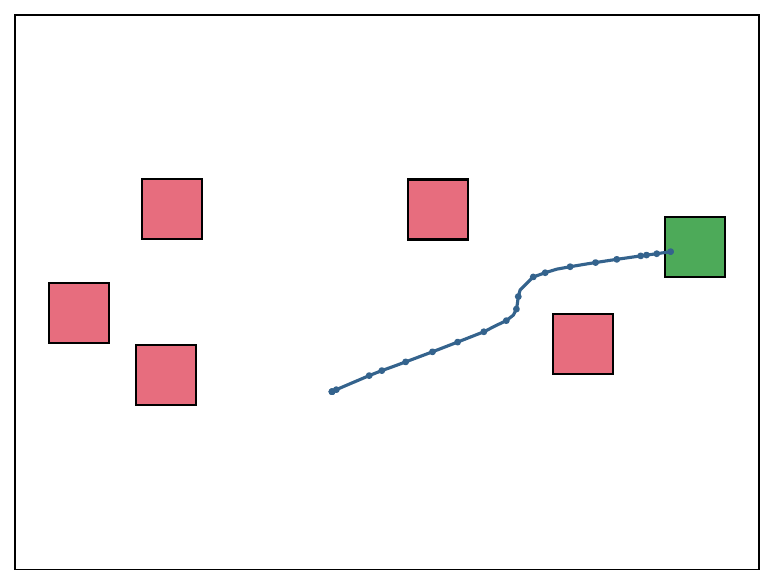}}
    \subfloat[\label{nav_b}]{
    \includegraphics[scale=0.17]{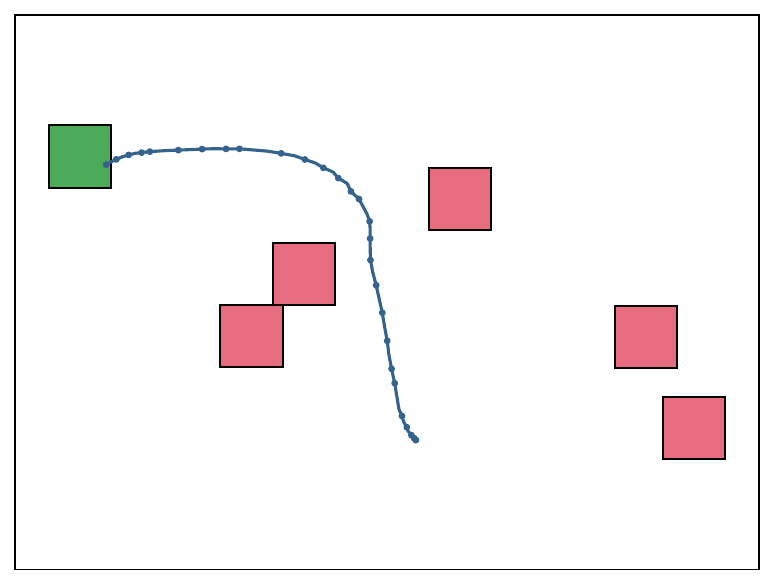}}
    \subfloat[\label{nav_c}]{
    \includegraphics[scale=0.17]{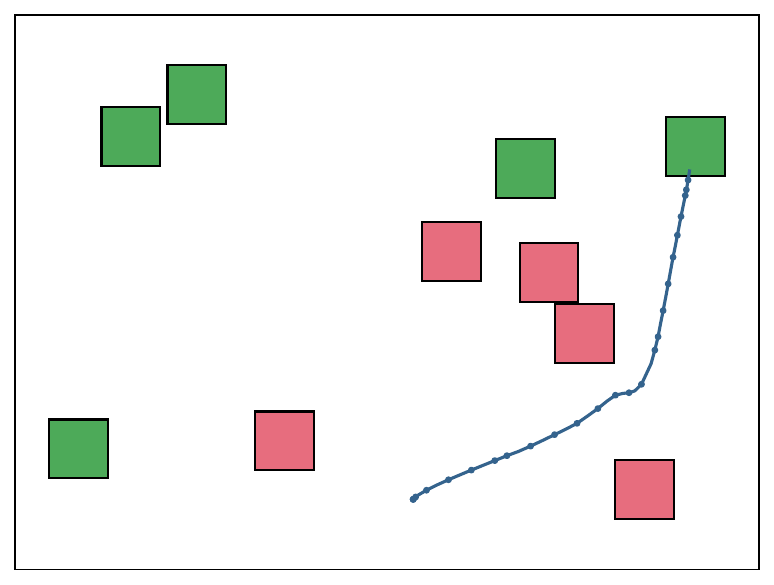}}
    \subfloat[\label{nav_d}]{
    \includegraphics[scale=0.17]{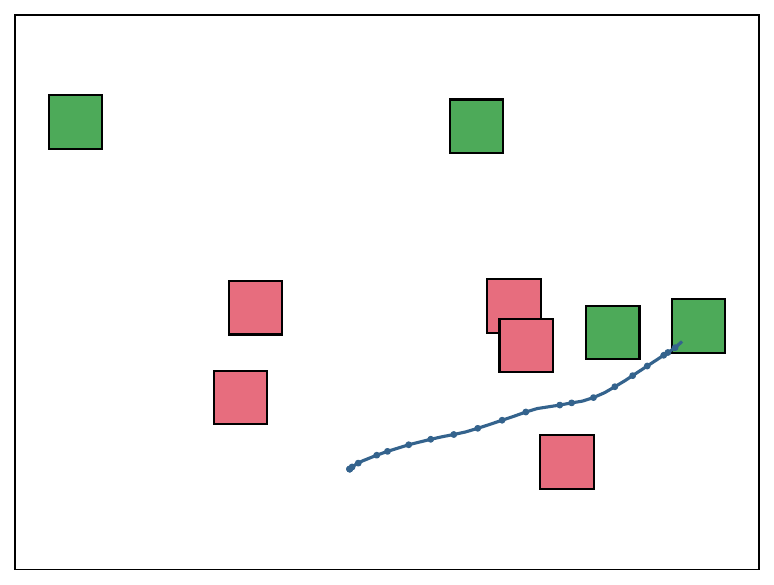}}
    \subfloat[\label{nav_e}]{
    \includegraphics[scale=0.17]{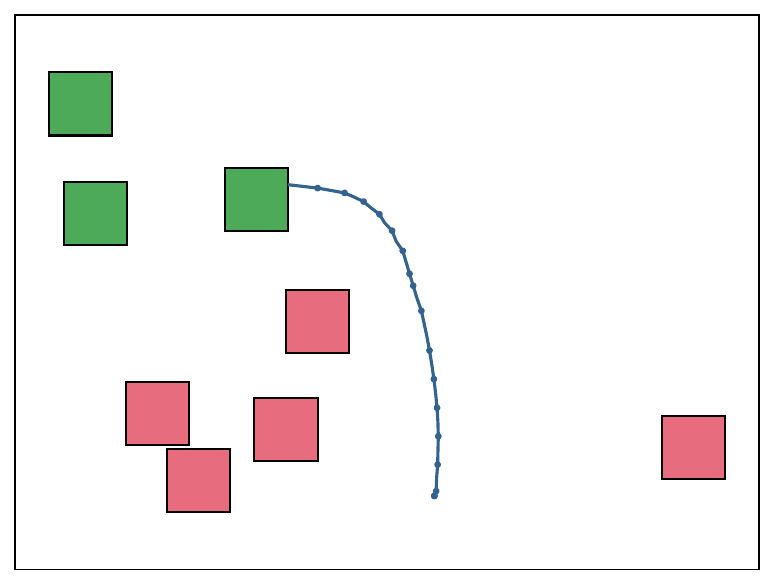}}
    \subfloat[\label{nav_f}]{
    \includegraphics[scale=0.17]{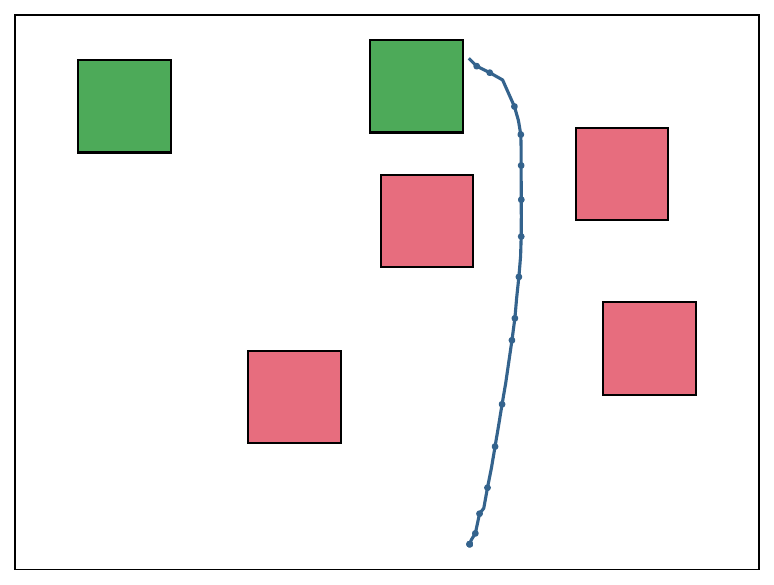}}
    }
    \caption{Visualization of RT-V2's navigation trajectories. The blue lines denote the executed trajectories. Pink boxes denote obstacles, while green boxes denote potential goals. All the sizes of goals and obstacles are the same. The different sizes in different figures only denote the different scales.}
\label{fig: viz_navi}
\end{figure*}

\subsection{Offline-Exp1: Trajectory Prediction}
In this experiment, we evaluate the performance of RT-V2 on a simulation dataset. Consistent with established practices in trajectory prediction \citep{mid,trajectron}, we use the widely-adopted evaluation metrics \emph{Average Displacement Error (ADE)} and \emph{Final Displacement Error (FDE)}:  
\begin{itemize}
    \item \textbf{ADE}: Measures the average distance between all ground truth and predicted positions across the trajectory.
    \item \textbf{FDE}: Measures the distance between the endpoints of the ground truth and predicted trajectories.
\end{itemize}
To assess the model's ability to capture the multi-modality of user intent, we compute these metrics based on the \emph{best-of-20} trajectories sampled from the model's predictions.

\noindent \textbf{Data collection and training setting.} 
To train the model, we collect a large-scale dataset in simulation. The experimental setup is illustrated in Fig.~\ref{fig: sim setup}. An omnidirectional virtual robot, represented as a sphere with a radius of $0.5~\text{m}$, is initially positioned at the origin $(0,0)$ in a $13 \times 18~\text{m}^2$ workspace. A random number of cubic obstacles with the length of $1.3~\text{m}$ are placed outside a distance $R_1$ from the origin and within a distance $R_2$, while a random number of goals are distributed outside $R_2$ and within the workspace boundaries. The robot is required to avoid obstacles and navigate to a randomly indicated goal.
Probabilistic Roadmaps (PRM) \citep{kavraki1996probabilistic} are used to plan a trajectory, which is then tracked using a noisy proportional controller. During this process, we record the positions of the conducted trajectory, the goals, and the obstacles. In total, we generate 1 million trajectories, of which 900,000 are used for training and 100,000 for testing. This dataset is referred to as \emph{Traj1M}. The model is trained with Adam optimizer, with a learning rate of 0.0001 and a batch size of 256. The models are trained on a single RTX 4070Ti GPU.

\noindent \textbf{Performance in the Traj1M dataset.}
The results are presented in Table~\ref{tab:Sim}. RT-V2 achieves an ADE of 158.61~mm and a FDE of 209.15~mm, significantly outperforming RT, which records an ADE of 226.42~mm and an FDE of 372.51~mm. The reduced error demonstrates the effectiveness of RT-V2's contextual encoding in improving the accuracy of future trajectory predictions.

\noindent \textbf{Visualization of trajectory prediction.}
Fig.~\ref{fig: trajectory prediction} presents an example of trajectory prediction using RT-V2. The trajectories of all modes from the action-GMMs are visualized at intervals of 4 iterations. The blue line represents the past trajectory, while the blue dash lines indicate predicted trajectories. The opacity of the predicted trajectories corresponds to the value of $q_{\omega}({\bm{z}|\bm{h}_{0}})$. Pink boxes depict obstacles, and green boxes indicate potential goals.
At the start of motion, RT-V2 predictions show significant uncertainty. Between Iter 4 and 12, multiple distinct trajectories are generated, with the most likely trajectory heading towards Goal A, showcasing the model's ability to represent multi-modal user intent. From Iter 16 to 24, the robot changes direction towards the left, and RT-V2 adapts its predictions to focus on Goal B, reflecting a shift in dynamics perception. Between Iter 28 and 40, RT-V2 recognizes that Goal B is not the intended target based on past dynamics and adjusts its predictions toward Goal C, progressively gaining confidence as alternative modes become transparent. 
This visualization demonstrates the multi-modal modeling capability and adaptability of RT-V2 in predicting user intent.

\subsection{Offline-Exp2: Navigation in Simulation Environment}
In this experiment, we evaluate the navigation ability of RT-V2 in a simulated environment identical to the one used for data collection in \emph{Offline-Exp1}. Success is defined as the robot reaching one of the goals, while failure occurs if the robot either collides with obstacles or fails to reach a goal within the time limit. The experiments are conducted over 100 rounds with fixed random seeds, and three metrics are recorded: success rate (SR), collision rate (CR), and out-of-time rate (OOT).
During each round, RT-V2 selects its optimal action using the following equations:  
$ \bar{\bm{z}} = \mathop{\textnormal{argmax}}_{\bm{z}} q_{\omega}(\bm{z}|\bm{h}_{0}),
\bar{\bm{a}}_t = \mathop{\textnormal{argmax}}_{\bm{a}_t} p_{\phi}(\bm{a}_t|\bm{h}_t,\bar{\bm{z}})$.
The navigation results are presented in Table~\ref{tab:Sim}. RT-V2 achieves a success rate (SR) of 93\%, a collision rate (CR) of 6\%, and an out-of-time rate (OOT) of 1\%, demonstrating its effective planning capability. Furthermore, by incorporating the constraints described in Section VI, RT-V2 achieves a perfect performance with 100\% SR and 0\% CR.

\noindent \textbf{Ablation studies}
Table~\ref{tab:Sim} illustrates the evolution from RT to RT-V2 through ablation studies. The table uses the following notation:  
\begin{itemize}
    \item \textbf{Map}: Denotes encoding the contextual map with a CNN. 
    \item \textbf{AO.}: Denotes actor optimization following the equation $\phi \leftarrow \phi + \alpha R \nabla \textnormal{log} p_{\phi}(\bm{a}_t|\bm{h}_t)$. 
    \item \textbf{CO.}: Denotes critic optimization in Eq.~\ref{eq: co}.  
    \item \textbf{Reinforce}: Specifies which $R$ is used in actor optimization.  
    \item \textbf{Clip Q}: Denotes whether $\hat{Q}(\bm{h}_t,\bm{a}_t)$ is clipped.  
    \item \textbf{MaxEnt}: Denotes which estimated action entropy is maximized during the training.  
\end{itemize}
The results demonstrate that pure imitation learning (``Pure IL'') achieves a low success rate (SR) of 28\%, highlighting the difficulty of capturing the correlation between future actions over the next $H$ steps and historical contextual information without reinforcement signals. Adding actor optimization with discounted returns (``+ AO.'') significantly improves SR from 28\% to 80\%. Introducing critic optimization (``+ CO.'') further increases SR to 85\%, addressing the reward sparsity problem. Using GAE and clipping $Q$ (``+ Clip Q'') raises SR to 89\% by reducing the variance of the estimated advantage to reinforce the action. 
However, maximizing entropy estimated via the Monte Carlo method (``+ MaxEnt (MC)'') leads to a significant drop in performance, with SR falling to 77\% and collision rate (CR) increasing from 9\% to 20\%. This decline occurs because maximizing entropy aims to encourage unexplored actions, but when entropy is estimated from the training batch, it inadvertently discourages positive actions and encourages negative ones. Maximizing the entropy lower bound avoids this issue, as the lower bound provides a conservative estimation of entropy, leading to better training stability.
Lastly, it is noteworthy that improvements in navigation performance often result in better trajectory prediction. This is because an intention policy inherently aims to avoid obstacles and achieve goals, further validating the approach.

\noindent \textbf{Visualization of navigation.} Fig.~\ref{fig: viz_navi} demonstrates some examples of RT-V2's navigation trajectories. Two observations can be made: First, RT-V2 can avoid the obstacles and reach one of the goals in a cluttered scene. Second, RT-V2 does not always choose the nearest goal to approach, which is not a concern in shared control since the intended goal is unknown.

\begin{figure*}[!t]
  \centering
  \includegraphics[width=\linewidth]{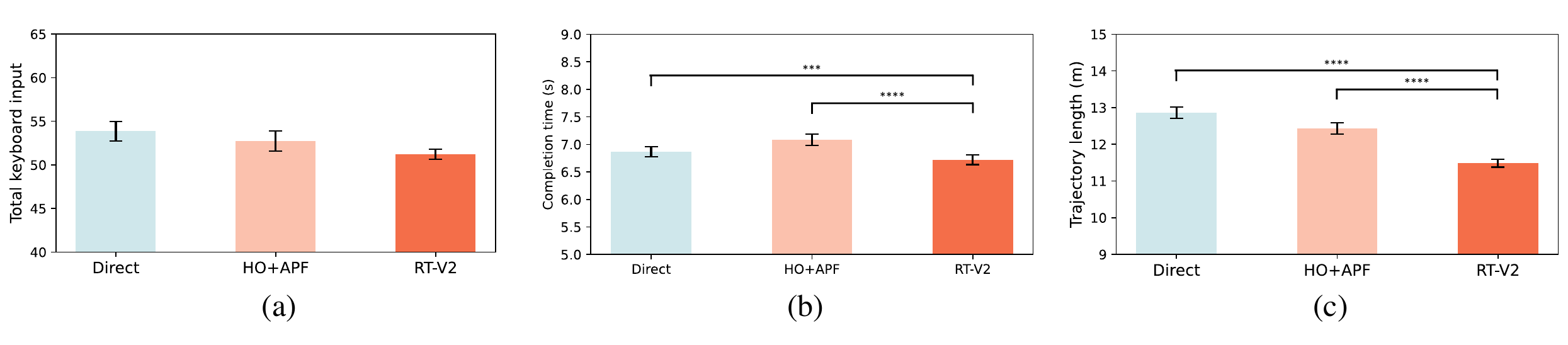}
  \caption{Some results for Human-Exp. (a) Total keyboard inputs. (b) Completion time. (c) Trajectory length. $*= p < 0.05$, $**= p < 0.01$, $***= p < 0.001$, and $ ****= p < 0.0001$.}
  \label{fig: performance}
\end{figure*}


\begin{figure*}[!t]
  \centering
  \includegraphics[width=\linewidth]{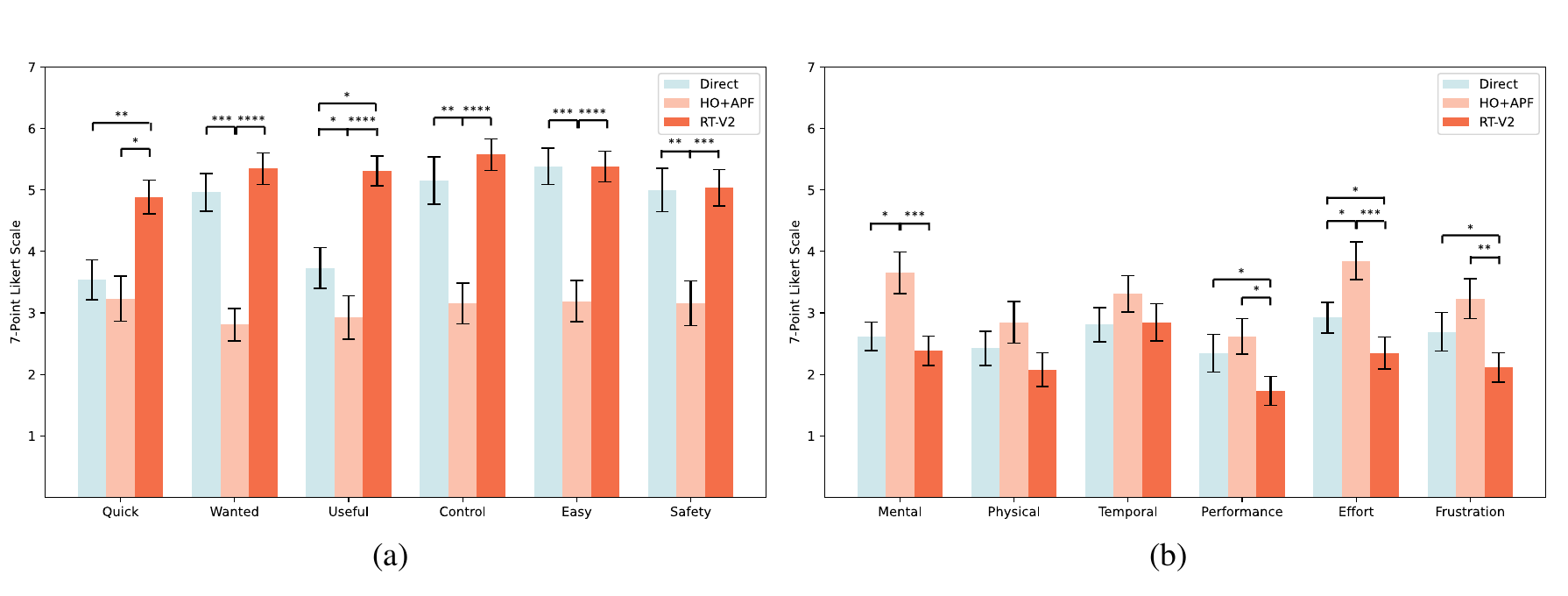}
  \caption{The user study for each method across all participants for Human-Exp in simulation. The plots are bar plots with error bars: (a) the agreement survey and (b) the NASA-TLX survey, where  $*= p < 0.05$, $**= p < 0.01$, $***= p < 0.001$, and $****= p < 0.0001$. For (a) a higher score is better, while for (b) a lower score is better. The color of the bars representing the control methods is consistent across all figures.}
  \label{fig: likert}
\end{figure*}

\begin{figure}[!t]
  \centering
  \includegraphics[width=0.9\linewidth]{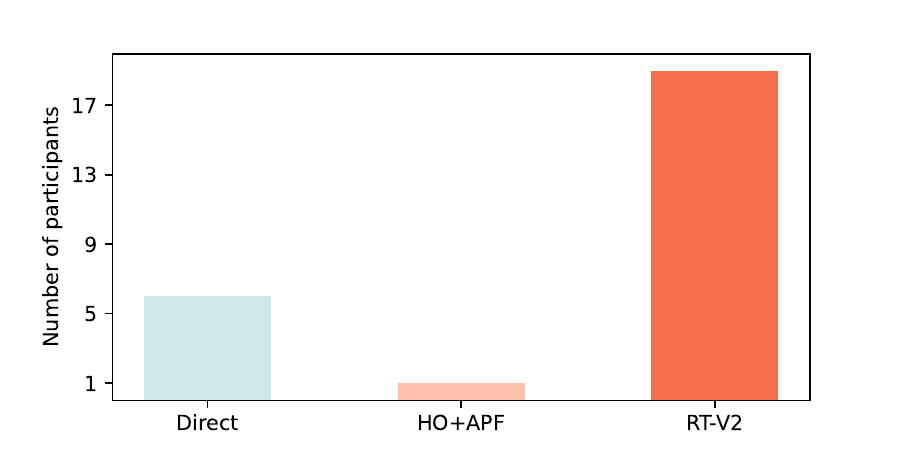}
  \caption{Preference analysis for three methods. Every participant voted for their favorite method after the experiment. RT-V2 is the most popular with 19 votes, while HO+APF is the least favorite with one vote.}
  \label{fig: preference}
\end{figure}

\begin{table}[t]
  \centering
  \caption{The success rate (SR), collision rate (CR), and unfinished rate (UR) of Human-Exp.}
  \begin{tabular}{cccc}
    \toprule
    Method & SR (\%) & CR (\%) & UR (\%)\\
    \hline
    Direct    & 94.6  & 5.4 & 0.0 \\
    HO+APF & 94.2   & 5.8 & 0.0 \\
    RT-V2  &  97.3 & 0.0  & 2.7  \\
    \bottomrule
  \end{tabular}
  \vspace{-0.4cm}
  \label{tab:obj1}
\end{table}

\subsection{Human-Exp: Shared-Autonomy with Human Users} \label{human-exp}
In this experiment, we aim to validate the performance of RT-V2 in a shared autonomy setting. By having human users complete a task in simulation, assisted by RT-V2, we can assess its performance in terms of speed, safety, and user experience.

\noindent \textbf{Design.} We compared three control conditions: (1) pure user control (Direct), (2) Hindsight Optimization combined with an Artificial Potential Field for obstacle avoidance (HO+APF) \citep{javdani2018shared,apf} (implementation details in the Appendix), and (3) RT-V2. The simulated environment followed the \emph{Offline-Exp1} scenario (Fig.~\ref{fig: sim setup}). Control frequency was 10 Hz, and the maximum robot speed was 3 m/s for all methods. Each trial required the participant to drive an omnidirectional virtual robot to a predefined target while avoiding obstacles; remaining potential goals acted as distractors for the assistive controllers, and the robot was permitted to pass through distractor locations. A trial was labeled a failure if the robot collided with an obstacle or if the participant failed to reach the goal within 100 control inputs.

\noindent \textbf{Protocol.} Twenty-six novice participants (19 male, 7 female; age 18–40) completed three rounds, each using a different assistive method (20 trials per round). Participants used a keyboard with four directional keys; diagonal motion was achieved via simultaneous keypresses, enabling eight movement directions. The order of the assistive methods (Direct / HO+APF / RT-V2) was randomized for each participant to mitigate the effects of novelty and practice. After each round, participants rated their agreement with the following statements on a seven-point Likert scale:
\begin{itemize}
    \item This algorithm helped me complete the task \emph{quickly}.
    \item This robot did what I \emph{wanted}.
    \item If I were to navigate a wheelchair, this algorithm would be \emph{useful} for me.
    \item I felt in \emph{control} while using this algorithm.
    \item I would find this algorithm \emph{easy to use}.
    \item I felt \emph{safe} when I used this algorithm.
\end{itemize}

Participants also completed a NASA-TLX survey \citep{nasatlx}, where they rated their workload across six subscales. \emph{Mental demand} assessed the cognitive and perceptual effort required; \emph{physical demand} measured the physical effort involved in the task; \emph{temporal demand} evaluated the time pressure experienced by the participant; \emph{effort} measured how hard the participant had to work to maintain performance; \emph{frustration} assessed feelings of annoyance, stress, and irritation; and \emph{performance} gauged how successful participants felt in completing the task. At the end of the three rounds, participants were also asked to indicate their preferred method and provide written comments. All participants provided signed consent for the experimental procedure, which was approved by the Social and Societal Ethics Committee of KU Leuven (G-2024-8371). The collected data were processed in accordance with the General Data Protection Regulation (GDPR) of the EU.

\noindent \textbf{Metrics.} Both objective and subjective metrics were used in the experiments. For objective metrics, we compared the success rate (SR), collision rate (CR), unfinished rate (UR), total keyboard inputs, completion time, and trajectory length. For subjective metrics, we compared participants' ratings from the agreement survey and their responses to the NASA-TLX survey across the three methods.

\begin{figure*}[!t]
  \centering
  \includegraphics[width=\linewidth]{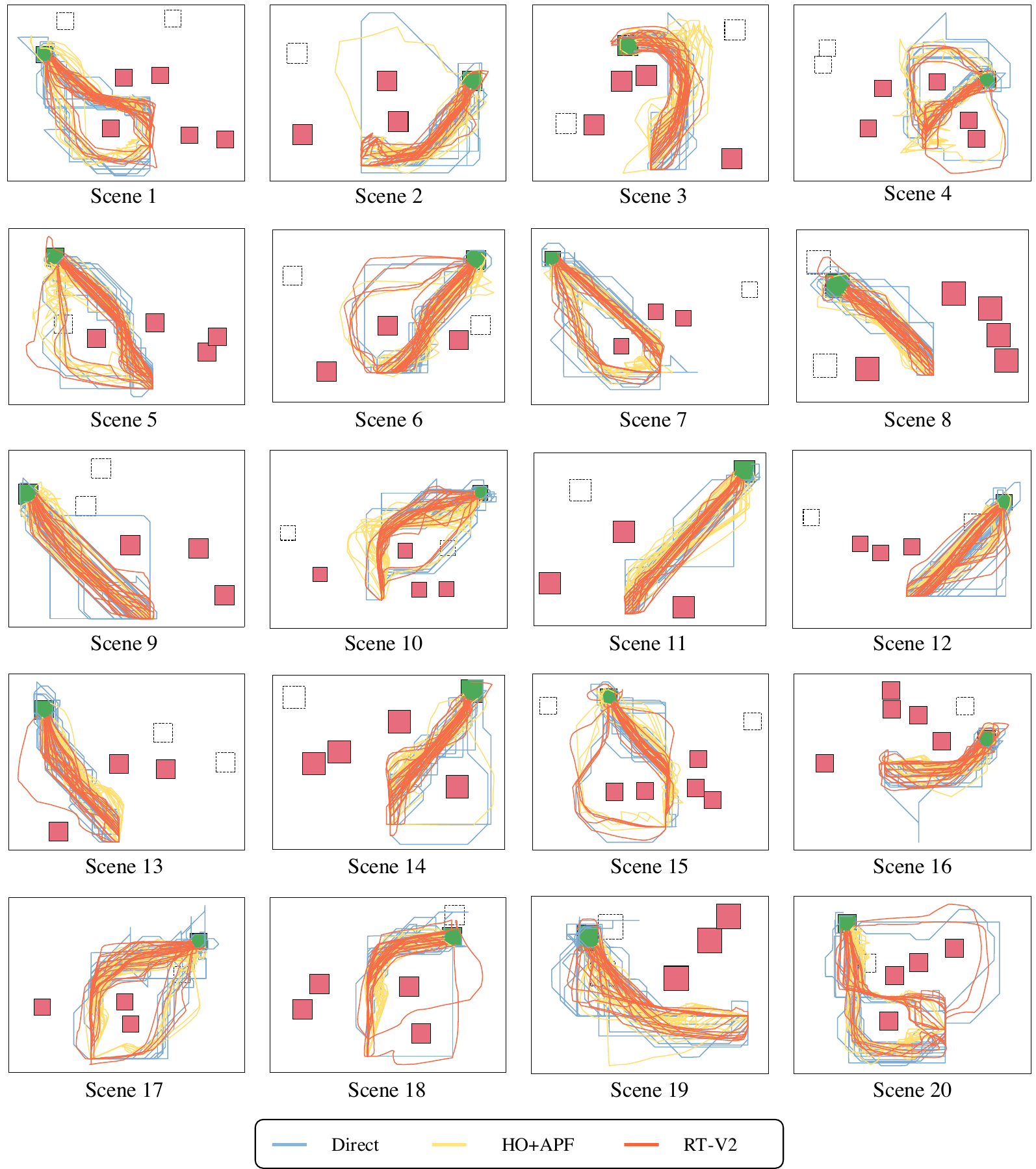}
  \caption{Visualization of successful demonstrations of navigation in simulation. All the trajectories for three methods (Direct, HO+APF, and RT-V2) in 20 scenes are plotted. \textcolor{377E22}{Green boxes}, white dash-line boxes, and \textcolor{E76D7E}{pink boxes} are true goals, distractors, and obstacles, respectively. \textcolor{7DABCF}{Light blue lines} denote the trajectories of Direct, \textcolor{FFDF70}{yellow lines} denote the trajectories of HO+APF, and \textcolor{F46E49}{orange lines} denote the trajectories of RT-V2.}
  \label{fig: trials}
\end{figure*}

\noindent \textbf{Results of Objective Measures.} Table \ref{tab:obj1} summarizes SR, CR, and UR by method. RT-V2 achieved the highest SR (97.3\%) and no collisions across analyzed trials. HO+APF exhibited a higher collision rate than Direct; inspection of trajectories shows that APF-induced reactive forces sometimes produced oscillatory “bouncing” behavior in narrow passages, which in turn provoked corrective overreactions by participants (example trajectories shown as blue lines in Scenes 1,5,6,7,15,20 of Fig.~\ref{fig: trials}).

\begin{figure*}[!t]
  \centering
  \includegraphics[width=\linewidth]{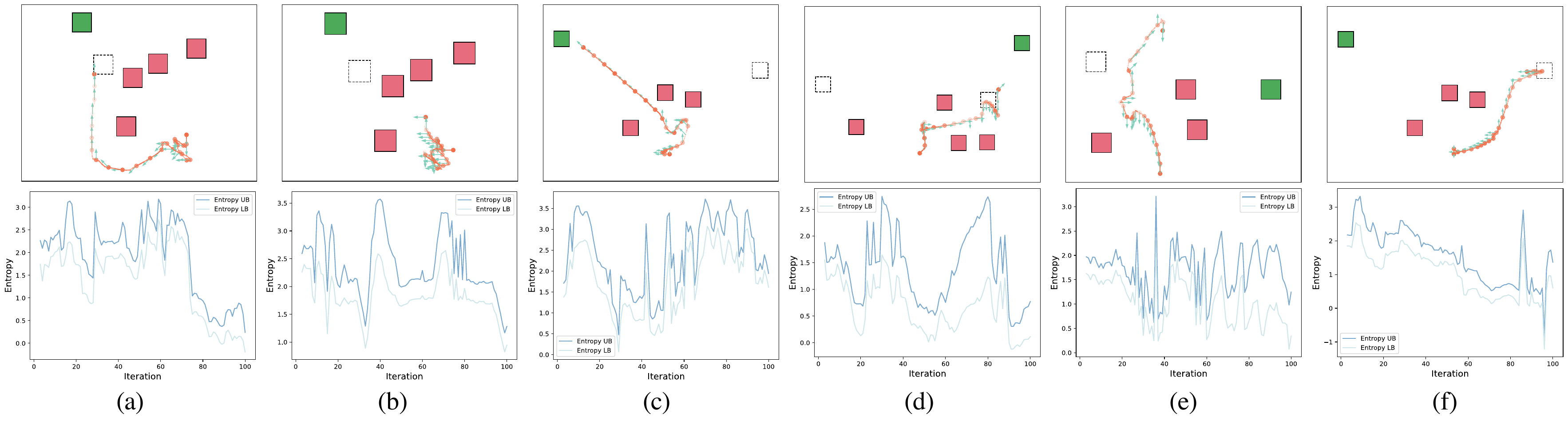}
  \caption{Visualization of failure trials of RT-V2 in Human-Exp. Both trajectory and entropy are plotted. The upper row for each subfigure visualizes the goals (\textcolor{377E22}{green boxes}), obstacles (\textcolor{E76D7E}{pink boxes}), distractors (white dash-line boxes), trajectories (\textcolor{F46E49}{orange lines}), and user commands (\textcolor{81D0BB}{light blue arrow}). We visualize the user commands every 4 iterations for better visibility. The lower row for each subfigure visualizes the upper bound (Entropy UB) and lower bound (Entropy LB) of the entropy of the action-GMMs generated by RT-V2 for each iteration. Besides, the opacity of the trajectories represents the normalized entropy lower bound.}
  \label{fig: failure}
\end{figure*}

Fig.~\ref{fig: performance} compares \emph{total keyboard input}, \emph{completion time}, and \emph{trajectory length}, excluding failure trials. A one-way repeated measures ANOVA was used to evaluate the effect of the method (Direct, HO+APF, RT-V2) on these metrics, and Tukey's test was applied for pairwise comparisons when significant effects were found.
For \emph{total keyboard input}, a significant difference was observed among the methods (\(F(2, 1476)=6.648\), \(p=0.0013\)), with RT-V2 differing significantly from HO+APF (\(p=0.0008\)). The unexpected behavior of HO+APF required users to make more adjustments, resulting in a higher number of total keyboard inputs. In contrast, RT-V2 accurately inferred users' intent, providing smoother control with fewer adjustments.
For \emph{completion time}, significant differences were found (\(F(2, 1476)=16.038\), \(p<0.0001\)), with RT-V2 differing from both Direct (\(p=0.0007\)) and HO+APF (\(p<0.0001\)). When using HO+APF, users sometimes paused to understand its behavior, leading to longer completion times.
For \emph{trajectory length}, significant differences were observed (\(F(2, 1476)=24.897\), \(p<0.0001\)), with RT-V2 differing from both Direct (\(p<0.0001\)) and HO+APF (\(p<0.0001\)). The shorter \emph{trajectory length} corresponds with the results showing fewer \emph{keyboard inputs} and reduced \emph{completion time}.

\begin{figure*}[!t]
  \centering
  \includegraphics[width=\linewidth]{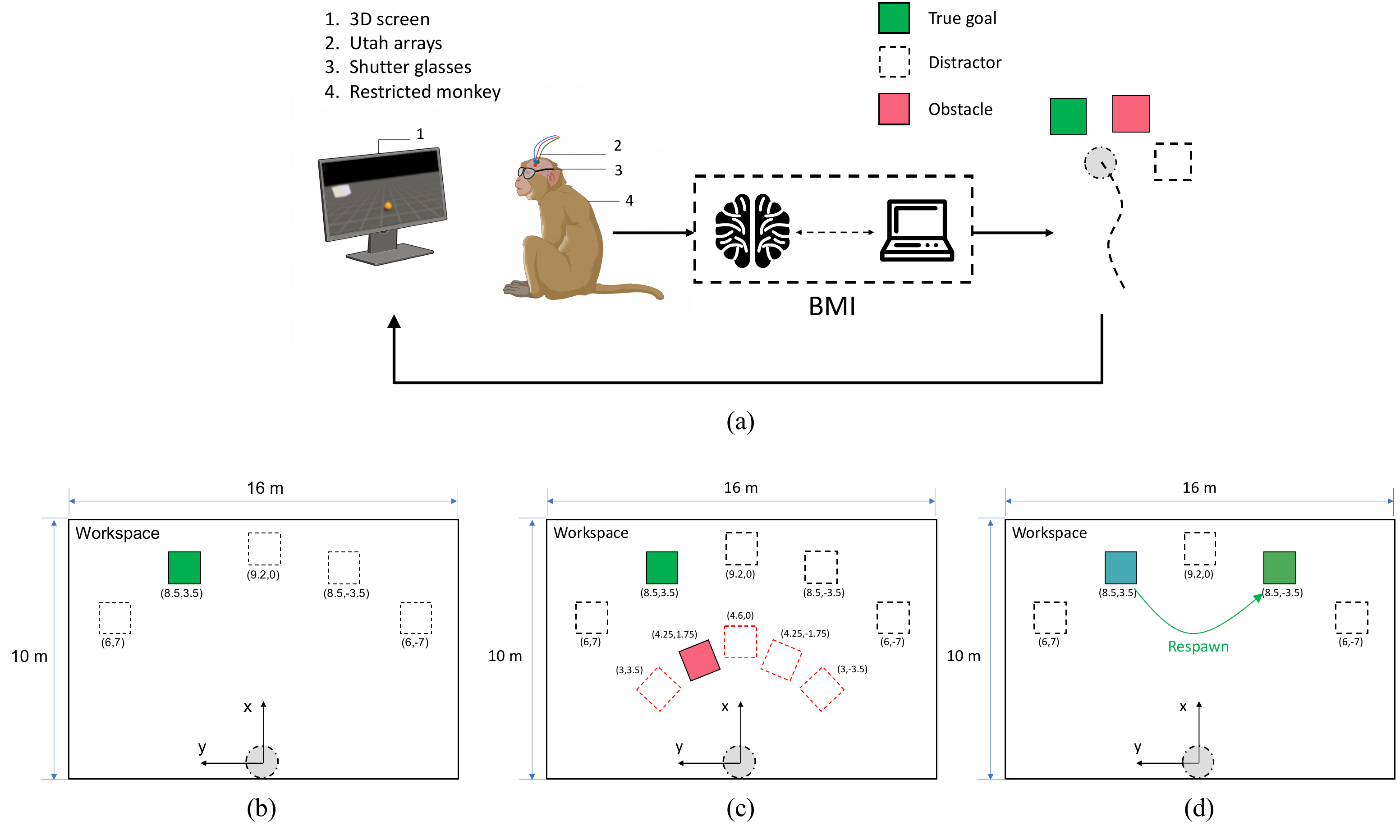}
  \caption{(a) An illustration of BMI experiments. This figure is based on the graphical elements shown in Fig.~4 of Saussus et al.'s work \citep{saussus2025intracortical}. (b) The experimental setup of \emph{Scene 1}: direct reaching task. (c) The experimental setup of \emph{Scene 2}: obstacle avoidance task. (d) The experimental setup of \emph{Scene 3}: target respawning task. See text for details.}
  \label{fig: bmi setup}
\end{figure*}

\noindent \textbf{Results of Subjective Measures.} Fig.~\ref{fig: likert}~(a) illustrates the comparison of the agreement survey results. The Kruskal-Wallis H-test was employed to evaluate the effect of the method (Direct, HO+APF, RT-V2) on these metrics. A Wilcoxon signed-rank test with a two-stage Benjamini-Hochberg correction was used for post hoc analysis of pairwise comparisons when significant effects were identified. Significant differences were observed for all measures. Post hoc analysis revealed significant differences between Direct and HO+APF in \emph{Wanted} ($p=0.0001$), \emph{Useful} ($p=0.0190$), \emph{Control} ($p=0.0020$), \emph{Easy} ($p=0.0002$), and \emph{Safety} ($p=0.0019$). Additionally, significant differences were found between RT-V2 and Direct in \emph{Quick} ($p=0.0012$) and \emph{Useful} ($p=0.0051$). Furthermore, RT-V2 differed significantly from HO+APF in \emph{Quick} ($p=0.0101$), \emph{Wanted} ($p<0.0001$), \emph{Useful} ($p<0.0001$), \emph{Control} ($p<0.0001$), \emph{Easy} ($p<0.0001$), and \emph{Safety} ($p=0.0006$).

Fig.~\ref{fig: likert}~(b) demonstrates the comparison of the NASA-TLX survey. The Kruskal-Wallis H-test evaluated the effect of the method (Direct, HO+APF, RT-V2) on these metrics. A Wilcoxon signed-rank test with two-stage Benjamini-Hochberg correction was used as post hoc analysis for pairwise comparisons when significant effects were found. Significant differences were found for \emph{Mental} ($H(77)=8.457, p=0.0146$), \emph{Performance} ($H(77)=7.763, p=0.026$), \emph{Effort} ($H(77)=13.211, p=0.0013$), \emph{Frustration} ($H(77)=6.329, p=0.0422$). Post hoc analysis found significant differences between Direct and HO+APF in \emph{Mental} ($p=0.0029$) and \emph{Effort} ($p=0.0192$), between RT-V2 and Direct in \emph{Performance} ($p=0.0114$), \emph{Effort} ($p=0.0320$) and \emph{Frustration} ($p=0.0403$), and between RT-V2 and HO+APF in \emph{Mental} ($p=0.0003$), \emph{Performance} ($p=0.0101$), \emph{Effort} ($p=0.0002$) and \emph{Frustration} ($p=0.0015$). No significant difference was found in \emph{Physical} demand, as the task involved only keyboard control and wasn't physically taxing. Similarly, the lack of urgency in achieving the goal resulted in no significant difference in \emph{Temporal} demand.

For preference analysis, most participants chose RT-V2 as their favorite, while only one participant preferred HO+APF. According to their comments, participants felt that the assistance provided by HO+APF was ``too present'' and ``too aggressive''. The robot seemed to have a mind of its own, occasionally not following their commands, and approached the goal in ways the participants did not want. HO+APF often forced participants to take a shorter path, which might not have aligned with their intent. As a result, participants spent more time wandering in place and figuring out their next actions, which is consistent with the shorter trajectory length but longer completion time and more total inputs of HO+APF shown in Fig.~\ref{fig: performance}.
Furthermore, because participants could not anticipate the behavior of HO+APF, they felt unsafe. These findings suggest that participants' sense of safety is not directly related to the actual safety level, as HO+APF had a similar collision rate (CR) to Direct in Table~\ref{tab:obj1}, but a much lower \emph{Safety} score in Fig.~\ref{fig: likert}~(a). Moreover, three participants reported that HO+APF took more time to master. They mentioned that, with more practice, they might have chosen HO+APF as their favorite. These comments align with the low \emph{Easy} score in Fig.~\ref{fig: likert}. All participants reported that RT-V2 provided a suitable level of assistance. Some expert participants, who could complete the task smoothly with Direct control, mentioned that they noticed few differences in user experience between Direct and RT-V2. This is because RT-V2 adjusts the level of assistance based on contextual uncertainty: it provides minimal assistance when the robot is far from potential goals and obstacles, and stronger assistance when the robot is about to collide with obstacles or reach the predicted goal. These comments align with the similar \emph{Easy} and \emph{Wanted} scores between Direct and RT-V2 in Fig.~\ref{fig: likert}.
However, participants also noted that sometimes RT-V2 took control and overrode their commands, leading to a 2.7\% unfinished rate (UR) in Table~\ref{tab:obj1}.

\noindent \textbf{Visualization of Demonstrations.} We visualize all successful trajectories for all methods in Fig.~\ref{fig: trials}, where blue lines represent the trajectories of Direct, yellow lines represent the trajectories of HO+APF, and orange lines represent the trajectories of RT-V2. Green boxes, white dash-line boxes, and pink boxes correspond to true goals, distractors, and obstacles, respectively. From Fig.~\ref{fig: trials}, the trajectories of RT-V2 are smoother and shorter compared to those of Direct and HO+APF. When the distractors were near the true goal, HO+APF sometimes dragged the robot toward the distractors, as shown in Scenes 4, 10, and 17. When the robot passed through one distractor to reach the true goal behind, HO+APF mistakenly predicted the distractor as the intended goal and prevented the robot from leaving, causing oscillations as shown in Scenes 5, 12, 17, 19, and 20. This inaccurate intent estimation of HO+APF led to low scores for \emph{Wanted} and \emph{Control} in Fig.~\ref{fig: likert}.

\noindent \textbf{Visualization of Failure Trials.} Fig.~\ref{fig: failure} shows the trajectories of some failure trials, in which the users failed to achieve the goals within 100 keyboard inputs. The upper row for each subfigure visualizes the goals (green boxes), obstacles (pink boxes), distractors (white dash-line boxes), trajectories (orange lines), and user commands (light blue arrows). Since RT-V2 uses implicit policy blending without an explicit arbitrator, to further analyze the authority of RT-V2, we also visualize the entropy of the action-GMMs in the lower row of each subfigure. The higher the entropy, the stronger the assistance. There is no analytical expression for the entropy of GMMs, so we plot the upper and lower bounds of the entropy based on \citep{huber2008entropy} (the math can be referred to in the Appendix). Fig.~\ref{fig: failure}~(a), (b), and (c) demonstrate the cases in which RT-V2 hindered the user from choosing another way to avoid the obstacles. In these cases, a drastic decrease in entropy was observed every time the user aimed to move downward to avoid the obstacles, where RT-V2 overrode the user's commands. Fig.~\ref{fig: failure}~(d), (e), and (f) demonstrate the cases where RT-V2 was overconfident about certain distractors and trapped the robot around the distractors, with a drastic decrease in entropy observed. The reason for these failures lies in the breaking of a preliminary condition: the training data should be collected from the user. Based on Eq.~\ref{eq: ideal shared control}, the intended trajectories are collected from the interaction with the user and the environment, and then RT-V2 can capture the behavioral patterns of that specific user. In this experiment, collecting a massive amount of data from participants was impossible. Therefore, we trained the model with trajectories generated by PRM. What RT-V2 captured were the behavior patterns of PRM, though the behavior of PRM shares some common characteristics with the behavior of human users. As a result, RT-V2 sometimes prohibited the user from moving in certain directions due to the lack of such patterns in the dataset. We believe that by developing RT-V2 for a specific user and training it with user-specific data, we can avoid the failure cases mentioned above.

\subsection{BMI-Exp1: Trajectory Prediction}
In this experiment, we aim to finetune the models pre-trained on Traj1M and evaluate the model's performance on the real BMI data to further validate the effectiveness of the proposed RT-V2.

\noindent \textbf{Experimental Setting.} We use two monkeys with invasive chip implants to collect demonstrations as a dataset. Our BMI experiment setup is similar to the work of Saussus et al. \citep{saussus2025intracortical}: As shown in Fig.~\ref{fig: bmi setup}~(a), the monkey sat in a chair with its head fixed and arms restrained in front of a ViewPixx 3D screen in a dark room. On the screen, pairs of images for the left and right eye with slight angular disparities were presented alternately at 120 Hz. The monkey wore shutter glasses with liquid crystal lenses that opened and closed when voltage was applied, which was perfectly synchronized with the screen at a frequency of 60Hz for each eye to achieve stereoscopic vision. The monkeys were implanted with three 96-channel Utah arrays, which transmit the BMI signals for decoding into velocity commands. They were trained to use BMI to control an omnidirectional virtual robot to reach the goal in the Unity simulation. We collect demonstrations in two scenes:

\noindent \textbf{(i) Scene 1}: Direct Reaching. As shown in Fig.~\ref{fig: bmi setup}~(b), the robot is initially placed at $(0,0)$ in a $10 \times 16~m^2$ Unity workspace. In each round, a goal is placed at one of the potential goal positions $(6,7)$, $(8.5,3.5)$, $(9.2,0)$, $(8.5,-3.5)$, $(6,-7)$ (dubbed Goal0$\sim$Goal4). Distractors are placed at the rest of the positions to distract intent estimators. The monkeys are required to control the robot to reach the indicated goal.

\noindent \textbf{(ii) Scene 2}: Collision Avoidance. As shown in Fig.~\ref{fig: bmi setup}~(c), the basic setup follows Scene 1. In addition, one obstacle with a side length of $1.5~m$ is placed at one of the potential obstacle positions $(3,3.5)$, $(4.25,1.75)$, $(4.6,0)$, $(4.25,-1.75)$, $(3,-3.5)$, which are in the middle of the way to the goal. The monkeys are required to control the robot to get around the obstacle and reach the indicated goal.

We collected 3180 successful trajectories in total, of which 2225 were for finetuning, and 955 were for evaluation. We call this dataset \emph{TrajBMI}. We compare the proposed RT-V2 with our previous work RT \citep{RT}. Following the evaluation paradigm in RT \citep{RT}, \emph{Best-of-20} and \emph{Most likely} trajectories are sampled to compute ADE and FDE metrics. Additionally, because both RT and RT-V2 are probabilistic models, we can evaluate the negative log-likelihood (NLL) of the ground-truth trajectory with both models. We report NLL as a metric in the comparison.

\noindent \textbf{Performance in the TrajBMI Dataset.} 
Results are shown in Table~\ref{tab:trajBMI}. In the Best-of-20 setting, RT-V2 achieves 331.23 mm ADE and 566.31 mm FDE, outperforming RT with 356.64 mm ADE and 621.98 mm FDE. In the Most-likely setting, RT-V2 achieves 584.07 mm ADE, which is slightly lower than RT's ADE (577.65 mm), but RT-V2 outperforms RT in FDE (1156.16 mm vs. 1173.52 mm). RT-V2's NLL also outperforms RT's. The lower NLL not only aligns with the better performance in ADE and FDE in both the Best-of-20 and Most-likely settings, but also demonstrates better intent estimation accuracy. In posterior decision-making, the lower NLL means that the user will agree with the predicted action with a high probability, which can provide smoother cooperation with RT-V2 for the user.

\begin{table}[t]
  \centering
  \caption{The performance comparison in TrajBMI. The best performance in each metric is highlighted in bold.}
  \begin{tabular}{cccccc}
    \toprule
    \multirow{2}*{Method}   & \multicolumn{2}{c}{Best-of-20 (mm)} & \multicolumn{2}{c}{Most likely (mm)} & \multirow{2}*{NLL} \\
    \cmidrule{2-5}
    & ADE& FDE& ADE & FDE \\
    \midrule
    RT & 356.64  &  621.98  &  \textbf{577.65} &  1173.52 & 27.02\\
    RT-V2     &  \textbf{331.23}  & \textbf{566.31} & 584.07 & \textbf{1156.16} & \textbf{23.54} \\
    \bottomrule
  \end{tabular}
  \label{tab:trajBMI}
\end{table}


\begin{figure*}[!t]
  \centering
  \includegraphics[width=\linewidth]{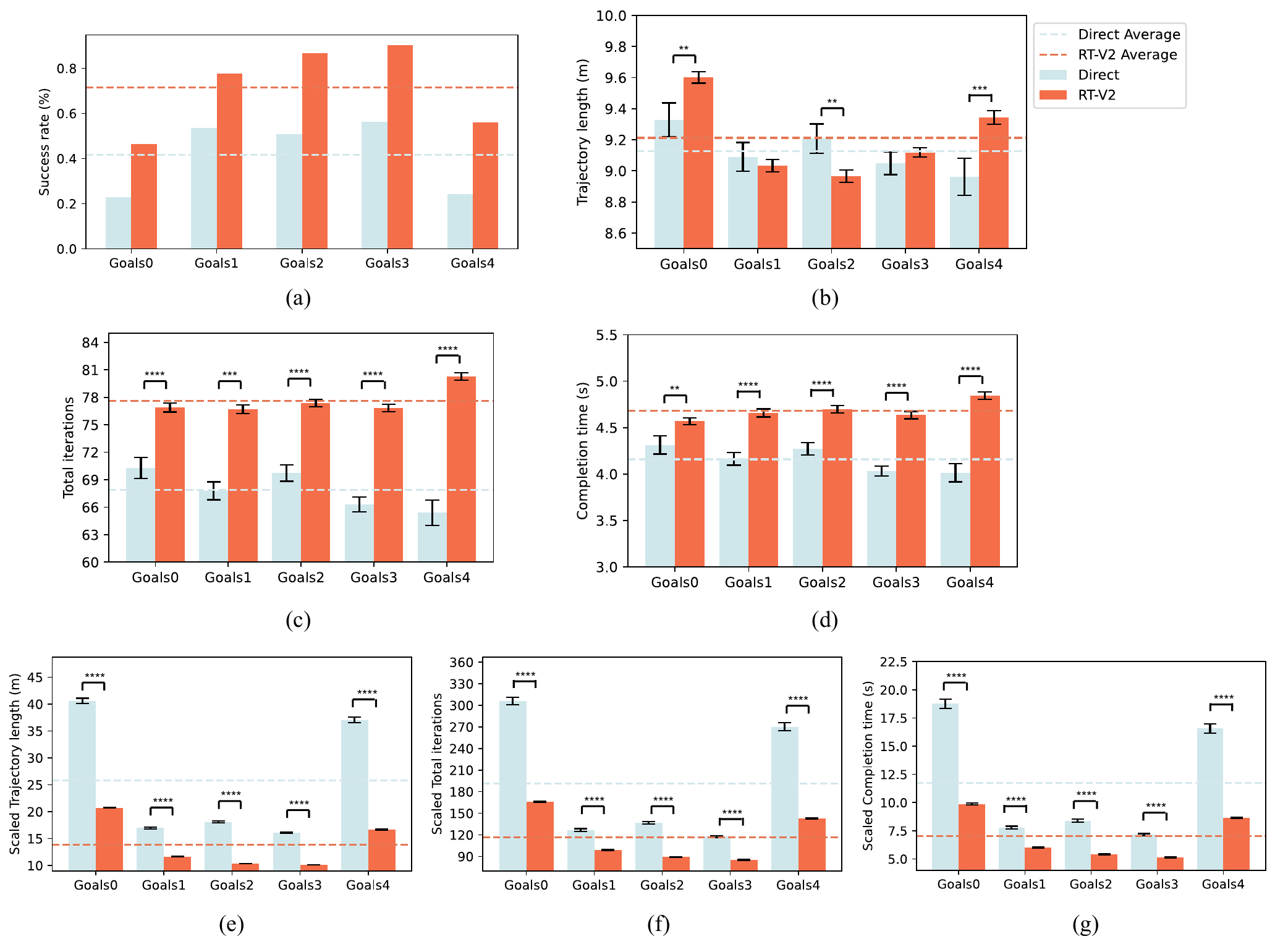}
  \caption{Results of the BMI fixed-obstacle shared autonomy experiments. The results are pooled for two monkeys. (a) Success rate. (b) Trajectory length. (c) Total iterations. (d) Completion time. (e) Scaled trajectory length. (f) Scaled total iterations. (g) Scaled completion time. $*= p < 0.05$, $**= p < 0.01$, $***= p < 0.001$, and $****= p < 0.0001$.}
  \label{fig: fixed_obs}
\end{figure*}

\begin{figure*}[!t]
  \centering
  \includegraphics[width=\linewidth]{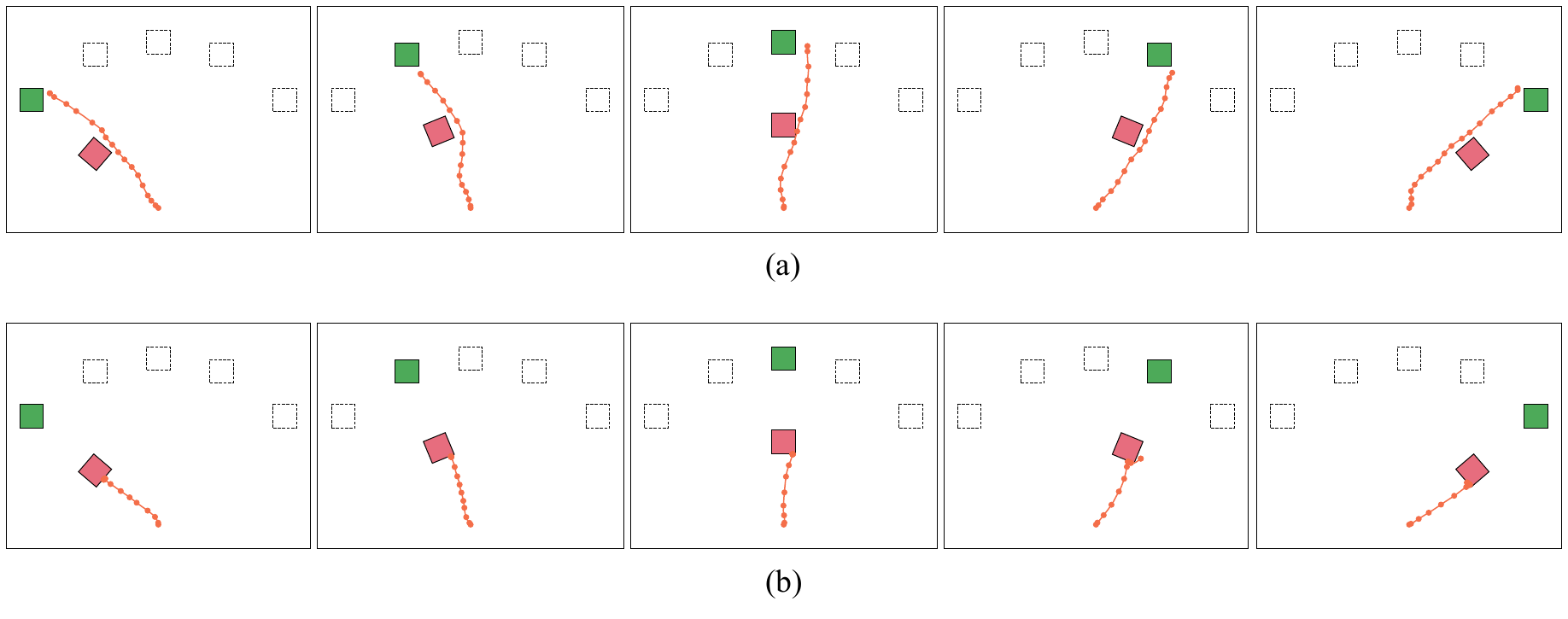}
  \caption{Visualization of trials of direct BMI control. (a) Successful trials. (b) Failure cases. Each subfigure visualizes the goals (\textcolor{377E22}{green boxes}), obstacles (\textcolor{E76D7E}{pink boxes}), distractors (white dash-line boxes), and trajectories (\textcolor{F46E49}{orange lines}). We visualize the user commands every 4 iterations for better visibility. With direct BMI control, the monkey behaves aggressively.}
  \label{fig: bmi_demo_direct}
\end{figure*}

\begin{figure*}[!t]
  \centering
  \includegraphics[width=\linewidth]{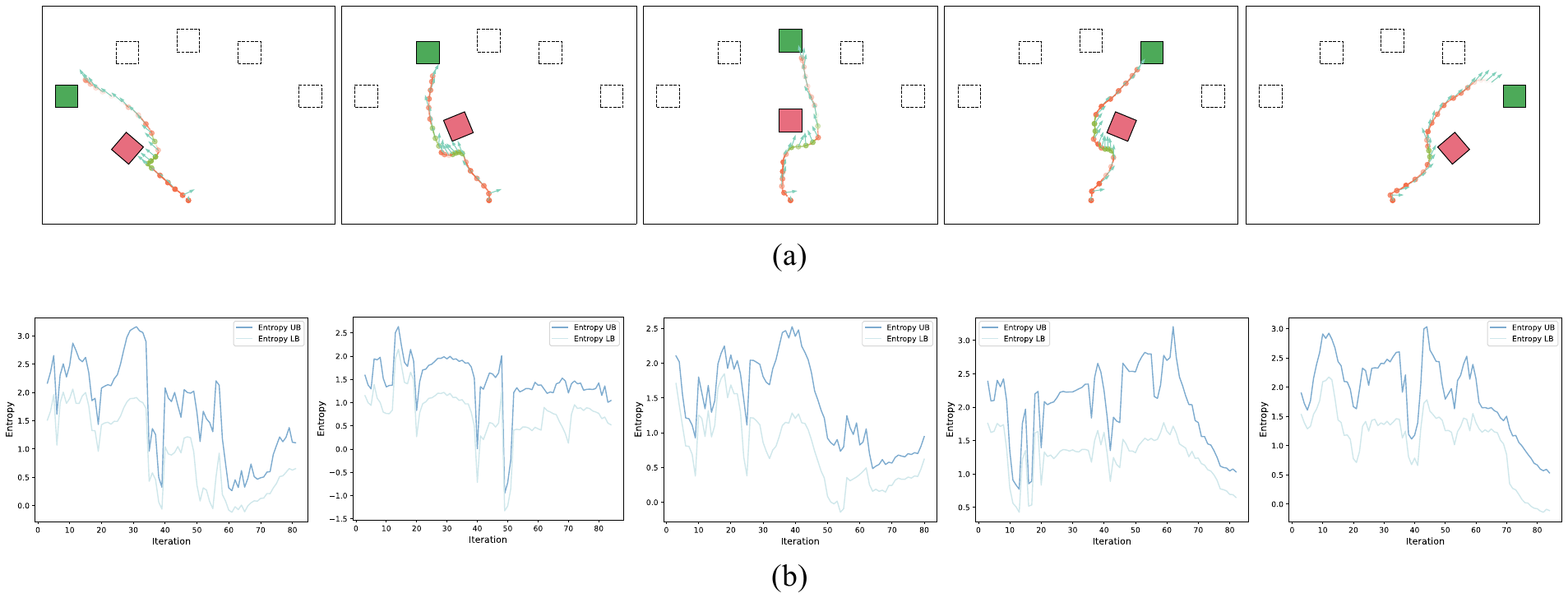}
  \caption{Visualization of trials assisted by RT-V2 in BMI-Exp2. Both (a) trajectory and (b) entropy are plotted. (a) Each subfigure visualizes the goals (\textcolor{377E22}{green boxes}), obstacles (\textcolor{E76D7E}{pink boxes}), distractors (white dash-line boxes), trajectories (\textcolor{F46E49}{orange lines}), and commands from BMI (\textcolor{81D0BB}{light blue arrow}). Besides, we highlight the part which is fully controlled by RT-V2 in \textcolor{lightgreen}{light green}. We visualize the user commands every 4 iterations for better visibility. (b) Each subfigure visualizes the upper bound (Entropy UB) and lower bound (Entropy LB) of the entropy of the action-GMMs generated by RT-V2 for each iteration. Besides, the opacity of the trajectories represents the normalized entropy lower bound.}
  \label{fig: bmi_demo_RT}
\end{figure*}

\begin{figure}[!t]
  \centering
  \includegraphics[width=\linewidth]{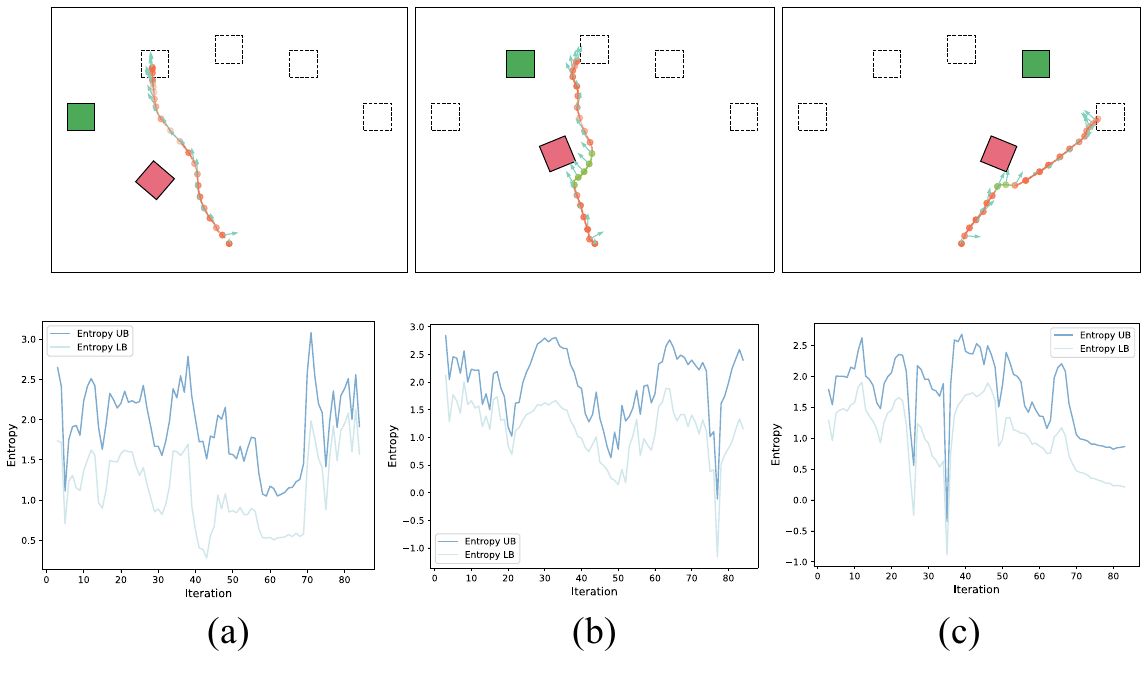}
  \caption{Visualization of failure trials assisted by RT-V2. The upper row shows the conducted trajectories and the bottom row shows the corresponding entropy for each iteration. }
  \label{fig: bmi_failure_RT}
\end{figure}

\subsection{BMI-Exp2: Shared Autonomy Experiment with a Fixed Obstacle}
We evaluated RT-V2 in a BMI shared-control task with two monkeys to test its generalizability across interfaces and species.

\noindent \textbf{Design.} 
We compared the RT-V2 assistive control against direct control using an invasive BMI interface. The experimental setup follows the \textit{Scene 2} in Fig.~\ref{fig: bmi setup}~(b), where the monkeys are required to reach the true target while an obstacle is placed between the target and the starting position. Each trial contained five potential goal locations; one was the true target, and the others were distractors. Notably, the distractors are visible only to RT-V2 (i.e., not visible on the screen).

\noindent \textbf{Protocol.} We conducted 12 sessions for Monkey 1 and 11 sessions for Monkey 2. In each session, both RT-V2 and direct BMI control (dubbed “Direct”) were activated randomly. If a monkey fails to complete the task within 10 seconds, the trial is considered a failure. As a result, we collected 1088 trials for Monkey 1 (495 from RT-V2 and 593 from Direct) and 888 trials for Monkey 2 (343 from RT-V2 and 545 from Direct). The control frequency of this experiment is 20 Hz (i.e., 20 commands sent to the simulation per second). 
Additionally, RT-V2 takes full control for three consecutive iterations if the current BMI command is about to cause a collision with the obstacle (i.e., a collision will happen if this command is executed directly).

All surgical and experimental procedures were approved by the ethical committee on animal experiments of the KU Leuven and performed according to the National Institutes of Health’s Guide for the Care and Use of Laboratory Animals and the EU Directive 2010/63/EU.

\noindent \textbf{Metrics.} We compared the success rate, trajectory length, total iterations, and completion time. Note that trajectory length, total iterations, and completion time are measured only for successful trials. Since different goal positions significantly affect these metrics (with the leftmost and rightmost goals being more challenging), we also provide a separate discussion for each goal position.

\noindent \textbf{Results.}
Fig.~\ref{fig: fixed_obs}~(a) illustrates the success rates for each potential goal position. RT-V2 achieves a higher success rate across all goal positions, achieving an average success rate of 71.5\%, compared to only 42.0\% with Direct control, with especially pronounced improvements for the challenging leftmost and rightmost goals. 

Fig.~\ref{fig: fixed_obs} (b-g) show the \textit{Trajectory Length}, \textit{Total Iterations}, and \textit{Completion Time} for each goal. Regarding trajectory length, a notable increase is observed for the leftmost (Goal0), middle (Goal2), and rightmost (Goal4) targets (see Fig.~\ref{fig: fixed_obs}~(b)). This increase is attributed to RT-V2 correcting mistaken actions in trials that would likely have otherwise failed. As for total iterations and completion time, a significant increase is observed across all goals (see Figs.~\ref{fig: fixed_obs} (c,d)). This increase occurs because, without RT-V2's assistance, monkeys approach goals more directly and often become trapped by the obstacle, resulting in failed trials (as shown in Fig.~\ref{fig: bmi_demo_direct}). RT-V2 helps the monkeys circumvent obstacles, which results in longer trajectories and more time spent, but leads to more successful task completion. To account for this bias, we introduce new metrics to more fairly evaluate RT-V2. Specifically, we normalize the trajectory length, total iterations, and completion time by dividing by the success rate, resulting in the scaled trajectory length, scaled total iterations, and scaled completion time. These new metrics can be interpreted as the trajectory length, number of iterations, and time required to accomplish one successful trial. As shown in Figs.~\ref{fig: fixed_obs} (e-g), we observe significant improvements across all three metrics for every goal, further demonstrating the effectiveness of the proposed RT-V2.



\noindent \textbf{Behavioral Analysis.} The BMI decoding signal has limited steering bandwidth and is relatively noisy, which makes fine steering maneuvers (for example, curving smoothly around an obstacle) difficult to execute. In direct BMI trials (Fig.~\ref{fig: bmi_demo_direct}), successful completion typically required the monkey to initiate a steering trajectory early and maintain a smooth curved path around the obstacle. Abrupt changes of direction near the obstacle were rarely achievable because the decoded commands tended to continue pointing into the obstacle, causing the robot to stall. By contrast, RT-V2 intervenes by modifying and smoothing decoded commands and by biasing the controller toward the inferred target; this enables the system to guide the robot around the obstacle (Fig.~\ref{fig: bmi_demo_RT}) and reduces failures caused by decoder noise.

We examined failure cases that occurred even when RT-V2 assisted (representative examples are shown in Fig.~\ref{fig: bmi_failure_RT}). We categorized failures into three types: (a) early, persistent errors — the monkey initially steered toward the wrong location and did not attempt corrective steering afterward; (b) overshoot-without-recovery — the monkey passed the obstacle but then failed to steer back toward the true target; and (c) late corrections blocked by low confidence — the monkey attempted a correction only when near a distractor, but RT-V2’s entropy was low (i.e., the planner was overconfident in the prediction to the distractor), so the controller did not apply a sufficient corrective bias and the robot became trapped near the distractor. Among these errors, (a-b) are caused by monkeys, and only (c) is caused by RT-V2.

\begin{figure*}[!t]
  \centering
  \includegraphics[width=\linewidth]{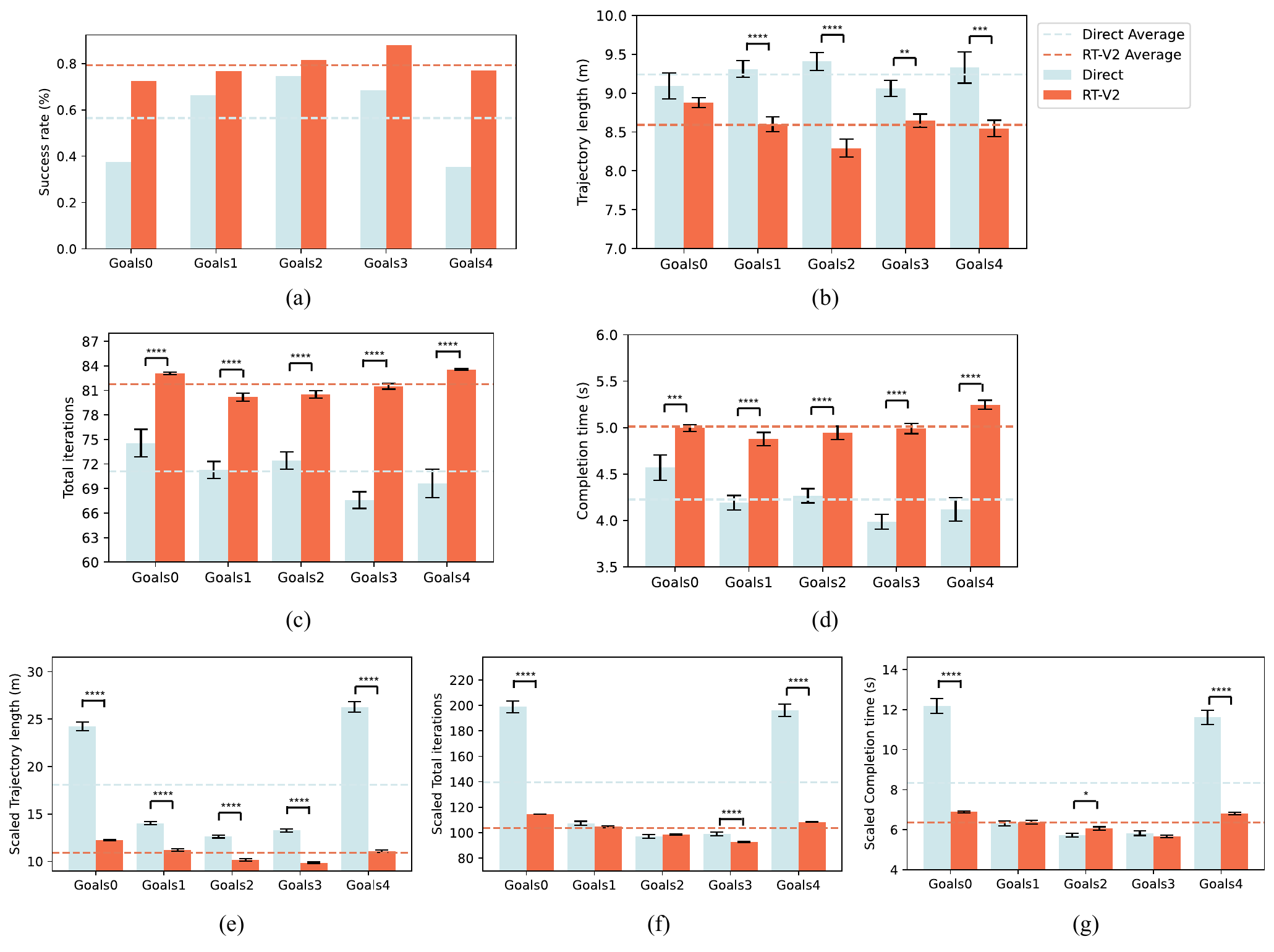}
  \caption{Results of the BMI appearing-obstacle shared autonomy experiments. (a) Success rate. (b) Trajectory length. (c) Total iterations. (d) Completion time. (e) Scaled trajectory length. (f) Scaled total iterations. (g) Scaled completion time. $*= p < 0.05$, $**= p < 0.01$, $***= p < 0.001$, and $****= p < 0.0001$.}
  \label{fig: appearing_obs}
\end{figure*}

\begin{figure*}[!t]
  \centering
  \includegraphics[width=\linewidth]{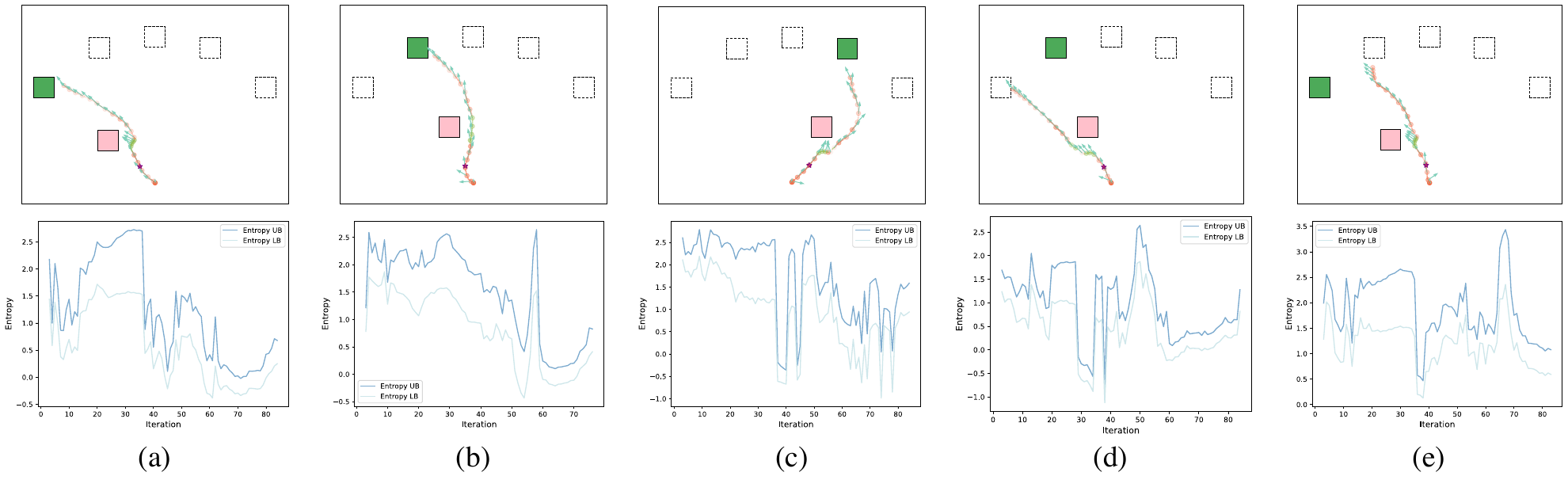}
  \caption{Visualization of trials of RT-V2 in BMI-Exp3. (a-c) Successful trials. (d-e) Failure cases. Each subfigure visualizes the goals (\textcolor{377E22}{green boxes}), appearing obstacles (\textcolor{pink}{light pink boxes}), distractors (white dash-line boxes), and trajectories (\textcolor{F46E49}{orange lines}). The star in the trajectory denotes the time when the obstacle appears. We visualize the user commands every 4 iterations for better visibility.}
  \label{fig: appearobs_demo}
\end{figure*}

\begin{figure*}[!t]
  \centering
  \includegraphics[width=\linewidth]{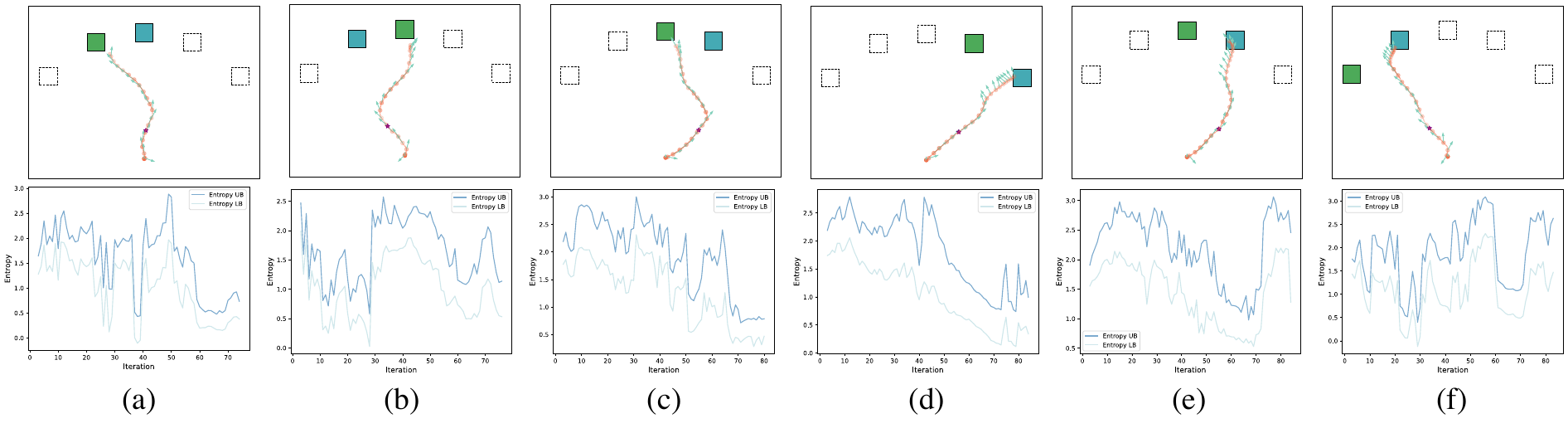}
  \caption{Visualization of trials of RT-V2 in BMI-Exp4. (a-c) Successful trials. (d-e) Failure cases. Each subfigure visualizes the target before respawning (\textcolor{45AAB4}{blue boxes}), the true target after respawning (\textcolor{377E22}{green boxes}), distractors (white dash-line boxes), and trajectories (\textcolor{F46E49}{orange lines}). The star in the trajectory denotes the time when the respawning happens. We visualize the user commands every 4 iterations for better visibility.}
  \label{fig: bmi_demo_respawning}
\end{figure*}

\subsection{BMI-Exp3: Shared Autonomy Experiment with an Appearing Obstacle} \label{bmi-exp3}
We evaluated RT-V2 on a harder task in which an obstacle appears during the movement, testing the method’s adaptability to sudden environmental changes. We used the same primary metrics as BMI-Exp2 (success rate, trajectory length, total iterations, completion time).

\noindent \textbf{Design.} The setup follows \textit{Scene 2} (Fig. \ref{fig: bmi setup}~(c)) and is identical to BMI-Exp2 except that the obstacle is not visible at trial start and appears midtrajectory. The target is one of several potential goal locations; distractor locations were not shown to the monkeys and were available only to RT-V2. To prevent the monkeys from learning to always expect an obstacle, occasional no-obstacle catch trials were interleaved; these catch trials were excluded from analysis.

\noindent \textbf{Protocol.} We ran 10 sessions for Monkey 1. In each session, RT-V2 and Direct (invasive BMI) control were randomly interleaved. Trials with completion time $>$ 10s were considered failures. We collected 1058 analyzed trials in total (515 Direct, 543 RT-V2). The control loop and arbitration ran at 20Hz (50ms per iteration). If the decoded BMI command predicted a collision, RT-V2 assumed full control for three consecutive iterations (150~ms).

\noindent \textbf{Results.} Fig.~\ref{fig: appearing_obs}~(a) illustrates per-goal success rates: RT-V2 achieved an overall success rate of 79.4\% versus 57.1\% for Direct, with the largest gains on the outer (leftmost/rightmost) goals. Figs.~\ref{fig: appearing_obs} (b-g) show the \textit{Trajectory Length}, \textit{Total Iterations}, and \textit{Completion Time} per goal. Figs.~\ref{fig: appearing_obs} (e-g) show the same metrics rescaled by success rate. RT-V2 produced shorter, smoother trajectories (reduced trajectory length) but had higher total iterations and completion time in raw metrics, indicating that RT-V2’s corrective behavior trades speed for path quality. After rescaling by success rate (metrics per successful trial), RT-V2 outperformed Direct across all three measures, especially for the leftmost and rightmost goals.

\noindent \textbf{Behavioral Analysis.} Representative successful RT-V2 trials are shown in Fig.~\ref{fig: appearobs_demo}(a–c) — RT-V2 detects the appearing obstacle, modifies decoded commands, and guides the robot around the obstacle toward the true target. Failure examples are in Fig. \ref{fig: appearobs_demo}(d–e): (d) the monkey passes the obstacle but does not attempt corrective steering afterward, preventing RT-V2 from inferring the true goal; (e) the monkey attempts a late correction, but RT-V2’s entropy is low (high confidence in an incorrect distractor) so the planner does not apply a sufficient corrective bias and overrides the BMI command, yielding a failure.

\begin{figure}[!t]
  \centering
  \includegraphics[width=\linewidth]{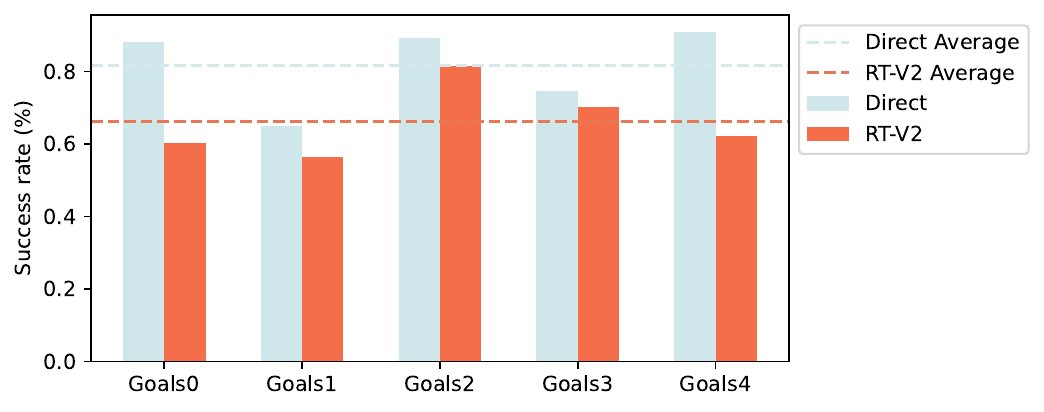}
  \caption{Success rates across all goal positions in BMI-Exp4.}
  \label{fig: acc respawning}
\end{figure}

\subsection{BMI-Exp4: Shared Control Experiment with a Respawning Target} \label{bmi-exp4}
Shared-control systems face an action–noise dilemma: the controller must distinguish deliberate changes in user intent from transient/noisy deviations in the interface signal. To test RT-V2’s ability to adapt to genuine intent changes, we designed a midtrial target-change (``respawn'') paradigm in which the true target switches to a different potential goal location during movement.

\noindent \textbf{Design.} The task follows Scene 3 (Fig.~\ref{fig: bmi setup}~(d)). At trial start, the monkey sees the initial target (distractors are not shown to the monkey and are available only to RT-V2). At a randomized midtrajectory time (when the robot has traveled approximately 40–60\% of the straight-line distance to the initial target), the true target switches to another potential goal; the monkey can see the new target position on the display. Occasional no-switch catch trials were interleaved to prevent expectation bias; catch trials were excluded from the analyzed set.

\noindent \textbf{Protocol.} We ran 10 sessions for Monkey 1. RT-V2 and Direct (invasive BMI) control were randomly interleaved within sessions. Trials exceeding 10s were marked as failures. After excluding catch trials, we analyzed 1786 trials total (891 RT-V2, 895 Direct). The control loop and arbitration ran at 20Hz (50ms per iteration). RT-V2 used the same takeover rule as in earlier experiments.

\noindent \textbf{Result.} Fig.~\ref{fig: acc respawning} shows the success rates for each potential goal position. RT-V2 suffers from a performance decrease by 15.4\%, achieving an average success rate of 67.4\%, compared to Direct control, which has a 79.7\% success rate. This drop is consistent with a loss of flexibility when the controller applies assistance: when monkeys change intent late, RT-V2’s confidence is often already concentrated on the initial (now incorrect) target (low entropy), so RT-V2 applies corrective bias that can override the monkey’s late steering attempts.
Fig.~\ref{fig: bmi_demo_respawning} visualizes the successful cases and failure cases with the assistance of RT-V2. In Fig.~\ref{fig: bmi_demo_respawning}~(a-c), the monkey responds to the respawn quickly, and RT-V2 also adapts to the change of intent. In Fig.~\ref{fig: bmi_demo_respawning}~(d-f), the monkey changes the motion at the last moment, when RT-V2 is confident of its prediction (low entropy) and overrides the monkey's command.

\section{Limitations and Future Work}
This paper introduced RT-V2, a probabilistic shared-control method that uses a Bayesian framework to infer user intent via posterior estimation. RT-V2 connects trajectory prediction, imitation learning, and shared control, and we validated it across prediction, navigation, and shared-autonomy tasks.

Despite these strengths, several limitations remain:

\noindent \textbf{Dependence on imitation data and distributional bias.} RT-V2 is trained to replicate user-like behavior by imitating intended trajectories. Collecting a large corpus of real user demonstrations (with failures labeled and removed) is expensive and often infeasible; in this work we used motion-planner trajectories as a practical proxy. Planner-generated trajectories share properties with human/animal behavior but are not identical, which can introduce a distributional mismatch: RT-V2 may disallow movements that a real user would produce. Future work should quantify this gap (e.g., compare curvature, speed profiles, and goal-switching statistics) and explore mitigations such as domain adaptation, data augmentation, inverse-reinforcement learning (to learn preferences from sparse demonstrations), or active learning that queries the user for corrective examples.

\noindent \textbf{Limited modeling of user adaptation.} Our current implementation treats the user model as static: RT-V2 assumes user behavior does not change in response to assistance. In practice, users adapt their behavior when assisted, causing covariate shifts that the current model does not capture. Future research should incorporate online or continual learning methods (Bayesian nonstationary models, meta-learning, or online policy refinement) so the prior can adapt to evolving user strategies while preserving safety.

\noindent \textbf{Scope of evaluation.} Evaluations were limited to relatively simple navigation scenarios (simulator-based and a set of shared-autonomy tasks with keyboard and BMI inputs). To establish broader applicability, RT-V2 should be tested on real mobile platforms, in denser and more dynamic environments (moving obstacles, multiple agents), and with diverse users. Besides, it is also promising to extend RT-V2's framework to $SE(3)$ robotic manipulation tasks, where the goals can be represented with grasp poses generated by grasp planners \citep{song2024implicit,song2025equivariant}. We will leave those works for the future.

\section{Conclusion}
In this work, we presented \emph{Robot Trajectron V2} (RT-V2), a Bayesian shared-control framework for navigation that models user intent via a learned prior and computes a posterior over candidate actions to produce safe, goal-directed assistance. RT-V2 combines imitation-based priors, posterior inference that accounts for interface uncertainty, and an Imagined-Rollout reinforcement step to refine policies using simulated interaction and reward feedback. A sampling-based controller then selects collision-free trajectories by maximizing the posterior under safety constraints.

Experimental results show that RT-V2 yields accurate trajectory predictions and improves safe navigation across simulated and shared-autonomy scenarios. In human and BMI experiments (keyboard and invasive BMI; human subjects and monkeys), RT-V2 provided smoother, safer assistance and increased task success under static and appearing-obstacle conditions. We also identified a tradeoff between assistance and flexibility during abrupt intent changes, and we discussed strategies to mitigate it.

Future extensions include adaptive online learning of user behavior, evaluations on real robotic platforms in dynamic environments, and formalizing robustness guarantees. We believe RT-V2 provides a principled probabilistic foundation for shared autonomy and a practical starting point for systems that robustly assist users across diverse interfaces and tasks.

\section*{Appendix 1: The relation between the goal-conditioned intention policy and the twin policy}
To be specific, $\pi_T(\bm{a}|\bm{s}) = p(\bm{a}|\bm{s},T)$, $\pi_I(\bm{a}|\bm{s}, \bm{g}) = p(\bm{a}|\bm{s},\bm{g},I)$, and $\pi_I(\bm{a}|\bm{s}) = p(\bm{a}|\bm{s},I)$. Proof:
\begin{equation}
\begin{aligned}
    & \mathcal{D}_{\textnormal{TV}}(\pi_T(\cdot|\bm{s}), \pi_I(\cdot|\bm{s}, \bm{g})) \\
    & =  \int_{\bm{a}} |\pi_T(\bm{a}|\bm{s})- \pi_I(\bm{a}|\bm{s}, \bm{g})| \mathrm{d} \bm{a} \\
    & =   \int_{\bm{a}} |\pi_T(\bm{a}|\bm{s})- \pi_I(\bm{a}|\bm{s}) + \pi_I(\bm{a}|\bm{s}) - \pi_I(\bm{a}|\bm{s}, \bm{g})| \mathrm{d} \bm{a} \\
    & \leq  \int_{\bm{a}} |\pi_T(\bm{a}|\bm{s})- \pi_I(\bm{a}|\bm{s})| + |\pi_I(\bm{a}|\bm{s}) - \pi_I(\bm{a}|\bm{s}, \bm{g})| \mathrm{d} \bm{a} \\
     & =  \int_{\bm{a}} |\pi_T(\bm{a}|\bm{s})- \pi_I(\bm{a}|\bm{s})|\mathrm{d} \bm{a} + \int_{\bm{a}} |\pi_I(\bm{a}|\bm{s}) - \pi_I(\bm{a}|\bm{s}, \bm{g})| \mathrm{d} \bm{a} \\
    & =   \mathcal{D}_{\textnormal{TV}}(\pi_T(\cdot|\bm{s}), \pi_I(\cdot|\bm{s})) + \mathcal{D}_{\textnormal{TV}}(\pi_I(\cdot|\bm{s}), \pi_I(\cdot|\bm{s}, \bm{g})). \label{eq: appendix 1-1}
\end{aligned}
\end{equation}
Due to Pinsker's inequality, the last term of Eq.~\ref{eq: appendix 1-1} can be bounded by:
\begin{equation}
\footnotesize
    \begin{aligned}
        & \mathcal{D}_{\textnormal{TV}}(\pi_T(\cdot|\bm{s}), \pi_I(\cdot|\bm{s})) + \mathcal{D}_{\textnormal{TV}}(\pi_I(\cdot|\bm{s}), \pi_I(\cdot|\bm{s}, \bm{g})) \\
        \leq & \sqrt{2\mathcal{D}_{\textnormal{KL}}(\pi_T(\cdot|\bm{s}), \pi_I(\cdot|\bm{s}))} + \mathcal{D}_{\textnormal{TV}}(\pi_I(\cdot|\bm{s}), \pi_I(\cdot|\bm{s}, \bm{g}))
    \end{aligned}
\end{equation}

\section*{Appendix 2: The Upper Bound and Lower Bound of Entropy of Gaussian Mixture Models}
Gaussian Mixture models can be written as:
\begin{equation}
    g(x) = \sum_{i=1}^{M} \alpha_i \mathcal{N}(\bm{x}|\bm{\mu}_i, \bm{\Sigma}_i).
\end{equation}
\noindent \textbf{Upper bound:}
\begin{equation}
\begin{aligned}
    H(g) & =   \int~g(\bm{x}) \textnormal{log}~g(\bm{x}) \mathrm{d}\bm{x}\\
     & =   -  \int~\sum_{i=1}^{M} \alpha_i \mathcal{N}(\bm{x}|\bm{\mu}_i, \bm{\Sigma}_i) \\
     & ~~~~~~~\cdot \textnormal{log}~(\sum_{j=1}^{M} \alpha_j \mathcal{N}(\bm{x}|\bm{\mu}_j, \bm{\Sigma}_j)) \mathrm{d}\bm{x} \\
     & =   -  \sum_{i=1}^{M} \alpha_i \int~\mathcal{N}(\bm{x}|\bm{\mu}_i, \bm{\Sigma}_i) \\
     & ~~~~~~~\cdot \textnormal{log}~(\alpha_i \mathcal{N}(\bm{x}|\bm{\mu}_i, \bm{\Sigma}_i)\cdot (1+\epsilon_i)) \mathrm{d}\bm{x} \\
      & =   -  \sum_{i=1}^{M} \alpha_i \int~\mathcal{N}(\bm{x}|\bm{\mu}_i, \bm{\Sigma}_i) \\
     & ~~~~~~~\cdot (\textnormal{log}~(\alpha_i \mathcal{N}(\bm{x}|\bm{\mu}_i, \bm{\Sigma}_i))+ \textnormal{log}~(1+\epsilon_i)) \mathrm{d}\bm{x}, \label{eq: upper bound1}
\end{aligned}
\end{equation}
where
\begin{equation}
\epsilon_i = \frac{\sum_{i \neq j=1}^M\alpha_j \cdot \mathcal{N}(\bm{x}|\bm{\mu}_j, \bm{\Sigma}_j)}{\alpha_i \cdot \mathcal{N}(\bm{x}|\bm{\mu}_i, \bm{\Sigma}_i)}
\end{equation}
Since $\textnormal{log}~(1+\epsilon_i)$ in Eq.~\ref{eq: upper bound1} is always non-negative, neglecting it yields the upper bound, as:
\begin{equation}
\begin{aligned}
    H(g) & \leq -\sum_{i=1}^{M} \alpha_i \int~\mathcal{N}(\bm{x}|\bm{\mu}_i, \bm{\Sigma}_i) \\
     & ~~~~~~~\cdot (\textnormal{log}~(\alpha_i \mathcal{N}(\bm{x}|\bm{\mu}_i, \bm{\Sigma}_i))) \mathrm{d}\bm{x},\\
     & = \sum_{i=1}^M \alpha_i \cdot (-\textnormal{log}~\alpha_i+H(\mathcal{N}(\cdot|\bm{\mu}_i, \bm{\Sigma}_i)))\\
     & = \sum_{i=1}^M \alpha_i \cdot (-\textnormal{log}~\alpha_i+\frac{1}{2}\textnormal{log}~((2\pi e)^N|\bm{\Sigma}_i|)),
\end{aligned}
\end{equation}
where $N$ is the dimension of Multivariate Gaussian distribution.

\noindent \textbf{Lower bound:}
\begin{equation}
\begin{aligned}
    H(g) & =   \int~g(\bm{x}) \textnormal{log}~g(\bm{x}) d\bm{x}\\
     & =   - \sum_{i=1}^{M} \alpha_i \cdot \int~\mathcal{N}(\bm{x}|\bm{\mu}_i, \bm{\Sigma}_i) \cdot \textnormal{log}~g(\bm{x}) d\bm{x} \\
    \geq & - \sum_{i=1}^{M} \alpha_i \cdot \textnormal{log}~\int~\mathcal{N}(\bm{x}|\bm{\mu}_i, \bm{\Sigma}_i) \cdot g(\bm{x}) d\bm{x} \\
    & =   - \sum_{i=1}^{M} \alpha_i \cdot \textnormal{log}~\sum_{j=1}^{M} \alpha_j e_{i,j}.
\end{aligned}
\end{equation}
where,
\begin{align}
    & e_{i,j} = \int~\mathcal{N}(\bm{x}|\bm{\mu}_i, \bm{\Sigma}_i) \cdot \mathcal{N}(\bm{x}|\bm{\mu}_j, \bm{\Sigma}_j) d\bm{x} \\
    & ~~~~= \mathcal{N}(\bm{\mu}_i|\bm{\mu}_j, \bm{\Sigma}_i+\bm{\Sigma}_j)  \nonumber
\end{align}

\section*{Appendix 3: The Derivation from Eq.~\ref{eq: Q* KL} to Eq.~\ref{eq: optimal action}}
First, we can analytically express $\frac{Q(\bm{A})}{P(\bm{A})}$ in Eq.~\ref{eq: Q* KL} as follows:
\begin{equation}
    \frac{Q(\bm{A})}{P(\bm{A})} = \mathop{\textnormal{exp}}(\sum_{t=0}^{H-1}-\frac{1}{2}\bm{u}_t^T\bm{\Omega}^{-1}_{t}\bm{u}_t + \bm{u}_t^T\bm{\Omega}^{-1}_{t}\bm{a}_t). \label{eq: q_div_p}
\end{equation}
Inserting this into Eq.~\ref{eq: Q* KL} yields:
\begin{equation}
\begin{aligned}
    & \mathcal{D}_{\textnormal{KL}}(Q^*||Q) \\
    & = \int Q^*(\bm{A})(\sum_{t=0}^{H-1}-\frac{1}{2}\bm{u}_t^T\bm{\Omega}^{-1}_{t}\bm{u}_t + \bm{u}_t^T\bm{\Omega}^{-1}_{t}\bm{a}_t)\mathrm{d}(\bm{A}) \\
    & =\sum_{t=0}^{H-1}(-\frac{1}{2}\bm{u}_t^T\bm{\Omega}_t^{-1}\bm{u}_t + \bm{u}_t^T \int Q^*(\bm{A})\bm{\Omega}_t^{-1}\bm{a}_t \mathrm{d} \bm{A}). \label{eq: Q* KL2}
\end{aligned}
\end{equation}
This is concave with respect to each $\bm{\mu}_t$. Thus, we can take the gradient of Eq.~\ref{eq: Q* KL2} and set it to zero to obtain the maximum, which yields:
\begin{equation}
\begin{aligned}
    &\frac{\mathrm{d} \mathcal{D}_{\textnormal{KL}}(Q^*||Q)}{\mathrm{d} \bm{u}_t} = 0\\
    &\Rightarrow \bm{u}_t^* = \int~Q^*(\bm{A})\bm{a}_t \mathrm{d}\bm{A}. \label{eq: optimal action2}
\end{aligned}
\end{equation}

\section*{Appendix 4: The implementation of Baseline in Human-Exp}
We choose Hindsight Optimization (HO) \citep{javdani2018shared} with APFs \citep{apf} as our baseline. HO is a well-recognized intent estimator widely used in many human-robot interaction and cooperation tasks \citep{yang2021review,selvaggio2021autonomy}. We leverage APFs to control and navigate. APFs are a popular method for navigation \citep{weerakoon2015artificial,sudhakara2018obstacle}. Besides, since APFs can be written as a form of constraint-based shared control \citep{iregui2021reconfigurable,sct}, APFs share some common characteristics with them, which means that the performance obtained by the baseline is representative to some extent.

HO considers the past position trajectories and the current user's command to infer the intent estimation probabilities $\{p_m\}_{m=1}^{M}$ for all the potential goals $\{\bm{g}_m\}_{m=1}^{M}$. In the implementation of APFs, we design a repulsive field to navigate around the obstacles and an attractive field to reach the intended goal. The repulsive field is written as:
\begin{equation}
    U_r(\bm{p})=\begin{cases}
    &w_{r} (\frac{1}{\rho(\bm{p})}-\frac{1}{\rho_0})^2,\ \text{if}~ \rho(\bm{p})\leq  \rho_0\\
    & 0,\ \ \  \ \ \ \ \ \ \ \ \ \ \ \ \ \ \ \text{if} \rho(\bm{p}) >  \rho_0,
     \end{cases} 
\end{equation}
where $w_r$ is the repulsive weight, $\rho(\bm{p})$ denotes the distance between the current robot position $\bm{p}$ and the closest obstacle, and $\rho_0$ denotes the influence ray of the obstacle.
The attractive field is written as:
\begin{equation}
    U_a(\bm{p}) = w_a \sum_{m=1}^{M} p_{m} \parallel \bm{g}_m - \bm{p} \parallel,
\end{equation}
where $w_a$ is the attraction weight. The repulsive field and the attractive field apply a force equal to the negative gradient of the potential. 
Combining two fields with the current user's command $\bm{v}_u$, we define the robot velocity $\bm{v}$ as:
\begin{equation}
\begin{aligned}
    \bm{v} & = \bm{v}_{u} - (\nabla U_a(\bm{p})+\nabla U_r(\bm{p})) \\
    & = \bm{v}_{u} + w_a  \sum_{m=1}^{M} p_{m} \frac{\bm{g}_m-\bm{p}}{\parallel \bm{g}_m-\bm{p} \parallel} \\ 
    &~~~~~~~~~~+\frac{w_{r}}{\rho^2(\bm{p})} (\frac{1}{\rho(\bm{p})}-\frac{1}{\rho_0}) \nabla \rho(\bm{p}) \\
    & = \bm{v}_{u} + w_a \bm{v}_{g} + w_{r} \bm{v}_{r}, \label{apf_share_control}
\end{aligned}
\end{equation}
In each iteration, given the user's command $\bm{v}_u$, we calculate the robot velocity $\bm{v}$ and apply it to the robot. The robot velocity $\bm{v}$ not only considers the user's command, but also the predicted goals with uncertainty and obstacle avoidance, which realizes shared control.

\bibliographystyle{SageH}

\bibliography{ref}

\begin{thebibliography}{67}
\providecommand{\natexlab}[1]{#1}
\providecommand{\url}[1]{\texttt{#1}}
\providecommand{\urlprefix}{URL }
\expandafter\ifx\csname urlstyle\endcsname\relax
  \providecommand{\doi}[1]{DOI:\discretionary{}{}{}#1}\else
  \providecommand{\doi}{DOI:\discretionary{}{}{}\begingroup \urlstyle{rm}\Url}\fi

\bibitem[{Aigner and McCarragher(2000)}]{aigner2000modeling}
Aigner P and McCarragher BJ (2000) Modeling and constraining human interactions in shared control utilizing a discrete event framework.
\newblock \emph{IEEE Transactions on Systems, Man, and Cybernetics-Part A: Systems and Humans} 30(3): 369--379.

\bibitem[{Backman et~al.(2023)Backman, Kuli{\'c} and Chung}]{backman2023reinforcement}
Backman K, Kuli{\'c} D and Chung H (2023) Reinforcement learning for shared autonomy drone landings.
\newblock \emph{Autonomous Robots} 47(8): 1419--1438.

\bibitem[{Bhardwaj et~al.(2022)Bhardwaj, Sundaralingam, Mousavian, Ratliff, Fox, Ramos and Boots}]{bhardwaj2022storm}
Bhardwaj M, Sundaralingam B, Mousavian A, Ratliff ND, Fox D, Ramos F and Boots B (2022) Storm: An integrated framework for fast joint-space model-predictive control for reactive manipulation.
\newblock In: \emph{Conference on Robot Learning}. PMLR, pp. 750--759.

\bibitem[{Demeester et~al.(2008)Demeester, H{\"u}ntemann, Vanhooydonck, Vanacker, Van~Brussel and Nuttin}]{demeester2008user}
Demeester E, H{\"u}ntemann A, Vanhooydonck D, Vanacker G, Van~Brussel H and Nuttin M (2008) User-adapted plan recognition and user-adapted shared control: A bayesian approach to semi-autonomous wheelchair driving.
\newblock \emph{Autonomous Robots} 24: 193--211.

\bibitem[{Dragan and Srinivasa(2013)}]{policyblending}
Dragan AD and Srinivasa SS (2013) A policy-blending formalism for shared control.
\newblock \emph{The International Journal of Robotics Research} 32(7): 790--805.

\bibitem[{Ezeh et~al.(2017{\natexlab{a}})Ezeh, Trautman, Devigne, Bureau, Babel and Carlson}]{pscwheel}
Ezeh C, Trautman P, Devigne L, Bureau V, Babel M and Carlson T (2017{\natexlab{a}}) Probabilistic vs linear blending approaches to shared control for wheelchair driving.
\newblock In: \emph{2017 International Conference on Rehabilitation Robotics (ICORR)}. IEEE, pp. 835--840.

\bibitem[{Ezeh et~al.(2017{\natexlab{b}})Ezeh, Trautman, Devigne, Bureau, Babel and Carlson}]{ezeh2017probabilistic}
Ezeh C, Trautman P, Devigne L, Bureau V, Babel M and Carlson T (2017{\natexlab{b}}) Probabilistic vs linear blending approaches to shared control for wheelchair driving.
\newblock In: \emph{2017 International Conference on Rehabilitation Robotics (ICORR)}. IEEE, pp. 835--840.

\bibitem[{Ezeh et~al.(2017{\natexlab{c}})Ezeh, Trautman, Holloway and Carlson}]{psccompare}
Ezeh C, Trautman P, Holloway C and Carlson T (2017{\natexlab{c}}) Comparing shared control approaches for alternative interfaces: A wheelchair simulator experiment.
\newblock In: \emph{2017 IEEE International Conference on Systems, Man, and Cybernetics (SMC)}. IEEE, pp. 93--98.

\bibitem[{Fang et~al.(2019)Fang, Jia, Guo, Xu, Wen and Sun}]{ILsurvey}
Fang B, Jia S, Guo D, Xu M, Wen S and Sun F (2019) Survey of imitation learning for robotic manipulation.
\newblock \emph{International Journal of Intelligent Robotics and Applications} 3: 362--369.

\bibitem[{Fu et~al.(2025)Fu, Song, Hu and Detry}]{fu2025a}
Fu Z, Song P, Hu Y and Detry R (2025) Tasc: Task-aware shared control for teleoperated manipulation.
\newblock \urlprefix\url{https://arxiv.org/abs/2509.10416}.

\bibitem[{Goodrich and Olsen(2003)}]{goodrich2003seven}
Goodrich MA and Olsen DR (2003) Seven principles of efficient human robot interaction.
\newblock In: \emph{SMC'03 Conference Proceedings. 2003 IEEE International Conference on Systems, Man and Cybernetics. Conference Theme-System Security and Assurance (Cat. No. 03CH37483)}, volume~4. IEEE, pp. 3942--3948.

\bibitem[{Gottardi et~al.(2022)Gottardi, Tortora, Tosello and Menegatti}]{gottardi2022shared}
Gottardi A, Tortora S, Tosello E and Menegatti E (2022) Shared control in robot teleoperation with improved potential fields.
\newblock \emph{IEEE Transactions on Human-Machine Systems} 52(3): 410--422.

\bibitem[{Gu et~al.(2022)Gu, Chen, Li, Lin, Rao, Zhou and Lu}]{mid}
Gu T, Chen G, Li J, Lin C, Rao Y, Zhou J and Lu J (2022) Stochastic trajectory prediction via motion indeterminacy diffusion.
\newblock In: \emph{Proceedings of the IEEE/CVF Conference on Computer Vision and Pattern Recognition}. pp. 17113--17122.

\bibitem[{Ha and Schmidhuber(2018)}]{ha2018world}
Ha D and Schmidhuber J (2018) World models.
\newblock \emph{arXiv preprint arXiv:1803.10122} .

\bibitem[{Haarnoja et~al.(2018)Haarnoja, Zhou, Abbeel and Levine}]{sac}
Haarnoja T, Zhou A, Abbeel P and Levine S (2018) Soft actor-critic: Off-policy maximum entropy deep reinforcement learning with a stochastic actor.
\newblock In: \emph{International conference on machine learning}. PMLR, pp. 1861--1870.

\bibitem[{Hafner et~al.(2020)Hafner, Lillicrap, Ba and Norouzi}]{dreamer}
Hafner D, Lillicrap T, Ba J and Norouzi M (2020) Dream to control: Learning behaviors by latent imagination.
\newblock In: \emph{International Conference on Learning Representations}.

\bibitem[{Hansen et~al.(2022)Hansen, Wang and Su}]{tdmpc}
Hansen N, Wang X and Su H (2022) Temporal difference learning for model predictive control.
\newblock In: \emph{International Conference on Machine Learning, PMLR}.

\bibitem[{Hart(1988)}]{nasatlx}
Hart S (1988) Development of nasa-tlx (task load index): Results of empirical and theoretical research.
\newblock \emph{Human mental workload/Elsevier} .

\bibitem[{Haviland and Corke(2020)}]{haviland2020purely}
Haviland J and Corke P (2020) A purely-reactive manipulability-maximising motion controller.
\newblock \emph{arXiv preprint arXiv:2002.11901} .

\bibitem[{Higgins et~al.(2016)Higgins, Matthey, Pal, Burgess, Glorot, Botvinick, Mohamed and Lerchner}]{beta-vae}
Higgins I, Matthey L, Pal A, Burgess C, Glorot X, Botvinick M, Mohamed S and Lerchner A (2016) beta-vae: Learning basic visual concepts with a constrained variational framework.
\newblock In: \emph{International conference on learning representations}.

\bibitem[{Ho and Ermon(2016)}]{gail}
Ho J and Ermon S (2016) Generative adversarial imitation learning.
\newblock \emph{Advances in neural information processing systems} 29.

\bibitem[{Hochberg et~al.(2012)Hochberg, Bacher, Jarosiewicz, Masse, Simeral, Vogel, Haddadin, Liu, Cash, Van Der~Smagt et~al.}]{hochberg2012reach}
Hochberg LR, Bacher D, Jarosiewicz B, Masse NY, Simeral JD, Vogel J, Haddadin S, Liu J, Cash SS, Van Der~Smagt P et~al. (2012) Reach and grasp by people with tetraplegia using a neurally controlled robotic arm.
\newblock \emph{Nature} 485(7398): 372--375.

\bibitem[{Honda et~al.(2023)Honda, Akai, Suzuki, Aoki, Hosogaya, Okuda and Suzuki}]{honda2023stein}
Honda K, Akai N, Suzuki K, Aoki M, Hosogaya H, Okuda H and Suzuki T (2023) Stein variational guided model predictive path integral control: Proposal and experiments with fast maneuvering vehicles.
\newblock \emph{arXiv preprint arXiv:2309.11040} .

\bibitem[{Hu et~al.(2022)Hu, Corrado, Griffiths, Murez, Gurau, Yeo, Kendall, Cipolla and Shotton}]{mile}
Hu A, Corrado G, Griffiths N, Murez Z, Gurau C, Yeo H, Kendall A, Cipolla R and Shotton J (2022) Model-based imitation learning for urban driving.
\newblock \emph{Advances in Neural Information Processing Systems} 35: 20703--20716.

\bibitem[{Huber et~al.(2008)Huber, Bailey, Durrant-Whyte and Hanebeck}]{huber2008entropy}
Huber MF, Bailey T, Durrant-Whyte H and Hanebeck UD (2008) On entropy approximation for gaussian mixture random vectors.
\newblock In: \emph{2008 IEEE International Conference on Multisensor Fusion and Integration for Intelligent Systems}. IEEE, pp. 181--188.

\bibitem[{Iregui et~al.(2021)Iregui, De~Schutter and Aertbeli{\"e}n}]{iregui2021reconfigurable}
Iregui S, De~Schutter J and Aertbeli{\"e}n E (2021) Reconfigurable constraint-based reactive framework for assistive robotics with adaptable levels of autonomy.
\newblock \emph{IEEE Robotics and Automation Letters} 6(4): 7397--7405.

\bibitem[{Ivanovic et~al.(2020)Ivanovic, Leung, Schmerling and Pavone}]{ivanovic2020multimodal}
Ivanovic B, Leung K, Schmerling E and Pavone M (2020) Multimodal deep generative models for trajectory prediction: A conditional variational autoencoder approach.
\newblock \emph{IEEE Robotics and Automation Letters} 6(2): 295--302.

\bibitem[{Ivanovic and Pavone(2019)}]{trajectron}
Ivanovic B and Pavone M (2019) The trajectron: Probabilistic multi-agent trajectory modeling with dynamic spatiotemporal graphs.
\newblock In: \emph{Proceedings of the IEEE/CVF International Conference on Computer Vision}. pp. 2375--2384.

\bibitem[{Javdani et~al.(2018)Javdani, Admoni, Pellegrinelli, Srinivasa and Bagnell}]{javdani2018shared}
Javdani S, Admoni H, Pellegrinelli S, Srinivasa SS and Bagnell JA (2018) Shared autonomy via hindsight optimization for teleoperation and teaming.
\newblock \emph{The International Journal of Robotics Research} 37(7): 717--742.

\bibitem[{Kavraki et~al.(1996)Kavraki, Svestka, Latombe and Overmars}]{kavraki1996probabilistic}
Kavraki LE, Svestka P, Latombe JC and Overmars MH (1996) Probabilistic roadmaps for path planning in high-dimensional configuration spaces.
\newblock \emph{IEEE transactions on Robotics and Automation} 12(4): 566--580.

\bibitem[{Khatib(1986)}]{apf}
Khatib O (1986) Real-time obstacle avoidance for manipulators and mobile robots.
\newblock \emph{The international journal of robotics research} 5(1): 90--98.

\bibitem[{Kolev et~al.(2024)Kolev, Rafailov, Hatch, Wu and Finn}]{kolev2024efficient}
Kolev V, Rafailov R, Hatch K, Wu J and Finn C (2024) Efficient imitation learning with conservative world models.
\newblock \emph{arXiv preprint arXiv:2405.13193} .

\bibitem[{Konda and Tsitsiklis(1999)}]{ac}
Konda V and Tsitsiklis J (1999) Actor-critic algorithms.
\newblock \emph{Advances in neural information processing systems} 12.

\bibitem[{Le~Mero et~al.(2022)Le~Mero, Yi, Dianati and Mouzakitis}]{ILsurveyDriving}
Le~Mero L, Yi D, Dianati M and Mouzakitis A (2022) A survey on imitation learning techniques for end-to-end autonomous vehicles.
\newblock \emph{IEEE Transactions on Intelligent Transportation Systems} 23(9): 14128--14147.

\bibitem[{Lei et~al.(2022)Lei, Tan, Garg, Li, Sidarta and Ang}]{lei2022intention}
Lei Z, Tan BY, Garg NP, Li L, Sidarta A and Ang WT (2022) An intention prediction based shared control system for point-to-point navigation of a robotic wheelchair.
\newblock \emph{IEEE Robotics and Automation Letters} 7(4): 8893--8900.

\bibitem[{Lu et~al.(2019)Lu, Bi and Li}]{lu2019model}
Lu Y, Bi L and Li H (2019) Model predictive-based shared control for brain-controlled driving.
\newblock \emph{IEEE Transactions on Intelligent Transportation Systems} 21(2): 630--640.

\bibitem[{Luo et~al.(2024)Luo, Peng, Lv, Hong, Driggs-Campbell, Lu and Li}]{luo2024human}
Luo S, Peng Q, Lv J, Hong K, Driggs-Campbell KR, Lu C and Li YL (2024) Human-agent joint learning for efficient robot manipulation skill acquisition.
\newblock \emph{arXiv preprint arXiv:2407.00299} .

\bibitem[{Maeda(2022)}]{maeda2022blending}
Maeda G (2022) Blending primitive policies in shared control for assisted teleoperation.
\newblock In: \emph{2022 International Conference on Robotics and Automation (ICRA)}. IEEE, pp. 9332--9338.

\bibitem[{Muelling et~al.(2017)Muelling, Venkatraman, Valois, Downey, Weiss, Javdani, Hebert, Schwartz, Collinger and Bagnell}]{muelling2017autonomy}
Muelling K, Venkatraman A, Valois JS, Downey JE, Weiss J, Javdani S, Hebert M, Schwartz AB, Collinger JL and Bagnell JA (2017) Autonomy infused teleoperation with application to brain computer interface controlled manipulation.
\newblock \emph{Autonomous Robots} 41: 1401--1422.

\bibitem[{Oh et~al.(2020)Oh, Wu, Toussaint and Mainprice}]{oh2020natural}
Oh Y, Wu SW, Toussaint M and Mainprice J (2020) Natural gradient shared control.
\newblock In: \emph{2020 29th IEEE International Conference on Robot and Human Interactive Communication (RO-MAN)}. IEEE, pp. 1223--1229.

\bibitem[{Padalkar et~al.(2023)Padalkar, Pooley, Jain, Bewley, Herzog, Irpan, Khazatsky, Rai, Singh, Brohan et~al.}]{rtx}
Padalkar A, Pooley A, Jain A, Bewley A, Herzog A, Irpan A, Khazatsky A, Rai A, Singh A, Brohan A et~al. (2023) Open x-embodiment: Robotic learning datasets and rt-x models.
\newblock \emph{arXiv preprint arXiv:2310.08864} .

\bibitem[{Quere et~al.(2020)Quere, Hagengruber, Iskandar, Bustamante, Leidner, Stulp and Vogel}]{sct}
Quere G, Hagengruber A, Iskandar M, Bustamante S, Leidner D, Stulp F and Vogel J (2020) Shared control templates for assistive robotics.
\newblock In: \emph{2020 IEEE international conference on robotics and automation (ICRA)}. IEEE, pp. 1956--1962.

\bibitem[{Rastgar et~al.(2024)Rastgar, Masnavi, Sharma, Aabloo, Swevers and Singh}]{rastgar2024priest}
Rastgar F, Masnavi H, Sharma B, Aabloo A, Swevers J and Singh AK (2024) Priest: Projection guided sampling-based optimization for autonomous navigation.
\newblock \emph{IEEE Robotics and Automation Letters} .

\bibitem[{Reddy et~al.(2018)Reddy, Dragan and Levine}]{reddy2018shared}
Reddy S, Dragan AD and Levine S (2018) Shared autonomy via deep reinforcement learning.
\newblock \emph{arXiv preprint arXiv:1802.01744} .

\bibitem[{Ross et~al.(2011)Ross, Gordon and Bagnell}]{dagger}
Ross S, Gordon G and Bagnell D (2011) A reduction of imitation learning and structured prediction to no-regret online learning.
\newblock In: \emph{Proceedings of the fourteenth international conference on artificial intelligence and statistics}. JMLR Workshop and Conference Proceedings, pp. 627--635.

\bibitem[{Rulik et~al.(2022)Rulik, Sunny, Sanjuan De~Caro, Zarif, Brahmi, Ahamed, Schultz, Wang, Leheng, Longxiang et~al.}]{rulik2022control}
Rulik I, Sunny MSH, Sanjuan De~Caro JD, Zarif MII, Brahmi B, Ahamed SI, Schultz K, Wang I, Leheng T, Longxiang JP et~al. (2022) Control of a wheelchair-mounted 6dof assistive robot with chin and finger joysticks.
\newblock \emph{Frontiers in Robotics and AI} 9: 885610.

\bibitem[{Salzmann et~al.(2020)Salzmann, Ivanovic, Chakravarty and Pavone}]{trajectron++}
Salzmann T, Ivanovic B, Chakravarty P and Pavone M (2020) Trajectron++: Dynamically-feasible trajectory forecasting with heterogeneous data.
\newblock In: \emph{Computer Vision--ECCV 2020: 16th European Conference, Glasgow, UK, August 23--28, 2020, Proceedings, Part XVIII 16}. Springer, pp. 683--700.

\bibitem[{Saussus et~al.(2025)Saussus, De~Schrijver, Garcia~Ramirez, Decramer and Janssen}]{saussus2025intracortical}
Saussus O, De~Schrijver S, Garcia~Ramirez J, Decramer T and Janssen P (2025) An intracortical brain-computer interface for navigation in virtual reality in macaque monkeys.
\newblock \emph{bioRxiv} : 2025--05.

\bibitem[{Schaff and Walter(2020)}]{schaff2020residual}
Schaff C and Walter MR (2020) Residual policy learning for shared autonomy.
\newblock \emph{arXiv preprint arXiv:2004.05097} .

\bibitem[{Schulman et~al.(2015)Schulman, Moritz, Levine, Jordan and Abbeel}]{gae}
Schulman J, Moritz P, Levine S, Jordan M and Abbeel P (2015) High-dimensional continuous control using generalized advantage estimation.
\newblock \emph{arXiv preprint arXiv:1506.02438} .

\bibitem[{Selvaggio et~al.(2021)Selvaggio, Cognetti, Nikolaidis, Ivaldi and Siciliano}]{selvaggio2021autonomy}
Selvaggio M, Cognetti M, Nikolaidis S, Ivaldi S and Siciliano B (2021) Autonomy in physical human-robot interaction: A brief survey.
\newblock \emph{IEEE Robotics and Automation Letters} 6(4): 7989--7996.

\bibitem[{Singh and Heard(2023)}]{singh2023probabilistic}
Singh S and Heard J (2023) Probabilistic policy blending for shared autonomy using deep reinforcement learning.
\newblock In: \emph{2023 32nd IEEE International Conference on Robot and Human Interactive Communication (RO-MAN)}. IEEE, pp. 1537--1544.

\bibitem[{Sohn et~al.(2015)Sohn, Lee and Yan}]{cvae}
Sohn K, Lee H and Yan X (2015) Learning structured output representation using deep conditional generative models.
\newblock \emph{Advances in neural information processing systems} 28.

\bibitem[{Song et~al.(2020)Song, Ding, Chen, Shen, Wang and Chen}]{song2020pip}
Song H, Ding W, Chen Y, Shen S, Wang MY and Chen Q (2020) Pip: Planning-informed trajectory prediction for autonomous driving.
\newblock In: \emph{Computer Vision--ECCV 2020: 16th European Conference, Glasgow, UK, August 23--28, 2020, Proceedings, Part XXI 16}. Springer, pp. 598--614.

\bibitem[{Song et~al.(2025)Song, Hu, Li and Detry}]{song2025equivariant}
Song P, Hu Y, Li P and Detry R (2025) Equivariant volumetric grasping.
\newblock \emph{arXiv preprint arXiv:2507.18847} .

\bibitem[{Song et~al.(2024{\natexlab{a}})Song, Li, Aertbeli{\"e}n and Detry}]{RT}
Song P, Li P, Aertbeli{\"e}n E and Detry R (2024{\natexlab{a}}) Robot trajectron: Trajectory prediction-based shared control for robot manipulation.
\newblock In: \emph{Proceedings Of IEEE International Conference on Robotics and Automation}.

\bibitem[{Song et~al.(2024{\natexlab{b}})Song, Li and Detry}]{song2024implicit}
Song P, Li P and Detry R (2024{\natexlab{b}}) Implicit grasp diffusion: Bridging the gap between dense prediction and sampling-based grasping.
\newblock In: \emph{8th Annual Conference on Robot Learning}.

\bibitem[{Sudhakara et~al.(2018)Sudhakara, Ganapathy, Priyadharshini and Sundaran}]{sudhakara2018obstacle}
Sudhakara P, Ganapathy V, Priyadharshini B and Sundaran K (2018) Obstacle avoidance and navigation planning of a wheeled mobile robot using amended artificial potential field method.
\newblock \emph{Procedia computer science} 133: 998--1004.

\bibitem[{Trautman(2015)}]{psc}
Trautman P (2015) Assistive planning in complex, dynamic environments: a probabilistic approach.
\newblock In: \emph{2015 IEEE International Conference on Systems, Man, and Cybernetics}. IEEE, pp. 3072--3078.

\bibitem[{Vanhooydonck et~al.(2003)Vanhooydonck, Demeester, Nuttin and Van~Brussel}]{vanhooydonck2003shared}
Vanhooydonck D, Demeester E, Nuttin M and Van~Brussel H (2003) Shared control for intelligent wheelchairs: an implicit estimation of the user intention.
\newblock In: \emph{Proceedings of the 1st international workshop on advances in service robotics (ASER’03)}. pp. 176--182.

\bibitem[{Weerakoon et~al.(2015)Weerakoon, Ishii and Nassiraei}]{weerakoon2015artificial}
Weerakoon T, Ishii K and Nassiraei AAF (2015) An artificial potential field based mobile robot navigation method to prevent from deadlock.
\newblock \emph{Journal of Artificial Intelligence and Soft Computing Research} 5(3): 189--203.

\bibitem[{Williams et~al.(2017)Williams, Wagener, Goldfain, Drews, Rehg, Boots and Theodorou}]{williams2017information}
Williams G, Wagener N, Goldfain B, Drews P, Rehg JM, Boots B and Theodorou EA (2017) Information theoretic mpc for model-based reinforcement learning.
\newblock In: \emph{2017 IEEE international conference on robotics and automation (ICRA)}. IEEE, pp. 1714--1721.

\bibitem[{Wu et~al.(2023)Wu, Zhou, Yang, Huang and Lv}]{wu2023human}
Wu J, Zhou Y, Yang H, Huang Z and Lv C (2023) Human-guided reinforcement learning with sim-to-real transfer for autonomous navigation.
\newblock \emph{IEEE Transactions on Pattern Analysis and Machine Intelligence} .

\bibitem[{Xu et~al.(2020)Xu, Zhang, Cao, Shu and Zhang}]{xu2020shared}
Xu Y, Zhang H, Cao L, Shu X and Zhang D (2020) A shared control strategy for reach and grasp of multiple objects using robot vision and noninvasive brain--computer interface.
\newblock \emph{IEEE Transactions on Automation Science and Engineering} 19(1): 360--372.

\bibitem[{Yang et~al.(2021)Yang, Zhu and Chen}]{yang2021review}
Yang C, Zhu Y and Chen Y (2021) A review of human--machine cooperation in the robotics domain.
\newblock \emph{IEEE Transactions on Human-Machine Systems} 52(1): 12--25.

\bibitem[{Zhang et~al.(2022)Zhang, Wu, Chen, Zhu, Munawar, Xiao, Guan, Su, Hong, Guo et~al.}]{zhang2022human}
Zhang D, Wu Z, Chen J, Zhu R, Munawar A, Xiao B, Guan Y, Su H, Hong W, Guo Y et~al. (2022) Human-robot shared control for surgical robot based on context-aware sim-to-real adaptation.
\newblock In: \emph{2022 International conference on robotics and automation (ICRA)}. IEEE, pp. 7694--7700.

\bibitem[{Ziebart et~al.(2008)Ziebart, Maas, Bagnell, Dey et~al.}]{ziebart2008maximum}
Ziebart BD, Maas AL, Bagnell JA, Dey AK et~al. (2008) Maximum entropy inverse reinforcement learning.
\newblock In: \emph{Aaai}, volume~8. Chicago, IL, USA, pp. 1433--1438.

\end{thebibliography}

\end{document}